\documentclass[12pt, English]{report}
\usepackage{amsmath,amssymb,amsfonts}

\usepackage[algo2e]{algorithm2e} 
\usepackage{graphics}
\usepackage{epsfig}
\usepackage{psfrag}
\usepackage{setspace}
\usepackage{cite}
\usepackage{url}
\usepackage{titlesec}
\usepackage{amsmath}
\usepackage{multirow}
\usepackage{algorithm}
\usepackage[paper=portrait,pagesize]{typearea}

\usepackage{array}
\usepackage{lscape}
\usepackage{longtable}
\usepackage{threeparttable}
\usepackage{amssymb}
\usepackage{balance}  
\usepackage{graphicx} 
\usepackage{url}      
\usepackage[lofdepth,lotdepth]{subfig}
\usepackage{mathtools}
\usepackage{rotating}
\usepackage{breakurl}
\usepackage{array}
\usepackage{graphicx}
\usepackage{color}
\usepackage{textcomp}
\usepackage{esvect}
\usepackage{amsmath,stackengine}
\usepackage{url}
\usepackage{graphicx}
\usepackage{ccicons}                              

\newcommand\given[1][]{\:#1\vert\:}

\usepackage{multirow}
\usepackage{amsmath}

\usepackage{comment}
\usepackage{lscape}

\usepackage{amsmath,amssymb,amsfonts}
\usepackage{mathtools}
\usepackage{rotating}
\usepackage{breakurl}
\usepackage{array}
\usepackage{graphicx}
\usepackage{color}
\usepackage{textcomp}
\usepackage{esvect}
\usepackage{amsmath,stackengine}
\usepackage{bbm}
\usepackage{pseudocode}
\usepackage{color}
\usepackage[scientific-notation=true]{siunitx}
\DeclareMathOperator*{\argmax}{arg\,max}

\usepackage[textwidth=6in,textheight=9in,left=1.5in,right=1in,top=1in,bottom=1in]{geometry} 
\usepackage[colorlinks=true,
            linkcolor=red,
            urlcolor=blue,
            citecolor=blue]{hyperref} 
\usepackage{rotating}

\titleformat{\paragraph}[hang]{\normalfont\normalsize\bfseries}{\theparagraph}{1em}{}
\titlespacing*{\paragraph}{0pt}{3.25ex plus 1ex minus .2ex}{1em}

\newcommand{\qu}[1]{``#1''}

\titleformat{\section}
  {\normalfont\large\bfseries}{\thesection}{1em}{} 
\titleformat{\subsection}
  {\normalfont\normalsize\bfseries}{\thesubsection}{1em}{} 
\titleformat{\subsubsection}
  {\normalfont\normalsize\itshape}{\thesubsubsection}{1em}{} 

\usepackage{adjustbox}
\begin{document}
\singlespacing
\pagenumbering{gobble} 
\newgeometry{top=2in}
\begin{center}
		{\LARGE Diagnosis and Analysis of Celiac Disease and Environmental Enteropathy on Biopsy Images using Deep Learning Approaches \par}
~\\~\\~\\~\\
    \rule{11.5cm}{1pt}\\
    ~\\
    {A Dissertation \\}
    ~\\
    {Presented to \\}
    ~\\
    {the faculty of the School of Engineering and Applied Science \\}
    ~\\
    {University of Virginia \\}
    ~\\
    \rule{11.5cm}{1pt}\\
    {in partial fulfillment\\}
    ~\\
    {of the requirements for the degree\\}
    ~\\
    ~\\
    {Doctor of Philosophy\\}
    ~\\
    {by\\}
    ~\\
    {Kamran Kowsari}
    
\vfill
    {May\\2020}

		\thispagestyle{empty}
		
		\newpage
\pagenumbering{arabic}

{\huge \textbf{Approval Sheet}}\\
\vspace{30pt}
This Dissertation is submitted in partial fulfillment of the requirements for the degree of
Doctor of Philosophy (Systems and Information Engineering)

\vspace{10pt}
\begin{flushleft}
\rule{14.5cm}{1pt}\vspace{-3pt}\\
Kamran Kowsari\\
\vspace{10pt}

This Dissertation has been read and approved by the Examining Committee:
\vspace{30pt}
{

\rule{14.5cm}{1pt}\vspace{-3pt}\\
Laura Barnes, Ph.D. (School of Engineering and Applied Science)\\~\\~\\
\rule{14.5cm}{1pt}\vspace{-3pt}\\
Donald Brown, Ph.D. (School of Engineering and Applied Science)\\~\\~\\
\rule{14.5cm}{1pt}\vspace{-3pt}\\
Michael Porter, Ph.D. (School of Engineering and Applied Science)\\~\\~\\
\rule{14.5cm}{1pt}\vspace{-3pt}\\
Sana Syed, MD, MS (School of Medicine)\\~\\~\\
\rule{14.5cm}{1pt}\vspace{-3pt}\\
Peter Beling, Ph.D. (School of Engineering and Applied Science)\\~\\
}

Accepted for the School of Engineering and Applied Science:

\vspace{100pt}
\centering

\rule{12.5cm}{1pt}\vspace{-3pt}\\
Craig H. Benson, Dean, School of Engineering and Applied Science\\\vspace{+10pt}
May~2020

\end{flushleft}

\end{center}

\restoregeometry
\newcommand{\RNum}[1]{%
  \textup{\uppercase\expandafter{\romannumeral#1}}%
}

\null
\vfill
\begin{center}
\copyright Copyright by Kamran Kowsari 2020\\
~\\
All Rights Reserved
\end{center}
\setcounter{page}{3}  
\doublespacing
\chapter*{Abstract}
Celiac  Disease~(CD) and Environmental Enteropathy~(EE) are common causes of malnutrition and adversely impact normal childhood development. Both conditions require a tissue biopsy for diagnosis and a major challenge of interpreting clinical biopsy images to differentiate between these gastrointestinal diseases is striking histopathologic overlap between them. In the current study, we propose four diagnosis techniques for these diseases and address their limitations and advantages. 

First, the diagnosis between CD, EE, and Normal biopsies is considered, but the main challenge with this diagnosis technique is the staining problem. The dataset used in this research is collected from different centers with different staining standards. To solve this problem, we use color balancing in order to train our model with a varying range of colors. 

Random Multimodel Deep Learning (RMDL) architecture has been used as another approach to mitigate the effects of the staining problem. RMDL combines different architectures and structures of deep learning and the final output of the model is based on the majority vote. 

CD is a chronic autoimmune disease that affects the small intestine genetically predisposed children and adults. Typically, CD rapidly progress from Marsh I to IIIa. Marsh III is sub-divided into IIIa~(partial villus atrophy), Marsh IIIb~(subtotal villous atrophy) and Marsh IIIc~(total villus atrophy) to explain the spectrum of villus atrophy along with crypt hypertrophy and increased intraepithelial lymphocytes. In the second part of this study, we proposed two ways for diagnosing different stages of CD. 

Finally, in the third part of this study, these two steps are combined as Hierarchical Medical Image Classification~(HMIC) to have a model to diagnose the disease data hierarchically.
\chapter*{Acknowledgements}
\doublespacing
It is a great pleasure to thank my committee members for their time and support. First, I would like to express my sincere gratitude to my advisor, Dr. Laura Barnes, for her constant encouragement and guidance throughout my study at UVa. Without her support, I wouldn't be able to complete this dissertation. Also, I would like to thank my co-advisor Donald Brown for his advice and help on my graduate study. To my wife, Sanam: thank you for everything you did and said for the last five years. To my mother, Giti, your support has been the driving motive for me to work hard and succeed. To my siblings and family: thank you for your support.
{ \hypersetup{linkcolor=black}
\setcounter{tocdepth}{7}
\tableofcontents
\cleardoublepage
\addcontentsline{toc}{chapter}{List of Tables}
\listoftables
\cleardoublepage
\addcontentsline{toc}{chapter}{List of Figures}
\listoffigures
\cleardoublepage
}

\doublespacing

\chapter{Introduction}
Celiac Disease~(CD) is a chronic autoimmune disease that affects the small intestine in genetically predisposed children and adults~\cite{kowsari2019diagnosis,sali2019celiacnet,syed2019assessment}. Gluten exposure triggers an inflammatory cascade which leads to compromised intestinal barrier function. If this enteropathy is unrecognized, it can lead to anemia, decreased bone density and in longstanding cases, intestinal cancer. Celiac Disease and Environmental Enteropathy~(EE)~\cite{iqbal2019study} are common causes of malnutrition and adversely impact normal childhood development. CD is an autoimmune disorder that is prevalent worldwide and is caused by an increased sensitivity to gluten. Gluten exposure destructs the small intestinal epithelial barrier, resulting in nutrient mal-absorption and childhood under-nutrition. EE also results in barrier dysfunction but is thought to be caused by an increased vulnerability to infections. EE has been implicated as the predominant cause of under-nutrition, oral vaccine failure, and impaired cognitive development in low-and-middle-income countries. Both conditions require a tissue biopsy for diagnosis, and a major challenge of interpreting clinical biopsy images to differentiate between these gastrointestinal diseases is striking histopathologic overlap between them. Studies have shown the ease of training Convolutional Neural Networks~(CNNs) for image recognition. These networks are an iterative family of machine learning architectures. CNNs have proven to have superior performance over a wide range of computer vision tasks such as classification and object detection. Due to the wide availability of robust open-source software and high-quality public datasets, these architectures are becoming the popular choice for being selected as the backbone of many modern computer vision technologies. Using large amounts of data, these models have shown to be effective in solving many biomedical imaging challenges. Currently, CNNs have been successfully applied to medical images such as MRI and X-rays~\cite{gulshan2016development, litjens2017survey}. The CNNs have also shown promising performance on histopathological images~\cite{kowsari2019diagnosis,Mohammad_al_boni}.

In this dissertation, we propose four techniques for diagnosis of celiac disease and Environmental Enteropathy based on biopsy images. The first technique uses Color Balancing~(CB) on Convolutional Neural Networks~(CNN). The second method is using Random Multimodel Deep Learning~(RMDL) for the diagnosis of celiac disease and Environmental Enteropathy. The third technique is used to diagnosis only in the stage of Celiac Disease Severity Diagnosis on Duodenal Histopathological images using Deep Learning~(DL). Finally, we use an approach we call Hierarchical Medical Image classification (HMIC) to combine these models as the hierarchical representations of theses biopsy images. In this dissertation, two main ideas have been covered to the diagnosis of diseased duodenal architecture on biopsy images. We proposed a data-driven model for diagnosis of diseased duodenal architecture on biopsy images using color balancing on convolutional neural networks. The validation results of this model show that it can be utilized by pathologists in diagnostic operations regarding CD and EE. The second approach called RMDL, is used for the medical image classification. It combines multiple deep learning architectures to produce random classification models. We investigated CD severity by applying Shallow CNNs and residual neural network architecture to histopathological images.

In this dissertation, we propose a technique called Hierarchical Medical Image Classification (HMIC), that combines multiple deep learning approaches to produce hierarchical classifications. The method that is presented in this dissertation described here can be improved in multiple ways. Additional training and testing with other hierarchically structured medical data sets will continue to identify architectures that work best for these problems. Also, it is possible to extend the hierarchy to more than two levels to capture more of the complexity in the hierarchical classification.

The rest of the dissertation is organized into five chapters which includes a literature review in Chapter~\ref{Chpt:Review}, discussing related research on Diagnosis of Celiac Disease and Environmental Enteropathy using Biopsy Images, the other kind of existing classification techniques~(traditional techniques). Chapter~\ref{Chpt:Review} also includes a summary of the existing medical image classification techniques, and finally, evaluation techniques have been discussed.

In Chapter~\ref{chpt:Celiac_color}, we talk about the technique to the diagnosis of Celiac Disease and Environmental Enteropathy on biopsy images using Color Balancing on Convolutional Neural Networks~(CNN). 

In Chapter~\ref{chpt:RMDL}, Random Multimodel Deep Learning~(RMDL) is introduced. We apply this technique on resized biopsy images. 

In Chapter~\ref{chpt:CeliacNet}, Celiac Disease Severity diagnosis on Duodenal Histopathological Images is discussed with using Deep Learning approaches. In this Chapter, we use two popular techniques call Shallow Convolutional Neural Networks~(CNN) and deep residual networks~(ResNet) with color normalization. All of data in this section is collected from one center so, we did not use color balancing in this section.

In Chapter~\ref{chpt:HMIC}, Hierarchical Medical Image classification (HMIC) is introduced which is a hierarchical deep learning approach. This Chapter in a combination of Chapter~\ref{chpt:Celiac_color} and Chapter~\ref{chpt:CeliacNet} present hierarchical representations of biopsy images.
\chapter{Literature Review}\label{Chpt:Review}
\section{Diagnosis of Celiac Disease and Environmental Enteropathy on Biopsy Images}\label{Sec:LR:Diagnosis}
Under-nutrition is the underlying cause of approximately~$45$\% of the~$5$ million under~$5$-year-old childhood deaths annually in low and middle-income countries (LMICs)~\cite{WHO.Children} and is a major cause of mortality in this population. Linear growth failure (or stunting) is a major complication of under-nutrition, and is associated with irreversible
physical and cognitive deficits, with profound developmental implications~\cite{syed2016environmental}. A common cause of stunting in
LMICs is EE, for which there are no universally accepted, clear diagnostic algorithms or non-invasive
biomarkers for accurate diagnosis~\cite{syed2016environmental}, making this a critical priority~\cite{naylor2015environmental}. EE has been described to be caused by 
chronic exposure to enteropathogens which results in a vicious cycle of constant
mucosal inflammation, villous blunting, and a damaged epithelium~\cite{syed2016environmental}. These deficiencies contribute to a markedly reduced nutrient absorption and thus under-nutrition and stunting~\cite{syed2016environmental}. Interestingly, CD, a common cause of stunting in
the United States, with an estimated~$1$\% prevalence, is an autoimmune disorder caused by a gluten sensitivity~\cite{husby2012european} and has many shared histological features with EE~(such as increased inflammatory cells and villous blunting)~\cite{syed2016environmental}. This resemblance has led to the major challenge of differentiating clinical biopsy images for these similar but distinct diseases. 
Therefore, there is a major clinical interest towards developing new, innovative methods to automate and enhance the detection of morphological features of
EE versus CD, and to differentiate between diseased and healthy small intestinal tissue~\cite{bejnordi2017diagnostic}. 

We propose a CNN-based model for the classification of biopsy images. In recent years, Deep Learning architectures have received great attention after achieving state-of-the-art results in a wide variety of fundamental tasks such as classification~\cite{Heidarysafa2018RMDL,kowsari2017hdltex,kowsari2018rmdl,heidarysafa2018analysis,info10040150,litjens2017survey,nobles2018identification,zhai2016doubly,kowsari2020gender} or other medical domains~\cite{hegde2019comparison,zhang2018patient2vec}. CNNs in particular have proven to be very effective in medical image processing. CNNs preserve local image relations, while reducing dimensionality and for this reason are the most popular machine learning algorithm in image recognition and visual learning tasks~\cite{ker2018deep}. CNNs have been widely used for classification and segmentation in various types of medical applications such as histopathological
images of breast tissues, lung images, MRI images,  medical X-Ray images, etc.~\cite{gulshan2016development,litjens2017survey}. Researchers produced advanced results on duodenal biopsies classification using CNNs~\cite{Mohammad_al_boni}, but those models are only robust to a single type of image stain or color distribution. Many researchers apply a stain normalization technique as part of the image pre-processing stage to both the training and validation datasets~\cite{nawaz2018classification}. In this section, varying levels of color balancing were applied during image pre-processing in order to account for multiple stain variations.

\subsection{Automated Detection of Celiac Disease using Deep Learning}
\begin{figure}[H]
    \centering
    \includegraphics[width=\textwidth]{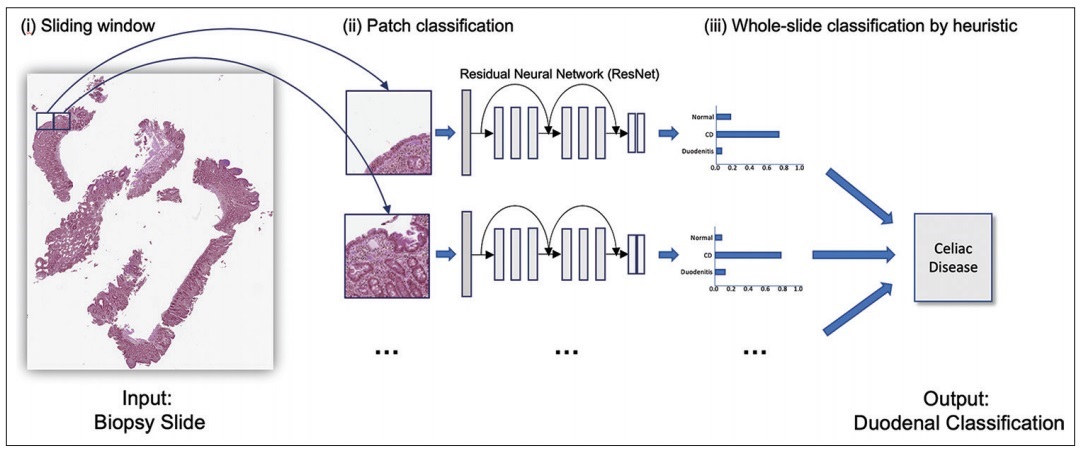}
    \caption{Wei et. al~\cite{wei2019automated} works on celiac disease on whole‑slide biopsy images}\label{fig:wei2019automated}
\end{figure}

Wei et. al~\cite{wei2019automated} introduced an automated detection technique for Celiac Disease on duodenal biopsy. As shown in Figure~\ref{fig:wei2019automated}, they convert their biopsy to patches and use deep residual network (ResNet-50) model to classify each patch, and they filtered out noise using thresholding to discard predictions of low confidence.

\begin{figure}[!b]
    \centering
    \includegraphics[width=\textwidth]{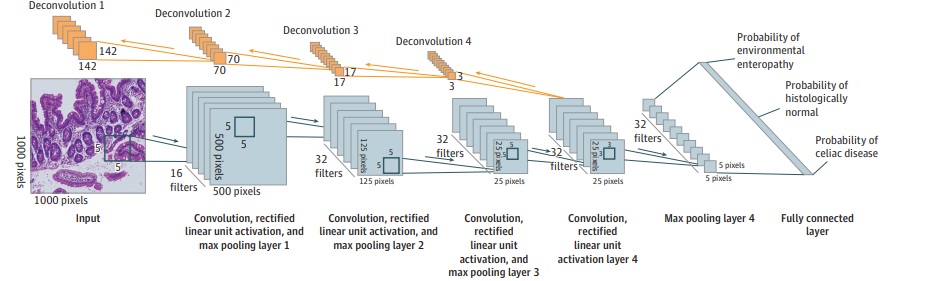}
    \caption{S. Syed et. al~\cite{syed2019assessment} works on Celiac Disease and Environmental
Enteropathy on biopsy images using CNN}\label{fig:syed2019assessment}
    \label{fig:my_label}
\end{figure}

S. Syed et. al~\cite{syed2019assessment,al2019duodenal}, use Convolutional Neural Network~(CNN) on duodenal biopsies of patients with environmental enteropathy, celiac disease, and histologically normal controls which automatically learned microlevel features in duodenal tissue, such as alterations in secretory cell populations. As shown in Figure~\ref{fig:syed2019assessment}, this model uses 4 Convolutional layers and followed by a pooling layer and connected to the fully connected  Neural Network layer.

\subsection{Automated Detection of Celiac Disease Using Traditional Technique}

A. Ermarth et. al~\cite{ermarth2017identification} works on the Identification of Pediatric Patients With Celiac Disease based on a classification and regression tree  Analysis~(CART). CART The CART is a predictive model that can be represented as a decision tree, and it recursively partitions the dataset by doing binary splits to find the optimal tree~\cite{lim2017failure}.  

K. Kowal~\cite{kowal2014computer} presents an automatic computer system a set of 84 features extracted from the nuclei is
used in the classification by the k-nearest neighbor (kNN) classifier. In this research, this technique was tested on 450 microscopic images from biopsies, and the proposed computer-aided diagnosis (CAD) system is able to classify with reasonable accuracy.

E. Radiya et. al.~\cite{radiya2017automated}, works on Active Feature Extraction (CAFE) consisting of two logistic regression algorithms on biopsy images from 392 biomarkers, and they get acceptable AUC when trained on data from one hospital and tested on samples of the other.

\section{Image Classification Techniques}\label{sec:techniques}
Many techniques are used in the past decades to diagnosis celiac disease such as SVM~\cite{hegenbart2011impact}, Random Forest~\cite{fathi2014differential}, ensemble learning~\cite{habtamu2015sensitive}, decision-tree~\cite{lionetti2014introduction}, Logistic Regression~\cite{pagadala2013diagnosis}, etc. These days, deep learning to diagnosis this disease from Normal and Environmental Enteropathy~(EE) is very popular~\cite{syed2019assessment,sali2019celiacnet,syed2018195,kowsari2019diagnosis}. In this sub-section, we describe the popular image classification methods within medical domain.   Over the last 50 years, many supervised learning such as image classification techniques have been developed and implemented in software to accurately label  data. For example, the researchers, K. Murphy in~2006~\cite{murphy2006naive} and I. Rish in~2001~\cite{rish2001empirical} introduced the Na\"ive Bayes Classifier~(NBC) as a simple approach to the more general representation of the supervised learning classification problem and this technique was developed for image classification task~\cite{mccann2012local}. This approach has provided a useful technique for data classification and information retrieval applications. As with most supervised learning classification techniques, NBC takes an input vector of numeric or categorical data values and produce the probability for each possible output labels. Another  popular classification technique in 1980s and 1990s was Support Vector Machines~(SVM), which has proven quite accurate over a wide variety of data. This technique constructs a set of hyper-planes in a transformed feature space. This transformation is not performed explicitly but rather through the kernal trick which allows the SVM classifier to perform well with highly nonlinear relationships between the predictor and response variables in the data.  A variety of approaches have been developed to further extend the basic methodology and obtain greater accuracy. C. Yu~\textit{et. al.} in 2009~\cite{yu2009learning} introduced  latent variables into the discriminative model as a new structure for SVM, and S. Tong~\textit{et. al.} in 2001~\cite{tong2001support} added active learning using SVM for data classification. For a large volume of data and  datasets with a huge number of features~(such as image or any kind of datasets), SVM implementations are computationally complex. Another technique that helps mediate the computational complexity of the SVM for classification tasks is stochastic gradient descent classifier~(SGDClassifier)~\cite{kabir2015bangla} which has been widely used in image classification. SGDClassifier is an iterative model for large datasets. The model is trained based on the SGD optimizer iteratively.

\subsection{Logistic Regression}
One of the earliest methods of classification is logistic regression~(LR) that is used for diagnosis CD~\cite{pagadala2013diagnosis}. LR was introduced and developed by statistician David Cox in 1958~\cite{cox2018analysis}. LR is a linear classifier with decision boundary of $\theta^Tx=0$. LR predicts probabilities rather than classes~\cite{fan2008liblinear,genkin2007large}. Many medical image classification application was developed by using LR~\cite{rao2011classification}.

\subsubsection{Basic Framework}
The goal of LR is to trained from the probability of variable~$Y$ being 0 or 1 given $x$. Let's have data data which is $X \in \mathbb{R}^{n\times d}$. If we have binary classification problems, the Bernoulli mixture models function should be used~\cite{juan2002use}.

\subsubsection*{Combining Instance-based Learning and LR:}
The LR model specifies the probability of binary output $y_i=\{0,1\}$ given the input $x_i$. To obey the basic principle underlying instance-based learning~(IBL)~\cite{cheng2009combining}, the classifier should be a function of the distance~$\delta_i$. $p$ will be large if ~$\delta_i\rightarrow 0$ then~$y_i=+1$, and  small for $y_i=-1$. $p$ should be close to~$1$ if ~$\delta_i\rightarrow \infty$; then, neither in favor of~$y_0 = +1$ nor in favor of~$y_0 = -1$. Multinomial (or multilabeled) logistic classification~\cite{krishnapuram2005sparse} uses the probability of~$x$ belonging to class~$i$. In a classification task as supervised learning context, the component of~$\theta$ is calculated from the subset of the training data~$D$ which belongs to class~$i$ where~$i \in\{1,\hdots,n\}$. To perform maximum likelihood~(ML) estimation of~$\theta$, we need to maximize the log-likelihood function~\cite{info10040150}.

\subsection{K-Nearest Neighbor}
The k-nearest-neighbors algorithm (k-NN) is a non-parametric technique used for classification. This method is used for data classification applications such as medical image classification in many research domains~\cite{jiang2012improved} in past decades.

\subsubsection{Basic Concept of k-NN}
\begin{figure}[t]
    \centering
    \includegraphics[width=0.7\textwidth]{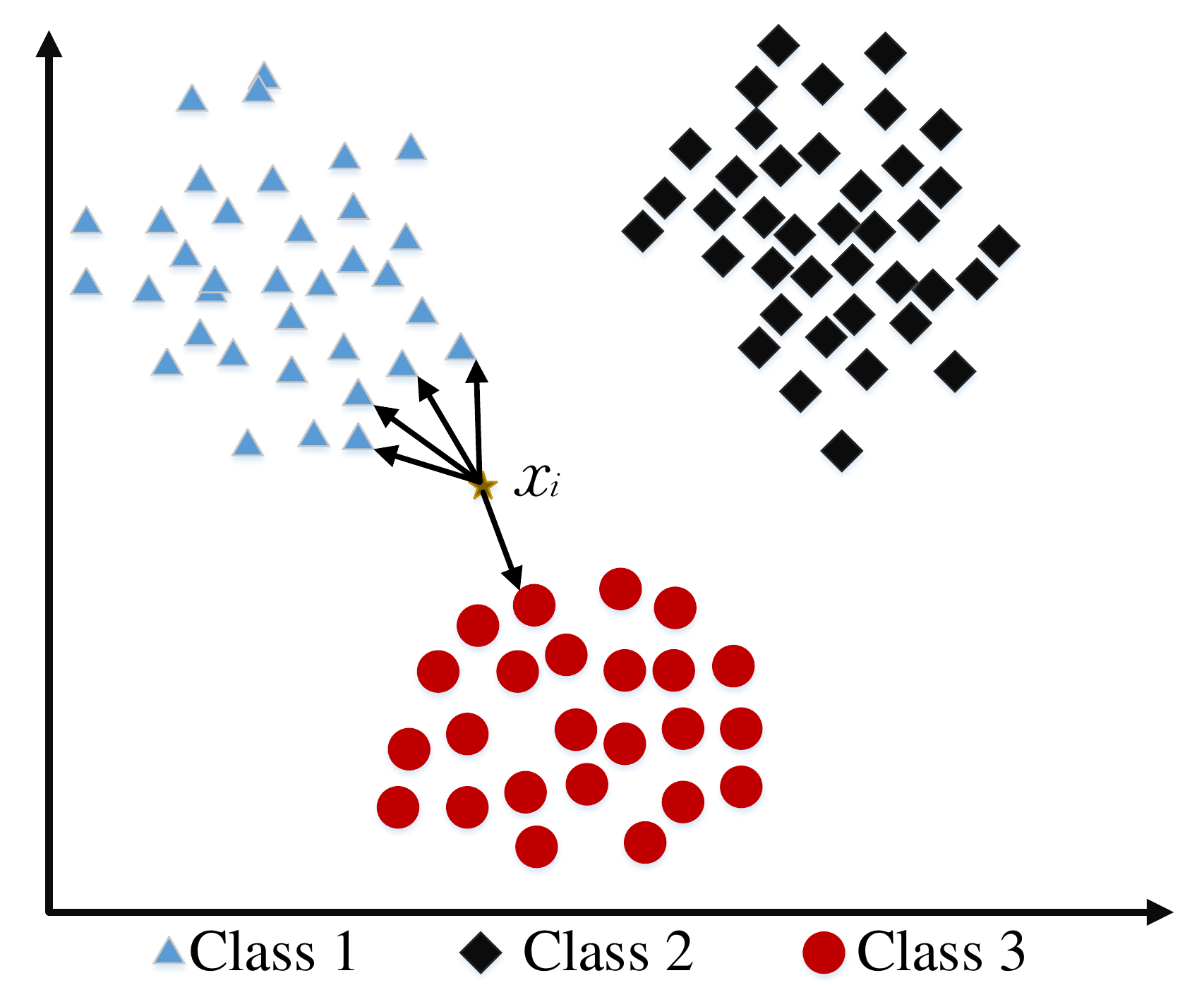}
    \caption{A architecture of k-NN model for 2D dataset and three classes} \label{fig:KNN}
\end{figure}

Given a test set~$x$, the k-NN algorithm finds the~$k$ nearest neighbors of~$x$ among all training set, and scores the category candidates based the class of~$k$ neighbors. The similarity of~$x$ and each neighbor's images could be the score of the category of the neighbor images. Multiple k-NN images may belong to the same category; in this case, the summation of these scores would be the similarity score of the class~$k$ with respect to the test image~$x$. After sorting the score values, the algorithm assigns the candidate to the class with the highest score from the test image~$x$~\cite{jiang2012improved}. The Figure~\ref{fig:KNN} illustrates the k-NN architecture, but for simplicity, this figure is designed by 2D dataset~(similar with higher dimensional space). The decision rule of k-NN is:

\begin{equation}
\begin{split}
    f(x) = &arg \max_j S(x,C_j) \\ = &\sum_{d_i\in k-NN} sim(x,d_i)y(d_i,C_j)
\end{split}
\end{equation}
where S refers to score value with respect to $S(x,C_j)$, the score value of candidate~$i$ to class of~$j$, and output of $f(x)$ is a label to the test set image.

\subsection{Support Vector Machine~(SVM)}\label{me_svm}
The original version of SVM was developed by ~\textit{Vapnik and  Chervonenkis} \cite{vapnik1964class} in 1963. ~\textit{BE. Boser et al.}~\cite{boser1992training} adapted this version into a nonlinear formulation in the early 1990s. SVM was originally designed for binary classification tasks. However, many researchers work on multi-class problems using this dominate technique~\cite{bo2006svm}. 

In 2015, Z. Camlica~\cite{camlica2015medical} designed a medical image classification using SVM which this technique using using local binary patterns (LBP) features from saliency-based folded data.

\begin{figure}[t]
    \centering
    \includegraphics[width=\textwidth]{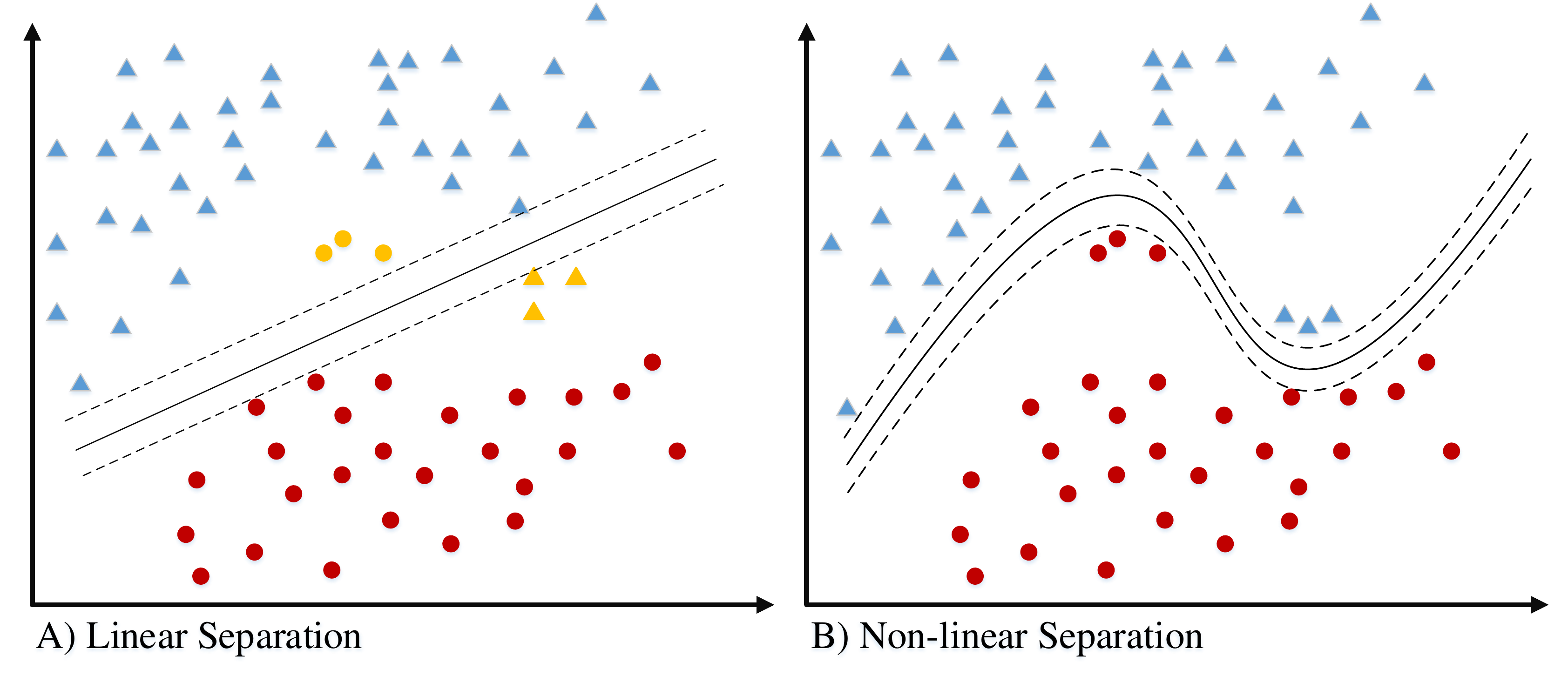}
    \caption{This figure shows the linear and non-linear SVM for 2D dataset~(for text data we have thousands of dimitions )}  \label{fig:SVM}
\end{figure}

In the context of image classification, let $x_1, x_2,...,x_l$ be training examples belonging to one class $X$, where X is a compact subset of $R^N$~\cite{manevitz2001one}. Since SVMs are traditionally used for the binary classification, we need to generate a Multiple-SVM ~(MSVM)~\cite{mohri2012foundations} for multi-class problems. One-vs-One is a technique for multi-class SVM that builds N(N-1) classifiers. The natural way to solve the k-class problem is to construct a decision function of all $k$ classes at once~\cite{chen2016turning,weston1998multi}. Another technique of multi-class classification using SVM is All-vs-One. 

\subsection{Decision Tree}
One earlier classification algorithm for image and data mining is decision tree~\cite{morgan1963problems}. This techniques is used for analysis the risk of celiac disease in children~\cite{lionetti2014introduction}. Decision tree classifiers (DTCs) are used successfully in many diverse areas for classification~\cite{safavian1991survey}. The structure of this technique is a hierarchical decomposition of the data space
~\cite{morgan1963problems,aggarwal2012survey}. Decision tree as classification task was introduced by~{D. Morgan}~\cite{magerman1995statistical} and developed by~{J.R. Quinlan}~\cite{quinlan1986induction}. The main idea is creating a tree based on the attribute for categorized data points, but the main challenge of a decision tree is which attribute or feature could be in parents' level and which one should be in child level.
To solve this problem,~{De M{\'a}ntaras}~\cite{de1991distance} introduced statistical modeling for feature selection in tree.

\subsection{Ensemble techniques}
Voting classification techniques, such as Bagging, Boosting, Random Forest, and RMDL have been successfully developed for  classification~\cite{farzi2016estimation} tasks. While Boosting adaptively changes the distribution of the training set based on the performance of previous classifiers, Bagging does not look at the previous classifier\cite{bauer1999empirical}.

\subsubsection{Random Forest}
\begin{figure}[!t]
\centering
\includegraphics[width=0.7\textwidth]{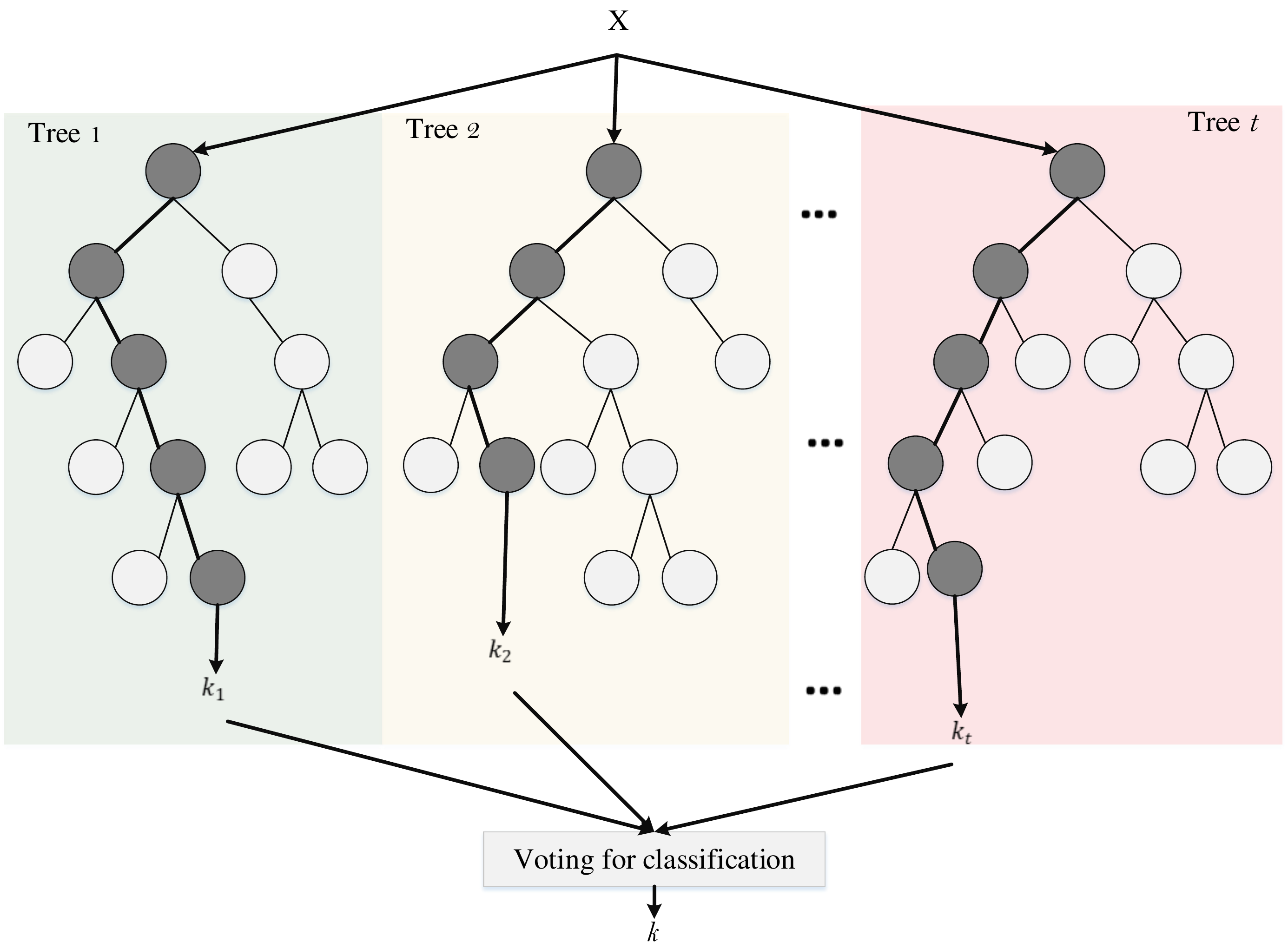}
\caption{Random Forest}\label{Fig:RF}
\end{figure}

Random Forest is used to diagnosis of Crohn’s disease and celiac disease using nuclear magnetic resonance spectroscopy~\cite{fathi2014differential}. This technique is used in many way for medical image classification which Random forests or random decision forests technique is an ensemble learning method for medical image classification~\cite{subudhi2020automated}. This method, which used $t$ tree as parallel, was introduced by~\textit{T. Kam Ho}~\cite{Ho1995RF} in 1995.  As shown in Figure~\ref{Fig:RF}, main idea of RF is generating random decision trees. In Chapter~\ref{chpt:RMDL}, we use very similar technique but instead of trees we use  Random Deep Learning in our forest. This technique was further developed in 1999 by~\textit{L. Breiman}~\cite{breiman1999random}, who found convergence for RF as margin measures~($mg(X,Y)$).

\paragraph*{Voting}
After training all trees as forest, predictions are assigned based on voting~\cite{wu2004probability} as follows:

\begin{equation}
    \delta _V  = arg\max_i \sum_{j:j \neq j} I_{\{r_{ij} > r_{ji}\}}
\end{equation}
such that

\begin{equation}
    r_{ij}+r_{ji} = 1
\end{equation}

\SetInd{0.25em}{0.1em}
\begin{algorithm}
\caption{A1}
\SetKwInOut{Input}{input}\SetKwInOut{Output}{output}
\SetKwFor{Foreach}{for}{do}{endfor}
\SetKwIF{If}{ElseIf}{Else}{if}{then}{else if}{else}{endif}
\BlankLine
\Input{training set~S of size m, inducer~$\tau$, integer $N$}
\BlankLine
\Foreach { $i=1$ to $N$}{

~~~~$C_i=\tau(S')$ 
\begin{flalign}\nonumber
    ~~~&\epsilon_i = \frac{1}{m} \sum_{x_j\in S'; C_i(x_j)\notin y_i} weight(x)&&
\end{flalign}
\If{$\epsilon_i > \frac{1}{2}$}{

~~~~~~~~set S' to a bootstrap sample from S with weight 1 for 

~~~~~~~~all instance and go top

~~~~}

~~~~$\beta_i = \frac{\epsilon_i}{1-\epsilon_i}$

\Foreach { $x_i\in S'$}{

\If{$C_i(x_j)=y_i$}{

~~~~~~~~$weight(x_j) = weight(x_j) . \beta_i$ 

}

}

~~~~Normalize weights of instances

}

\vspace{-20pt}

\begin{flalign}\nonumber
    &C^*(x)= arg\max_{y\in Y} \sum_{i,C_i(x)=y} \log \frac{1}{\beta_i}&&
\end{flalign}

\Output{Classifier $C^*$}
\caption{The AdaBoost method}\label{alg:AdaBoost_Method}
\end{algorithm}

\begin{figure}[!b]
    \centering
    \includegraphics[width=\textwidth]{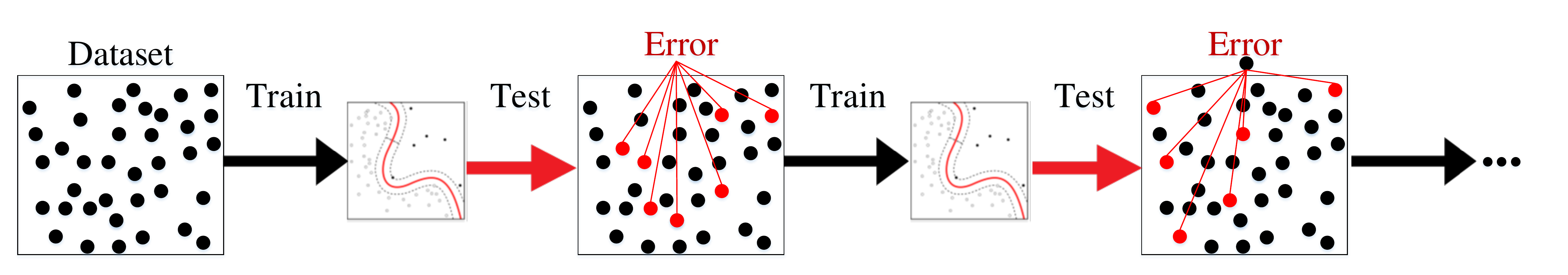}
    \caption{This figure is the Boosting technique architecture} \label{fig:Boosting}
\end{figure}

\subsubsection{Boosting}
The Boosting algorithm was first introduced by~\textit{RE. Schapire}~\cite{schapire1990strength} in 1990 as a technique for boosting the performance of a weak learning algorithm. This technique was further developed by Freund~\cite{freund1992improved,bloehdorn2004boosting}.

The Figure~\ref{fig:Boosting} shows how boosting algorithm works for 2D datasets, as it shown we have labeled data then we trained by multi-model architectures~(ensemble learning). These developments resulted in the AdaBoost (Adaptive Boosting)~\cite{freund1995efficient}.

\begin{algorithm}[!b]
\SetKwInOut{Input}{input}\SetKwInOut{Output}{output}
\SetKwFor{Foreach}{for}{do}{endfor}
\SetKwIF{If}{ElseIf}{Else}{if}{then}{else if}{else}{endif}
\BlankLine
\Input{training set~S, Inducer~$\tau$, integer $N$}
\BlankLine
\Foreach { $i=1$ to $N$}{
~~~~$S'$ = bootstrap sample from S

~~~~$C_i$ = $\tau(S')$
}
\vspace{-50pt}
\begin{flalign}\nonumber
    &C^*(x)= arg\max_{y\in Y} \sum_{i,C_i=y} 1&&
\end{flalign}

\Output{Classifier $C^*$}
\caption{Bagging}\label{Method}
\end{algorithm}

\subsubsection{Bagging}
\begin{figure}[t]
    \centering
    \includegraphics[width=0.8\textwidth]{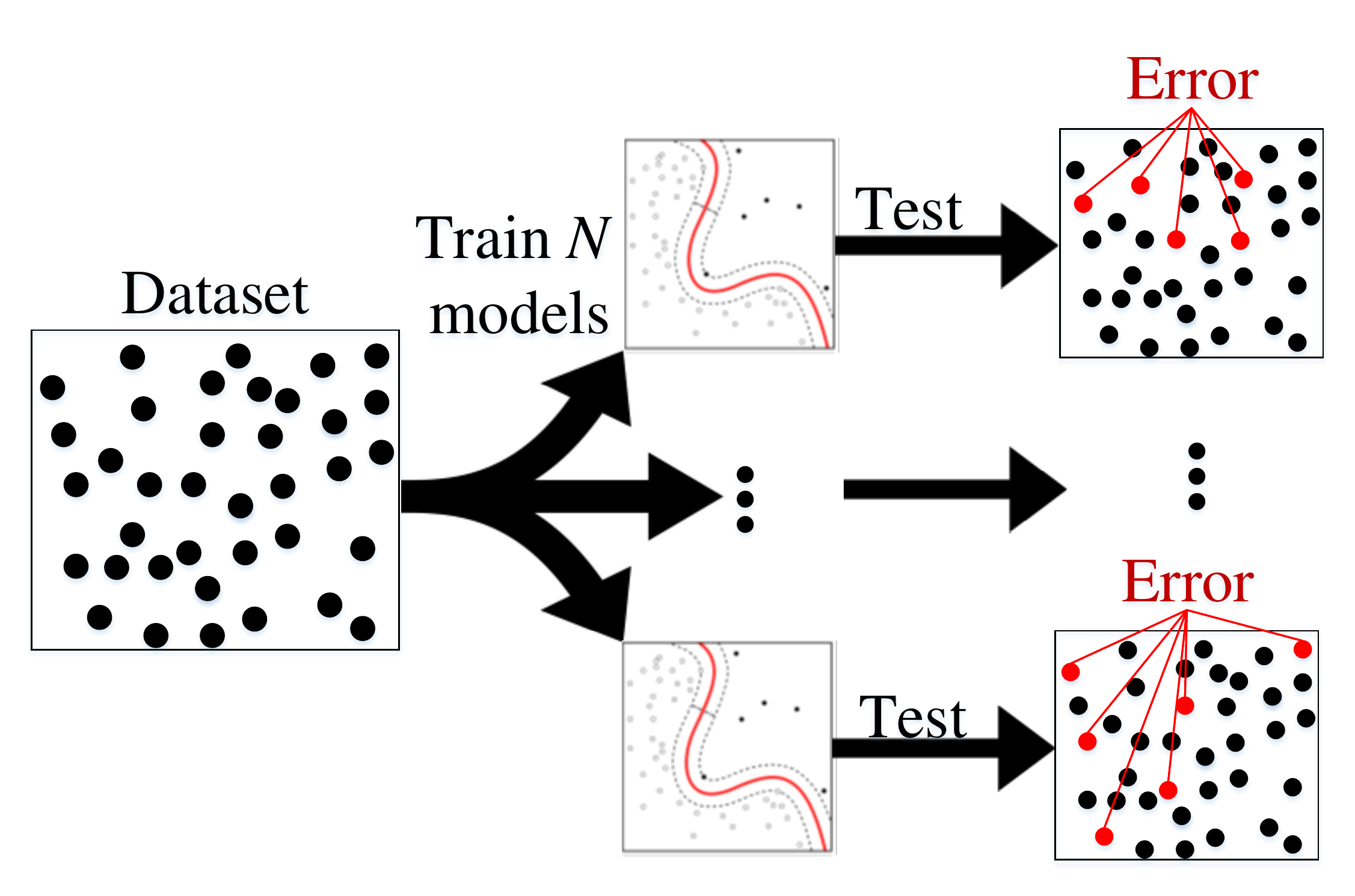}
    \caption{This figure shows a simple model of Bagging technique} \label{fig:Bagging}
\end{figure}

The Bagging algorithm was introduced by~\textit{L. Breiman}~\cite{breiman1996bagging} in 1996 as a voting classifier method. The algorithm is generated by different bootstrap samples\cite{bauer1999empirical}. A bootstrap generates a uniform sample from the training set. If~$N$ bootstrap samples $B_1, B_2,\hdots, B_N$ have been generated, then we have N classifiers~($C$)  which $C_i$ is built from each bootstrap sample $B_i$. Finally, our classifier~$C$ contain or generated from $C_1,C_2,...,C_N$ whose output is the class predicted most often by its sub-classifiers, with ties broken arbitrarily\cite{bauer1999empirical,breiman1996bagging}.
Figure~\ref{fig:Bagging} shows a simple Bagging algorithm which trained~$N$ models. 

\subsection{Deep Learning:}\label{subsec:related3}
Neural networks derive their architecture as a relatively simple representation of the neurons in the human's brain. They are essentially weighted combinations of inputs that pass through multiple non-linear functions. Neural networks use an iterative learning method known as back-propagation and an optimizer~(such as stochastic gradient descent~(SGD)). 

Multilayer perceptron or Deep Neural Networks~(DNN) are based on simple neural networks architectures but they contain multiple hidden layers. These networks have been widely used for classification. For example, D. Cire{\c{s}}An~\textit{et. al.} in 2012~\cite{ci2012multitraffic} used multi-column deep neural networks for classification tasks, where  multi-column deep neural networks use Multilayer perceptron architectures. Convolutional Neural Networks~(CNN) provide a different architectural approach to learning with neural networks. The main idea of CNN is to use feed-forward networks with convolutional layers that include local and global pooling layers. A. Krizhevsky in 2012~\cite{krizhevsky2012imagenet} used CNN, but they have used~$2D$ convolutional layers combined with the ~$2D$ feature space of the image. Another example of CNN in~\cite{lecun2015deep} showed excellent accuracy for image classification. The final type of deep learning architecture is Recurrent Neural Networks~(RNN) where outputs from the neurons are fed back into the network as inputs for the next step. Some recent extensions to this architecture uses Gated Recurrent Units~(GRUs)~\cite{chung2014empirical} or Long Short-Term Memory~(LSTM) units~\cite{hochreiter1997long}. These new units help control for instability problems in the original network architecure. RNN have been successfully used for natural language processing~\cite{mikolov2010recurrent}. Recently, In ~\cite{mao2016hierarchical} in~2016 developed A hierarchical convolutional neural network for mitosis detection in phase-contrast microscopy images. These networks have two important characteristics: hierarchical structure and an attention mechanism. 

New work has combined these three basic models of the deep learning structure and developed a novel technique for enhancing accuracy and robustness. The work of M. Turan~\textit{et. al.} in 2017~\cite{turan2017deep} and M. Liang~\textit{et. al.}in 2015~\cite{liang2015recurrent} implemented innovative combinations of CNN and RNN called \textit{A Recurrent Convolutional Neural Network~(RCNN)}.

\section{Autoencoder}
An autoencoder is a type of neural network that is trained to attempt to copy its input to its output~\cite{goodfellow2016deep}. In this dissertation, we use autoencoder in all chapters to select only useful images for training and discard useless information~(see Figure~\ref{fig:AE}). The autoencoder has achieved great success as a dimensionality reduction, abnormally detection, etc.  method via the powerful reprehensibility of neural networks~\cite{wang2014generalized}. The first version of autoencoder was introduced by {D.E. Rumelhart et~al.}~\cite{rumelhart1985learning} in 1985. The main idea is that one hidden layer between input and output layers has fewer units~\cite{liang2017text} and could thus be used to reduce the dimensions of a feature space. Especially for texts, documents, and sequences that contain many features, using an autoencoder could help allow for faster, more efficient data processsing.

\subsection{General Framework}
\begin{figure}[h]
\centering
\includegraphics[width=0.9\textwidth]{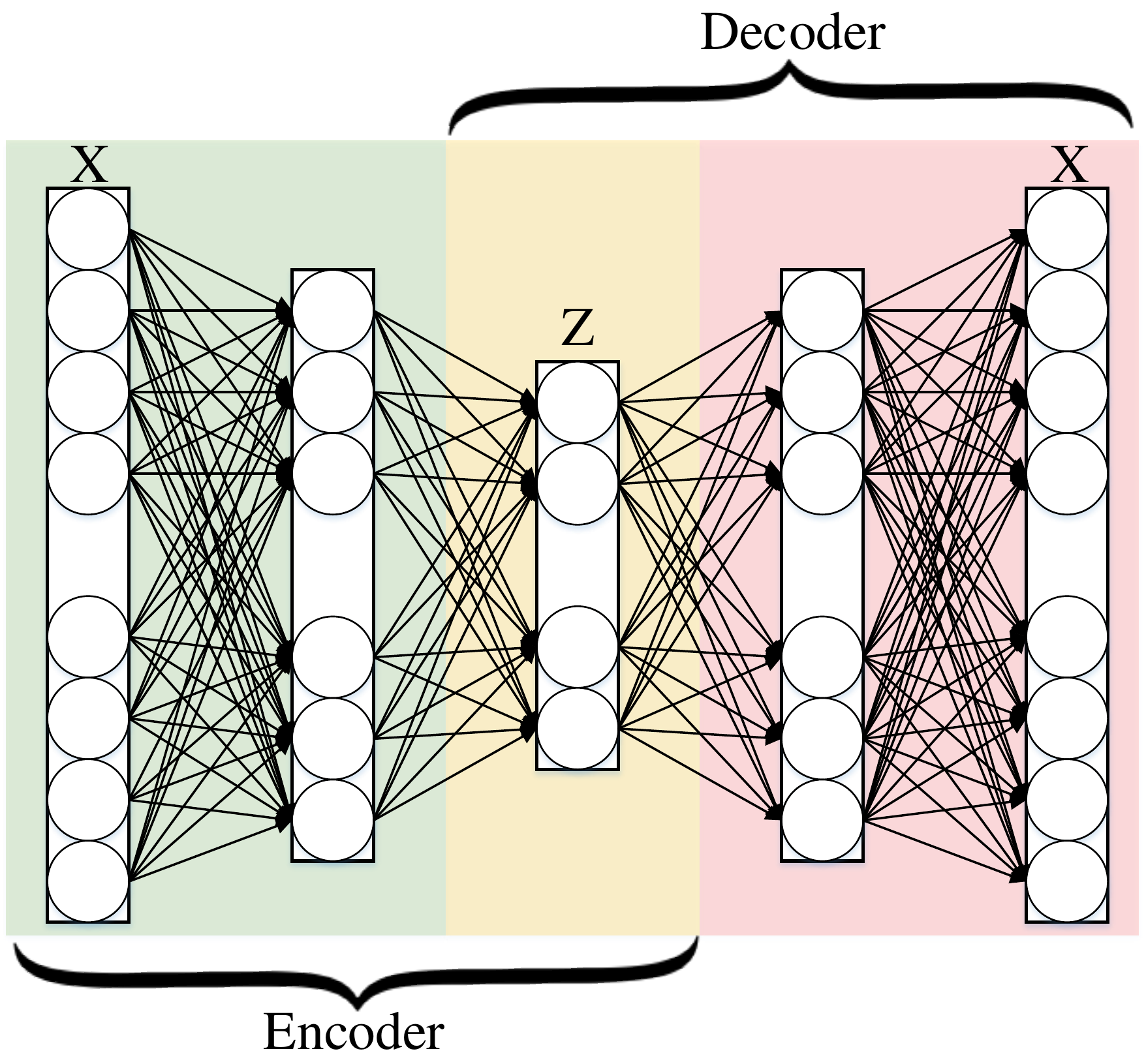}
\caption{This figure shows how a simple autoencoder works. The model depicted contains the following layers: $Z$ is code and two hidden layers are used for encoding and two are used for decoding.} \label{fig:Autoencoder}
\end{figure}

As shown in Figure~\ref{fig:Autoencoder}, the input and output layers of an autoencoder contain~$n$ units where~$x = \mathbb{R}^n$, and hidden layer~$Z$ contains~$p$ units with respect to~$p<n$~\cite{baldi2012autoencoders}. For this technique of dimensionality reduction, the dimensions of the final feature space are reduced from~$n\rightarrow p$. The encoder representation involves a sum of the representation of all words~(for bag-of-words), reflecting the relative frequency of each word~\cite{ap2014autoencoder}:
\begin{equation}
a(x)= c+\sum_{i=1}^{|x|} W_{.,x_i}, \phi(x)=h(a(x))
\end{equation}
where~$h(.)$ is an element-wise non-linearity such as the sigmoid~(see Equation~\eqref{sigmoid}).

\subsection{Conventional Autoencoder Architecture}
As shown in Figure~\ref{fig:AE}, we used Conventional Autoencoder Architecture in this study. A Convolutional Neural Network~(CNN)-based autoencoder can be divided into two main steps~\cite{masci2011stacked}~(encoding and decoding).
\begin{equation}
O_m(i, j) = a\bigg(\sum_{d=1}^{D}\sum_{u=-2k-1}^{2k+1}\sum_{v=-2k -1}^{2k +1} F^{(1)}_{m_d}(u, v)I_d(i -u, j -v)\bigg)~~\quad \forall  m = 1, \cdots, n
\end{equation}
where~$F \in \{F^{(1)}_{1},F^{(1)}_{2},\hdots,F^{(1)}_{n},\}$ which is a convolutional filter, with convolution among an input volume defined by~$I = \left\{I_1,\cdots, I_D\right\}$ which learns to represent input combining non-linear functions:
\begin{equation}
z_m = O_m = a(I * F^{(1)}_{m} + b^{(1)}_m) \quad m = 1, \cdots, m
\end{equation}

where~$b^{(1)}_m$ is the bias, and the number of zeros we want to pad the input with is such that: \text{dim}(I)~=~\text{dim}(\text{decode}(\text{encode}(I))). Finally, the encoding convolution is equal to:
\begin{equation}
\begin{split}
O_w = O_h &= (I_w + 2(2k +1) -2) - (2k + 1) + 1 \\&= I_w + (2k + 1) - 1
\end{split}
\end{equation}

The decoding convolution step produces~$n$ feature maps~$z_{m=1,\hdots,n}$. The reconstructed results~$\hat{I}$ is the result of the convolution between the volume of feature maps~$Z=\{z_{i=1}\}^n$ and this convolutional filters volume~$F^{(2)}$~\cite{chen2015page,geng2015high,masci2011stacked}.
\begin{equation}
\tilde{I} = a(Z * F^{(2)}_{m} + b^{(2)})
\end{equation}
\begin{equation}\label{eq:a:CNN1}
\begin{split}
O_w = O_h = &( I_w + (2k + 1) - 1 ) -\\&  (2k + 1) + 1 = I_w = I_h
\end{split}
\end{equation}
where Equation~\eqref{eq:a:CNN1} shows the decoding convolution with~$I$ dimensions. Input's dimensions are equal to the output's dimensions.

\subsection{Recurrent Autoencoder Architecture}
A recurrent neural network (RNN) is a natural generalization of feedforward neural networks to feature sequences~\cite{sutskever2014sequence}.  A standard RNN compute the encoding as a feature sequences of output by iteration:
\begin{align}
h_t &= sigm(W^{hx}x_t +W^{hh}h_{t-1})\\
y_t &= W^{y^h}h_t
\end{align}
where x is inputs $ (x_1, . . . , x_T )$ and $y$ refers to output~($y_1,...,y_T$). A multinomial distribution (1-of-K coding) can be output using a softmax activation function~\cite{cho2014learning}:
\begin{equation}
p(x_{t,j}=1\given x_{t-1,...,x_1}) = \frac{\exp(w_jh_t)}{\sum_{j'=1}^K \exp(w_{j'}h_t)}
\end{equation}

By combining these probabilities, we can compute the probability of the sequence~$x$ as:
\begin{equation}
p(x) = \prod_{t=1}^T p(x_t \given x_{t-1},...,x_1)
\end{equation}

\section{Medical Image Staining Problem for Machine Learning Algorithms}\label{sec:staining}

In this section, we explain the medical image staining problem for machine learning algorithms which include medical image staining; then, we discuss staining problems for machine learning, and finally, we explain possible solutions including color normalization and color balancing. 

\subsection{Medical Image Staining}
Hematoxylin and eosin (H\&E) stains have been used for at least a century and are still essential for recognizing various tissue types and the morphologic changes that form the basis of contemporary CD, EE, and cancer diagnosis~\cite{fischer2008hematoxylin}.  H\&E is used routinely in histopathology laboratories as it provides the pathologist/researcher a very detailed view of the tissue\cite{anderson2011introduction}.

\begin{figure}
    \centering
     \includegraphics[width=\textwidth]{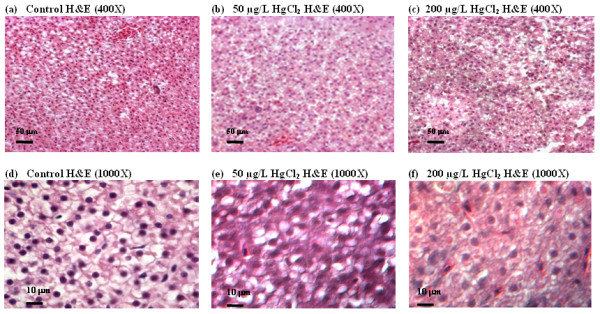}
    \caption{Hematoxylin and eosin (H\&E) staining ith 0 (control, a and b), 50 mg/L (b and e) and 200 mg/L HgCl2 (c and f). Panels (a-c) are shown in low magnification (400×) while (d-f) in high magnification (1000×).~\cite{ung2010mercury}.}
    \label{fig:h_and_e}
\end{figure}

\subsection{Staining Problem for Machine Learning Algorithm}
\begin{figure}[!b]
    \centering
     \includegraphics[width=\textwidth]{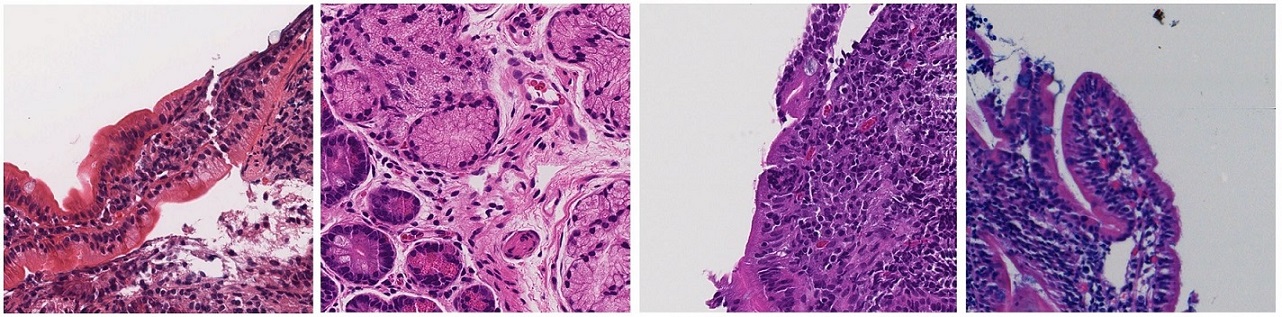}
    \caption{Different H\&E Color staining }
    \label{fig:d_h_e}
\end{figure}

Color variation has been a very important problem in histopathology based on light microscopy. A range of factors makes this problem even more complex such as the use of different scanners, variable chemical coloring/reactivity from different manufacturers/batches of stains, coloring being dependent on staining procedure (timing, concentrations, etc.), and light transmission being a function of section thickness~\cite{khan2014nonlinear}. This different H\&E staining standard~(as shown in Figure~\ref{fig:d_h_e}) can mislead the machine learning algorithms by color; for example, if we have all~$"A"$ images from center $One$ and all labeled $"B"$ images from center $"Two"$ and they used different H\&E standard which makes these based color slightly different with each other; thus, the training machine learning model based on the original images (without any pre-processing) misled by the color of these images instead of characteristic of images such as Crypt Epithelium, Goblet cells, Inflammatory cells, etc.

\subsection{Color Normalization and Color Balancing}
 In color normalization, Images are first deconvolved into their principal staining constituents then normalization is performed on each staining channel separately, which involves a mapping between reference and
source image~\cite{bentaieb2017adversarial}. However, this technique is limited to one standard color staining from one center which means non-linear machine learning algorithms such as deep learning could detect these color staining normalization. Many researchers try to solve this problem by modifying these algorithm~\cite{piorkowski2018color,sano2019fast,zarella2017alternative,roberto2019classification,yuan2018neural,ziaei2020characterization}. 

The another problem in this part is the computational complexity of these models which the original color normalization could not remove all intensity values from the image while preserving color values, so it needs to be a more complex model which increase computation time in these algorithms. Although many researchers proposed different color normalization techniques, but still these algorithms cannot remove all intensity values from the images.

The other way to solve this problem is the charging classification model instead of pre-processing. Many researchers using color balancing techniques to remove the effectiveness of color in the machine learning model~\cite{bianco2017improving,kowsari2019diagnosis} In this approach, the model is trained with different range of color balanced images.

\section{Evaluation Techniques}\label{sec:Evaluation}
In the research community, having shared and comparable performance measures to evaluate algorithms is preferable. However, in reality such measures may only exist for a handful of methods. The major problem when evaluating image classification methods is the absence of standard data collection protocols. Even if a common collection method existed (\textit{e.g.} Reuters news corpus), simply choosing different training and test sets can introduce inconsistencies in model performance~\cite{yang1999evaluation}. Another challenge with respect to method evaluation is being able to compare different performance measures used in separate experiments. Performance measures generally evaluate specific aspects of classification task performance, and thus do not always present identical information. In this section, we discuss evaluation metrics and performance measures and highlight ways in which the performance of classifiers can be compared.

\begin{figure}[!b]
\centering
    \includegraphics[width=0.8\textwidth]{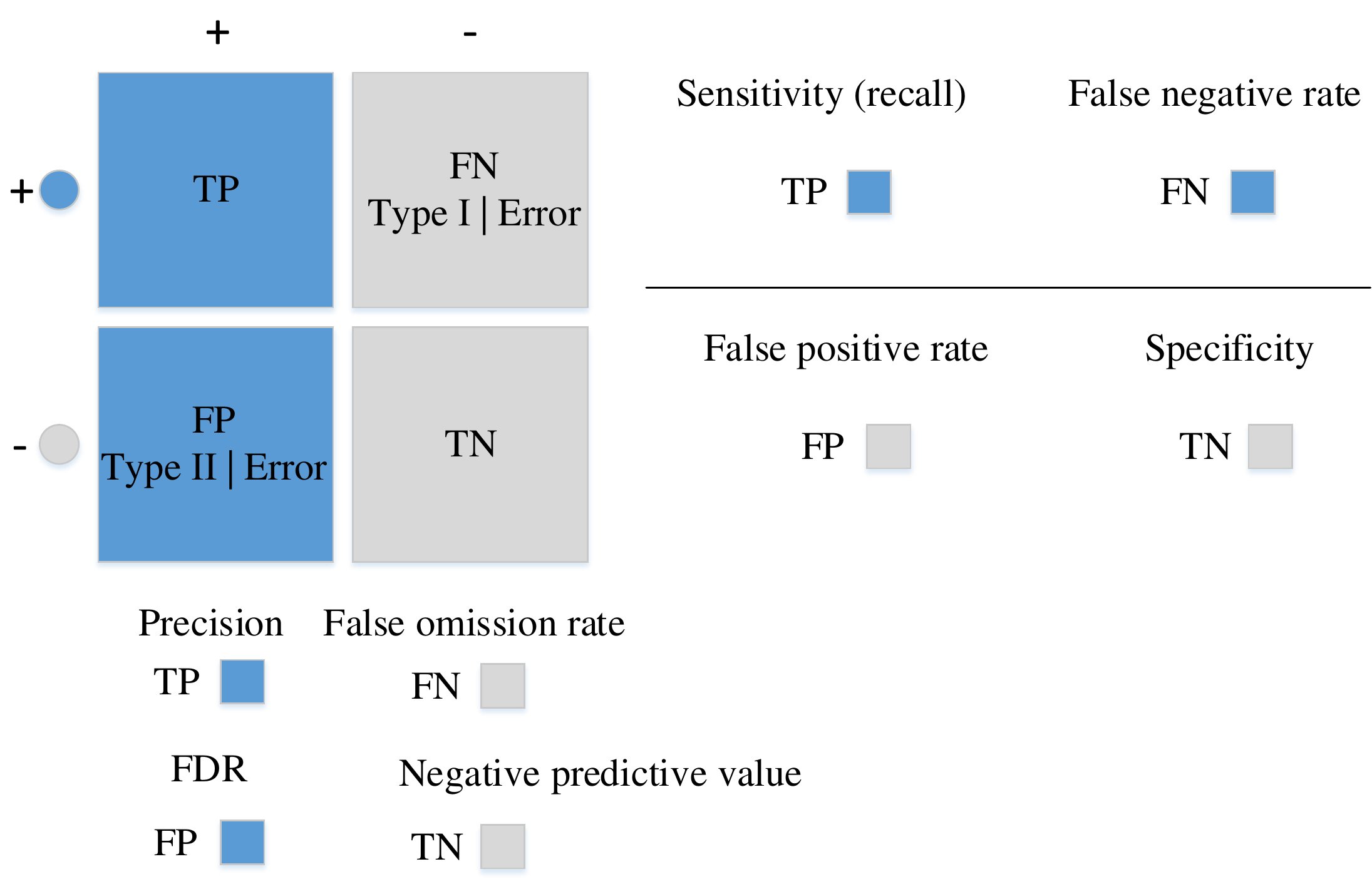}
    \caption{Confusion matrix}  \label{fig:F1}
\end{figure}

Since the underlying mechanics of different evaluation metrics may vary, understanding what exactly each of these metrics represents and what kind of information they are trying to convey is crucial for comparability. Some examples of these metrics include recall, precision, accuracy, F-measure, micro-average, and macro average. These metrics are based on a~``confusion matrix''~(shown in Figure~\ref{fig:F1}) that comprises true positives~(TP), false positives~(FP), false negatives~(FN) and true negatives~(TN)~\cite{lever2016points}. The significance of these four elements may vary based on the classification application. The fraction of correct predictions over all predictions is called accuracy~(Eq. \ref{eq:acc}). The fraction of known positives that are correctly predicted is called sensitivity~\textit{i.e.} true positive rate or recall~(Eq. \ref{eq:recall}). The ratio of correctly predicted negatives is called specificity~(Eq. \ref{eq:spec}). The proportion of correctly predicted positives to all positives is called precision,~\textit{i.e.} positive predictive value (Eq. \ref{eq:pres}). 

\begin{align}
accuracy&=\frac{(TP+TN)}{(TP+FP+FN+TN)}\label{eq:acc}\\
sensitivity&=\frac{TP}{(TP+FN)}\label{eq:recall}\\
specificity&=\frac{TN}{(TN+FP)}\label{eq:spec}\\
precision &= \frac{\sum_{l=1}^LTP_l}{\sum_{l=1}^LTP_l+FP_l}\label{eq:pres}\\
recall&= \frac{\sum_{l=1}^LTP_l}{\sum_{l=1}^LTP_l+FN_l}\\
F1-Score &=  \frac{\sum_{l=1}^L2TP_l}{\sum_{l=1}^L2TP_l+FP_l+FN_l}
\end{align}

The pitfalls of each of the above-mentioned metrics are listed in Table~\ref{table:pitfall}.

\begin{table}[h]
\centering
\caption{Metrics Pitfalls}
\label{table:pitfall}
\begin{tabular}{>{\centering\arraybackslash}m{3cm} >{\raggedright\arraybackslash}m{10cm}}
\hline
            & \multicolumn{1}{c}{Limitation}                                                                                                \\ \hline
Accuracy    & gives us no information on FN and FP                                                                                        \\ \hline
Sensitivity & does not evaluate TN and FP and any classifier that predicts data points as positives considered to have high sensitivity   \\ \hline
Specificity & similar to sensitivity and does not account for FN and TP                                                                   \\ \hline
Precision   & does not evaluate TN and FN and considered to be very conservative and goes for a case which is most certain to be positive \\ \hline
\end{tabular}
\end{table}

\subsection{Macro-Averaging and Micro-Averaging}
A single aggregate measure is required when several two-class classifiers are being used to process a collection. Macro-averaging gives a simple average over classes while micro-averaging combines per-datapoint decisions across classes and then outputs an effective measure on the pooled contingency table~\cite{manning2008matrix}. Macro-averaged results can be computed as follows:

\begin{equation}
    B_{macro} = \frac{1}{q}\sum_{\lambda=1}^q B(TP_{\lambda}+FP_{\lambda}+TN_{\lambda}+FN_{\lambda})
\end{equation}
where $B$ is a binary evaluation measure calculated based on true positives~(TP), false positives~(FP), false negatives~(FN) and true negatives~(TN), and $L=\{\lambda_j : j=1...q\}$ is the set of all labels. 

Micro-averaged results~\cite{sebastiani2002machine,tsoumakas2009mining} can be computed as follows:

\begin{equation}
\begin{split}
         B_{macro} = B\bigg(&\sum_{\lambda=1}^q TP_{\lambda}, \sum_{\lambda=1}^q FP_{\lambda},\sum_{\lambda=1}^q TN_{\lambda}, \sum_{\lambda=1}^q FN_{\lambda} \bigg)
\end{split}
\end{equation}
Micro-average score assigns equal weights to every images and as a consequence, and it is considered to be a per-image average. On the other hand, macro-average score assigns equal weights to each category without accounting for frequency and therefore it is a per-category average. 

\subsection{F$_{\beta}$ Score}
$F_{\beta}$ is one of the most popular aggregated evaluation metrics for classifier evaluation~\cite{lever2016points}. The parameter $\beta$ is used to balance recall and precision and defined as follows:

\begin{equation}{\label{eq:fbeta}}
F_{\beta} = \frac{(1+\beta^2)(precision \times recall)}{\beta^2 \times precision+recall}    
\end{equation}
For commonly used $\beta=1$~\textit{i.e.} $F_1$, recall and precision are given equal weights and Eq. \ref{eq:fbeta} can be simplified to:

\begin{equation}{\label{eq:f1}}
 F_1=\frac{2TP}{2TP+FP+FN}   
\end{equation}
Since $F_\beta$ is based on recall and precision, it does not represent the confusion matrix fully. 

\subsection{Matthews Correlation Coefficient (MCC)}
The Matthews correlation coefficient~(MCC)~\cite{matthews1975comparison} captures all the data in a confusion matrix and measures the quality of binary classification methods. MCC can be used for problems with uneven class sizes and is still considered a balanced measure. MCC ranges from $-1$ to $0$ (i.e. the classification is always wrong and always true respectively). MCC can be calculated as follows:

\begin{equation}{\label{eq:mcc}}
    MCC=\frac{TP \times TN - FP \times FN}{\sqrt{\splitdfrac{(TP+FP)\times(TP+FN)\times}{(TN+FP)\times(TN+FN)}}}
\end{equation}
While comparing two classifiers, one may have a higher score using MCC and the other one has a higher score using $F_1$ and as a result one specific metric cannot captures all the strengths and weaknesses of a classifier~\cite{lever2016points}. 

\subsection{Receiver Operating Characteristics (ROC)}
Receiver operating characteristics~(ROC)~\cite{yonelinas2007receiver} curves are valuable graphical tools for evaluating classifiers. However, class imbalances (i.e. differences in prior class probabilities~\cite{japkowicz2002class}) can cause ROC curves to poorly represent the classifier performance. ROC curve plots true positive rate~(TPR) and false positive rate~(FPR):

\begin{equation}
    TPR = \frac{TP}{TP + FN}
\end{equation}
\begin{equation}
    FPR = \frac{FP}{FP + TN}
\end{equation}

\subsection{Area Under ROC Curve (AUC)}
The area under ROC curve~(AUC)~\cite{hanley1982meaning,pencina2008evaluating} measures the entire area underneath the ROC curve. AUC leverages helpful properties such as increased sensitivity in the analysis of variance~(ANOVA) tests, independence from decision threshold, invariance to \textit{a priori} class probabilities, and indication of how well negative and positive classes are regarding decision index~\cite{bradley1997use}.\\

For binary classification tasks, AUC can be formulated as:

\begin{equation}
\begin{split}
        AUC &= \int_{-\infty}^{\infty} TPR(T) FPR'(T) dT\\ 
    &= \int_{-\infty}^{\infty}\int_{-\infty}^{\infty} I(T'>T)f_1(T')f_0(T)dT dT'\\ 
    &= P(X_1>X_0)
\end{split}
\end{equation}

For multi-class AUC, an average AUC can be defined~\cite{hand2001simple} as follows:

\begin{equation}
    AUC = \frac{2}{|C|(|C|-1)} \sum_{i=1}^{|C|} AUC_i
\end{equation}
where C is the number of the classes.\\
Yang~\cite{yang1999evaluation} evaluated statistical approaches for classification which following important factors that should be considered when comparing classifier algorithms:

\begin{itemize}
    \item Comparative evaluation across methods and experiments which gives insight about factors underlying performance variations and will lead to better evaluation methodology in the future
    \item Impact of collection variability such as including unlabelled images in training or test set and treat them as negative instances can be a serious problem.
    \item Category ranking evaluation and binary classification evaluation show the usefulness of classifier in interactive applications and emphasize their use in a batch mode respectively. Having both types of performance measurements to rank classifiers is helpful in detecting the effects of thresholding strategies.
    \item Evaluation of the scalability of classifiers in large category spaces is rarely investigated area.  
\end{itemize}

\section{Summary}\label{sec:Literature.Review.Summary}
The classification task is one of the most indispensable problems in machine learning. As~Medical image and image data sets proliferate, the development and documentation of supervised machine learning algorithms becomes an imperative issue, especially for medical image classification. Having a better categorization system for these images require discerning these algorithms. However, the existing image classification algorithms work more efficiently if we have a better understanding of pre-processing methods and how to evaluate them correctly. Existing classification algorithms, such as the Rocchio algorithm, bagging and boosting, logistic regression~(LR), k-nearest Neighbor~(KNN), Support Vector Machine~(SVM), random forest, and deep learning, are the primary focus of this part. Evaluation methods, such as accuracy, $F_{\beta}$, Matthew correlation coefficient~(MCC), receiver  operating  characteristics~(ROC), and area  under curve~(AUC), are explained. With these metrics, the image classification algorithm can be evaluated.
\chapter{Diagnosis of Celiac Disease and Environmental Enteropathy on Biopsy Images Using Color Balancing on Convolutional Neural Networks}\label{chpt:Celiac_color}
Celiac Disease~(CD) and Environmental Enteropathy~(EE) are common causes of malnutrition and adversely impact normal childhood development. CD is an autoimmune disorder that is prevalent worldwide and is caused by an increased sensitivity to gluten. Gluten exposure destructs the small intestinal epithelial barrier, resulting in nutrient mal-absorption and childhood under-nutrition. EE also results in barrier dysfunction but is thought to be caused by an increased vulnerability to infections. EE has been implicated as the predominant cause of under-nutrition, oral vaccine failure, and impaired cognitive development in low-and-middle-income countries. Both conditions require a tissue biopsy for diagnosis, and a major challenge of interpreting clinical biopsy images to differentiate between these gastrointestinal diseases is striking histopathologic overlap between them. In the current
study, we propose a convolutional neural network~(CNN) to classify duodenal biopsy images from subjects with CD, EE, and healthy controls. We evaluated the performance of our proposed model using a large cohort containing 1000 biopsy images. Our evaluations show that the proposed model achieves an area under ROC of 0.99, 1.00, and 0.97 for CD, EE, and healthy controls, respectively. These results demonstrate the discriminative power of the proposed model in duodenal biopsies classification.

\section{Introduction and Related Works}

In this section, we propose a CNN-based model for the classification of biopsy images. In recent years, Deep Learning architectures have received great attention after achieving state-of-the-art results in a wide variety of fundamental tasks such classification~\cite{Heidarysafa2018RMDL,kowsari2017hdltex,kowsari2018rmdl,info10040150,litjens2017survey,nobles2018identification,zhai2016doubly} or other medical domains~\cite{hegde2019comparison,zhang2018patient2vec}. CNNs, in particular, have proven to be very effective in medical image processing. CNNs preserve local image relations while reducing dimensionality and for this reason are the most popular machine learning algorithm in image recognition and visual learning tasks~\cite{ker2018deep}. CNNs have been widely used for classification and segmentation in various types of medical applications such as histopathological
images of breast tissues, lung images, MRI images,  medical X-Ray images, etc.~\cite{gulshan2016development,litjens2017survey}. Researchers produced advanced results on duodenal biopsies classification using CNNs~\cite{Mohammad_al_boni}, but those models are only robust to a single type of image stain or color distribution. Many researchers apply a stain normalization technique as part of the image pre-processing stage to both the training and validation datasets~\cite{nawaz2018classification,syed2019mo1992,shrivastava2019self}. In this dissertation section, varying levels of color balancing were applied during image pre-processing in order to account for multiple stain variations. 
\begin{figure}[h]
    \centering
    \includegraphics[width=\columnwidth]{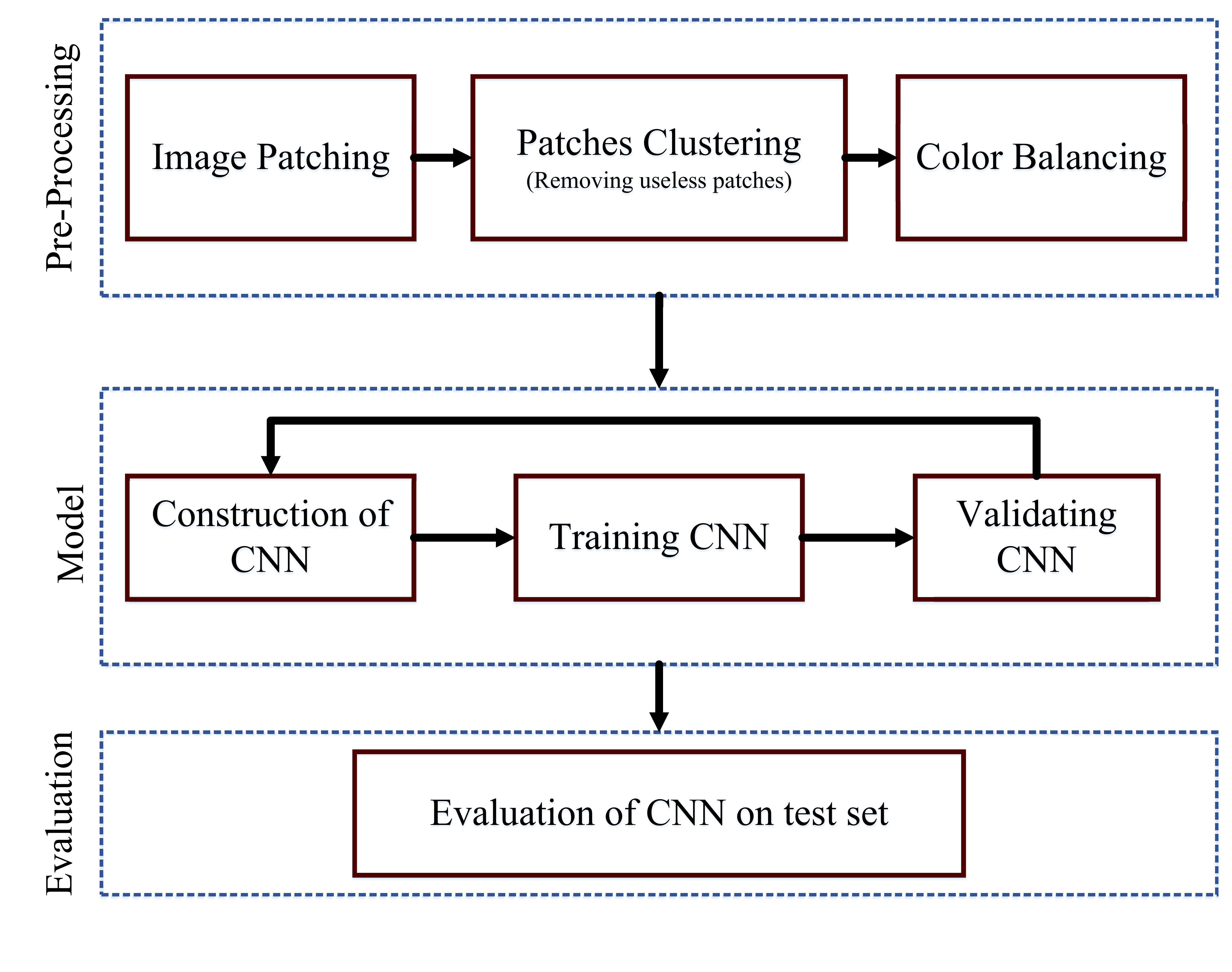}
    \caption{Overview of methodology} \label{fig:Pipeline}
\end{figure}

The rest of this Section is organized as follows: In Section~\ref{sec:Data_Source_3}, we describe the different data sets used in this work, as well as, the required pre-processing steps. The architecture of the model is explained in Section~\ref{sec:Method_3}. Empirical results are elaborated in Section~\ref{sec:Empirical_Results_3}. Finally, Section~\ref{sec:Conclusion_3} concludes the dissertation section along with outlining future directions.

\section{Data Source}\label{sec:Data_Source_3}

For this project, $121$ Hematoxylin and Eosin (H\&E) stained duodenal biopsy glass slides were retrieved from~$102$ patients. The slides were converted into~$3118$ whole slide images and labeled as either EE, CD, or normal. The biopsy slides for EE patients were from the Aga Khan University Hospital~(AKUH) in Karachi, Pakistan~($n = 29$ slides from~$10$ patients) and the University of Zambia Medical Center in Lusaka, Zambia ($n = 16$). The slides for CD patients ($n = 34$) and normal ($n = 42$) were retrieved from archives at the University of Virginia~(UVa). The CD and normal slides were converted into whole slide images at~$40$x magnification using the Leica SCN~$400$ slide scanner (Meyer Instruments, Houston, TX) at UVa, and the digitized EE slides were of 20x magnification and shared via the Environmental Enteric Dysfunction Biopsy Investigators~(EEDBI) Consortium shared WUPAX server. Characteristics of our patient population are as follows: the median~($Q1$, $Q3$) age of our entire study population was~$31$~($20.25$, $75.5$) months, and we had a roughly equal distribution of males~($52$\%, $n = 53$) and females~($48$\%, $n = 49$). The majority of our study population were histologically normal controls~$(41.2\%)$, followed by CD patients~$(33.3\%)$, and EE patients~$(25.5\%)$.

\section{Pre-Processing}\label{sec:Pre-Processing_3}
In this section, we cover all of the pre-processing steps which include image patching, image clustering, and color balancing. The biopsy images are unstructured~(varying image sizes) and too large to process with deep neural networks; thus, requiring that images are split into multiple smaller images. After executing the split, some of the images do not contain much useful information. For instance, some only contain the mostly blank border region of the original image. In the image clustering section, the process to select useful images is described. Finally, color balancing is used to correct for varying color stains which is a common issue in histological image processing.

\subsection{Image Patching}\label{subsec:lower:Patching}
Although the effectiveness of CNNs in image classification has been shown in various studies in different domains, training on high-resolution Whole Slide Tissue Images (WSI) is not commonly preferred due to a high computational cost. However, applying CNNs on WSI enables losing a large amount of discriminative information due to extensive downsampling~\cite{hou2016patch}. Due to a cellular level difference between Celiac, Environmental Enteropathy and normal cases, a trained classifier on image patches are likely to perform as well as or even better than a trained WSI-level classifier. Many researchers in pathology image analysis have considered classification or feature extraction on image patches~\cite{hou2016patch}.
In this project, after generating patches from each image, labels were applied to each patch according to its associated original image. CNN was trained to generate predictions on each individual patch.

\begin{figure}[!b]
    \centering
    \includegraphics[width= \textwidth]{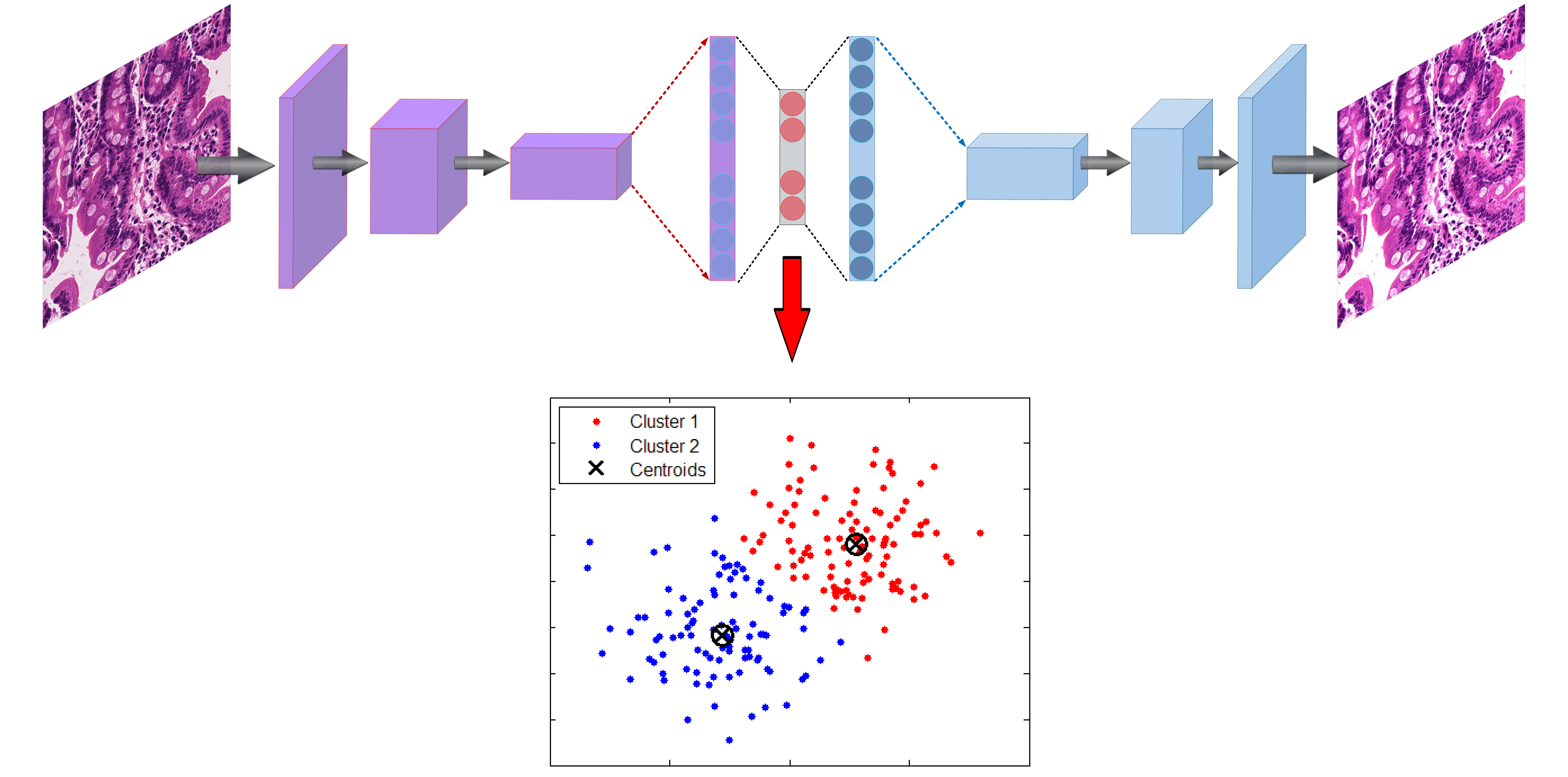}
    \caption{Structure of clustering model with autoencoder and K-means combination} \label{fig:AE}
\end{figure}

\subsection{Clustering}\label{subsec:lower:Clustering}
\subsubsection{Autoencoder}
In this study, after image patching, some of the created patches do not contain any useful information regarding biopsies and should be removed from the data. These patches have been created from mostly background parts of WSIs. A two-step clustering process was applied to identify the unimportant patches. For the first step, a convolutional autoencoder was used to learn embedded features of each patch and in the second step, we used k-means to cluster embedded features into two clusters: useful and not useful. In Figure~\ref{fig:AE}, the pipeline of our clustering technique is shown which contains both the autoencoder and k-mean clustering.  

An autoencoder is a type of neural network that is designed to match the model's inputs to the outputs~\cite{goodfellow2016deep}. The autoencoder has achieved great success as a dimensionality reduction method via the powerful reprehensibility of neural networks~\cite{wang2014generalized}. The first version of the autoencoder was introduced by~\textit{DE. Rumelhart el at.}~\cite{rumelhart1985learning} in 1985. The main idea is that one hidden layer between input and output layers has much fewer units~\cite{liang2017text} and can be used to reduce the dimensions of feature space. For medical images which typically contain many features, using an autoencoder can help allow for faster, more efficient data processing.

A CNN-based autoencoder can be divided into two main steps~\cite{masci2011stacked}~: encoding and decoding.
\begin{equation}
\begin{split}
    O_m(i, j) = a\bigg(&\sum_{d=1}^{D}\sum_{u=-2k-1}^{2k+1}\sum_{v=-2k -1}^{2k +1}F^{(1)}_{m_d}(u, v)I_d(i -u, j -v)\bigg) \\&\quad m = 1, \cdots, n
\end{split}
\end{equation}
Where~$F \in \{F^{(1)}_{1},F^{(1)}_{2},\hdots,F^{(1)}_{n},\}$ is a convolutional filter, with convolution among an input volume defined by~$I = \left\{I_1,\cdots, I_D\right\}$ which it learns to represent the input by combining non-linear functions:

\begin{equation}
    z_m = O_m = a(I * F^{(1)}_{m} + b^{(1)}_m) \quad m = 1, \cdots, m
\end{equation}
where~$b^{(1)}_m$ is the bias, and the number of zeros we want to pad the input with is such that: \text{dim}(I) = \text{dim}(\text{decode}(\text{encode}(I))) Finally, the encoding convolution is equal to:
\begin{equation}
\begin{split}
     O_w = O_h &= (I_w + 2(2k +1) -2) - (2k + 1) + 1 \\&= I_w + (2k + 1) - 1
\end{split}
\end{equation}
The decoding convolution step produces~$n$ feature maps~$z_{m=1,\hdots,n}$. The reconstructed results~$\hat{I}$ is the result of the convolution between the volume of feature maps~$Z=\{z_{i=1}\}^n$ and this convolutional filters volume~$F^{(2)}$~\cite{chen2015page,geng2015high}.
\begin{equation}
    \tilde{I} = a(Z * F^{(2)}_{m} + b^{(2)})
\end{equation}
\begin{equation}\label{eq:a:CNN}
\begin{split}
      O_w = O_h = ( I_w + (2k + 1) - 1 ) -  (2k + 1) + 1 = I_w = I_h  
\end{split}
\end{equation}
Where Equation~\ref{eq:a:CNN} shows the decoding convolution with ~$I$ dimensions. The input's dimensions are equal to the output's dimensions.

\subsubsection{K-Means}
\label{sec:kmeans3}
K-means clustering is one of the easiest methods of clustering algorithm~\cite{jain2010data,kowsari2014investigation} which data is given  $D\in\{x_1,x_2,...,x_n\}$ in $d$ dimensional vectors which is $x\in f^d$. The aim is to identify groups of data points and assign each point to one of the groups. What is good clustering? There is no universal approach, but we will try the K-means approach, but it has a lot of limitations that we could not calculate the error rate and error percentage although the loss function is available~\cite{kowsari2016weighted}. 

If we want to find the error rate the time complexity is equal to exponential. One measure of how good clustering is would be the sum of distances to the center. Therefore, K-means wants to minimize this $\xi$~(quantity), choosing $\mu$ and a’s to minimize but it is difficult to do analytically because a’s are binary assignments. So the K-means algorithm tries to iteratively solve the minimization~(sort of greedy algorithm)~\cite{alassaf2015automatic,yammahi2014efficient}.

Minimize~$\xi$ with respect to~$a$ and~$\mu$ by:
\begin{align}
    \xi = \sum_{j=1}^k \sum_{x_i} ||x_i-\mu_j||^2 = \sum_{j=1}^k \sum_{i=1}^n A_{ij}||x_i-\mu_j||
\end{align}
where $\xi$ stand for quantity of dataset, and $\mu$ is centroid of each cluster    
\begin{enumerate}
\item Initialize $\mu_1$ to $\mu_k$ arbitrarily.
\item 	Choose the optimal assignment “a” for given centers $\mu$. [Fix µ optimize a].
\item 	Choose optimal µ for fixed “a” [Fix a optimize $\mu$].
\item 	Repeat~(iteratively) 2 and 3 until convergence.
\end{enumerate}

\begin{algorithm}[!t]
\SetKwFor{Foreach}{for}{do}{endfor}
\SetKwFor{WHILE}{while}{do}{endwhile}
\caption{K-means $n$ images for $K$ Clusters (in our expriment k=2)}

\textbf{Input: }$D= \{\overrightarrow{x_1},\overrightarrow{x_2},...,\overrightarrow{x_n}\}$

\textbf{Output: }$\mu = \overrightarrow{\mu_1},\overrightarrow{\mu_2},...,\overrightarrow{\mu_k}\} $

$\{\overrightarrow{s_1}, \overrightarrow{s_2},...,\overrightarrow{s_k}\}$ set random seeds\\ $(\{\overrightarrow{x_1},\overrightarrow{x_2},...,\overrightarrow{x_n}\},K)$

    \Foreach {$k \leftarrow 1$ to K}{
    
$\overrightarrow{\mu_k} \leftarrow \overrightarrow{s_k}$
    
    }
    \WHILE{Criterion has not been met}
    {
      \Foreach {$k \leftarrow 1$ to K}{
      
      $w_k \leftarrow \{\}$
      }
      
    \Foreach {$n \leftarrow 1$ to $N$}{
      
      $j \leftarrow arg \min_{j'} |\overrightarrow{\mu_{j'}} - \overrightarrow{\mu_{x_n}}|$

      $w_j \leftarrow w_j \bigcup ~\{\overrightarrow{x_n}\}$

     }
    \Foreach {$k \leftarrow 1$ to K}{
      
      $\mu_k \leftarrow \frac{1}{|w_k|} \sum_{\overrightarrow{x}\in w_k} \overrightarrow{x}$
      
      }
    }
\end{algorithm}

K-Means is not guaranteed to converge to the global minimum of this function. In fact, finding a global minimum is a NP-hard problem, but it is reasonable to some extent.
\begin{enumerate}
\item Length of vectors is M (point and a centroid) so computing the distance between two points is O (M).
\item Reassign clusters: k = number of clusters, and N is number of points = $O(KN)$. So computing distance = $O(MKN)$
\item Computing centroids:  Where each point assigned once to a certain centroid = $O(NM)$ in that cluster k.
\item Assuming reassigning of clusters and computing centroids each done at least once for I iterations = $O(IKNM)$.
\end{enumerate}

So the time complexity of K-means is equal to if K value which the number of clusters and I value which the number of the iteration will be available; although, if the number of the iteration value is not available, time complexity can be exponential time. K-means also is used for image and data clustering especially broadly used in information retrieval~\cite{jain2010data,manning20introduction}. data-points (image representation) have the same cluster behave similarly
with respect to relevance to datapoint's information which is extracted as feature space. The centroid $\mu$ of a datapoint is calculated as follows:
\begin{equation}
    \mu (w) = \frac{1}{|w|} \sum_{\bar{x}\in w} \bar{x}
\end{equation}

 Results of patch clustering have been summarized in Table~\ref{tb:clustering}. Obviously, patches in cluster~$1$, which were deemed useful, are used for the analysis in this section.

\begin{figure}[!b]
    \centering
    \includegraphics[width=\columnwidth]{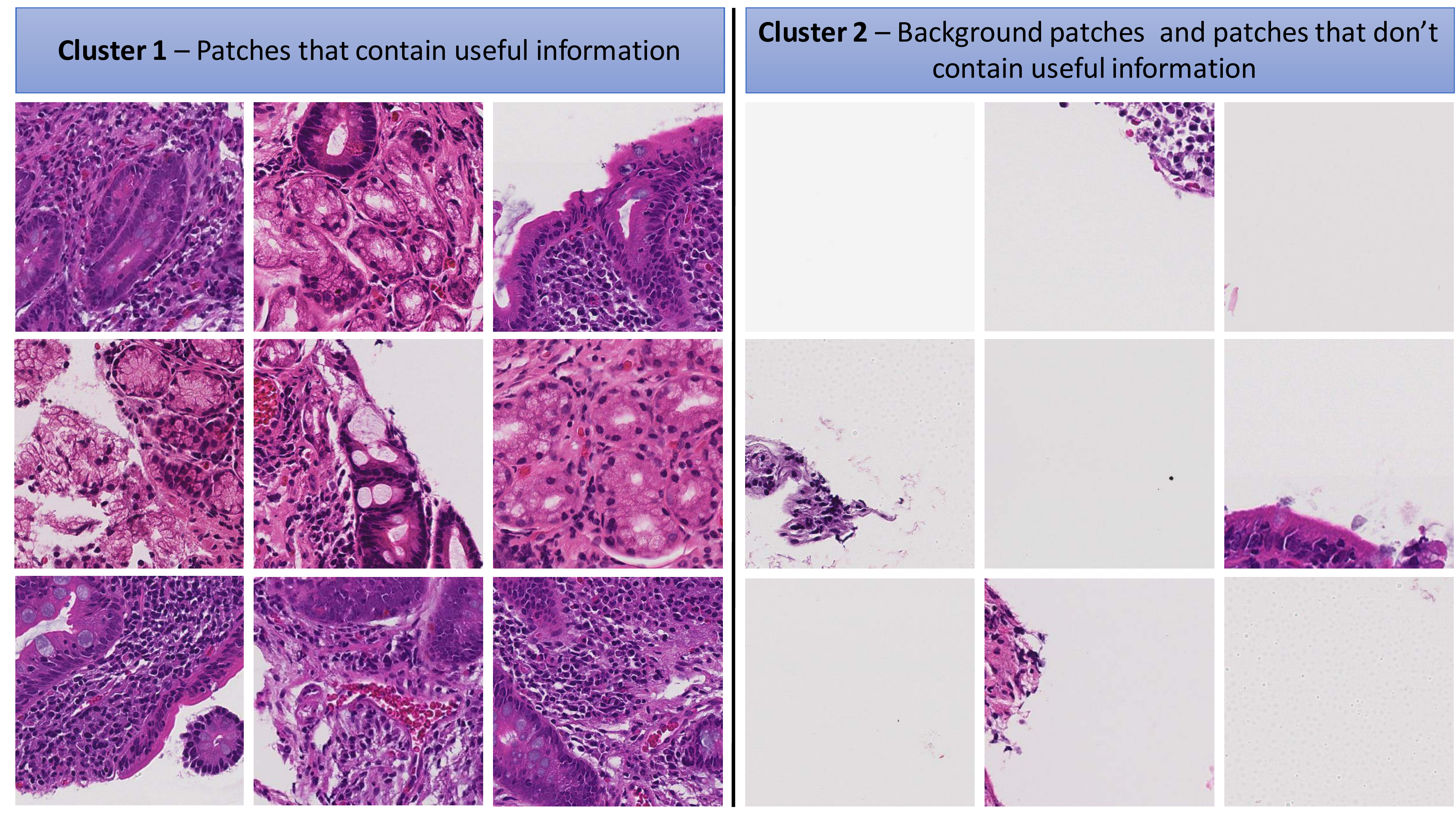}
    \caption{Some samples of clustering results - cluster 1 includes patches with useful information and cluster 2 includes patches without useful information (mostly created from background parts of WSIs)} \label{fig_Clustering}
\end{figure}

\begin{table}[!t]
\caption{The clustering results for all patches into two clusters}\label{tb:clustering}
\centering
\begin{tabular}{|c|c|c|c|}
\hline
                                                                         & Total & Cluster 1   & Cluster 2    \\ \hline
Celiac Disease (CD)                                                      & $16,832$  & $7,742~(46\%)$ & $9,090~(54\%)$  \\ \hline
Normal                                                                   & $15,983$  & $8,953~(56\%)$ & $7,030~(44\%)$  \\ \hline
Environmental Enteropathy~(EE)& $22,625$ & $2,034~(9\%)$   & $20,591~(91\%)$ \\ \hline
Total                                                                    & $55,440$ & $18,729~(34\%)$ & $36,711~(66\%)$ \\ \hline
\end{tabular}
\end{table}

\begin{figure}[!b]
    \centering
    \includegraphics[width=\columnwidth]{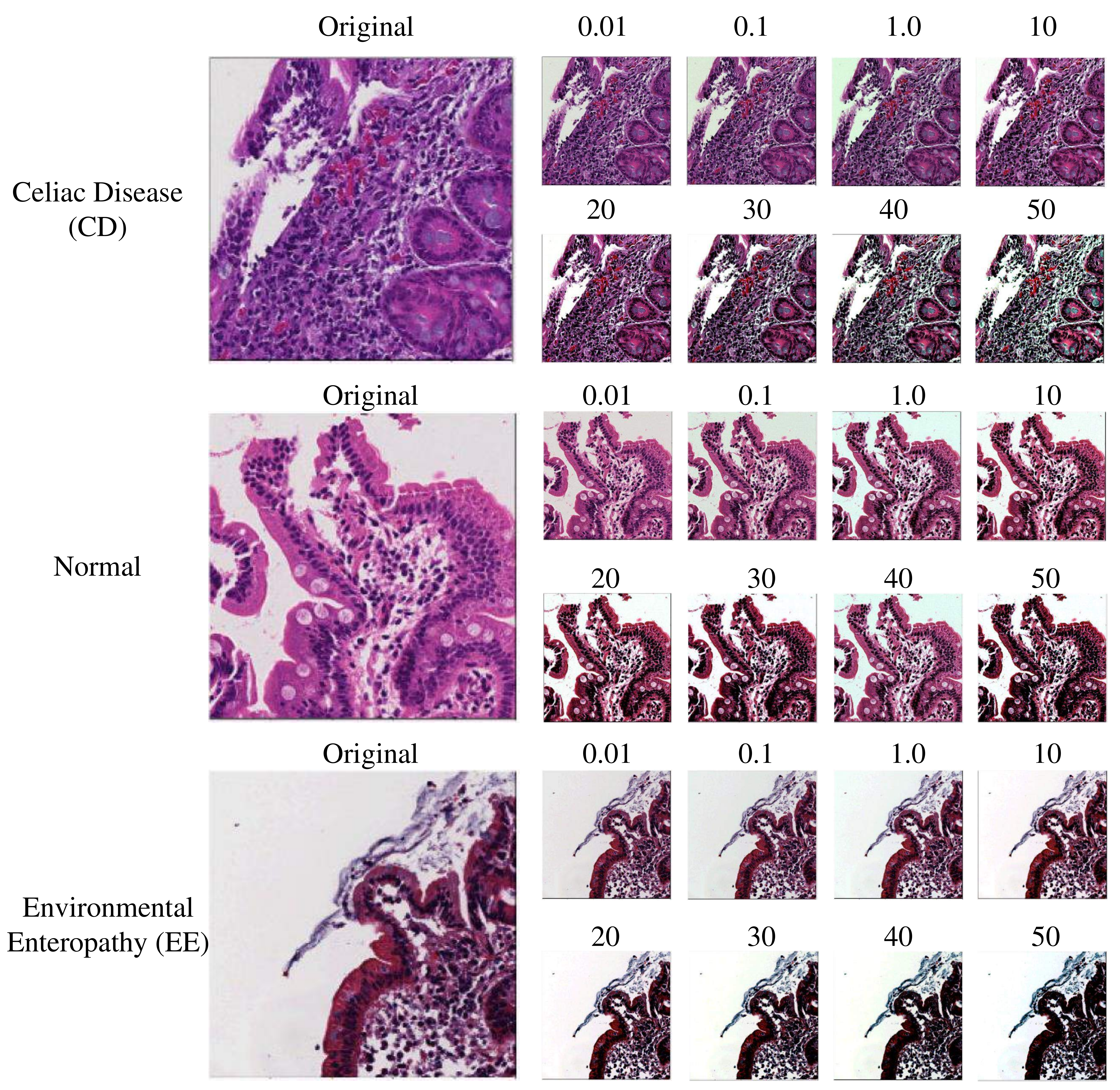}
    \caption{Color Balancing samples for the three classes} \label{fig:CB1}
\end{figure}

\subsection{Color Balancing}\label{subsec:CB_3}
The concept of color balancing for this dissertation section  is to convert all images to the same color space to account for variations in H\&E staining. The images can be represented with the illuminant spectral power distribution~$I(\lambda)$, the surface spectral reflectance~$S(\lambda)$, and the sensor spectral sensitivities~$C(\lambda)$~\cite{bianco2017improving,bianco2014error}. Using this notation~\cite{bianco2014error}, the sensor responses at the pixel with coordinates~$(x,y)$ can be thus described as:

\begin{equation}
    p(x,y) = \int_w I(x,y,\lambda) S(x,y,\lambda) C(\lambda) d\lambda
\end{equation}
where~$w$ is the wavelength range of the visible light spectrum, p and~$C(\lambda)$ are three-component vectors.

\begin{equation}\label{eq_RGB_IN_OUT}
    \begin{aligned} \left [ \begin{array}{c} R \\ G \\ B \\ \end{array} \right ]_{out} =& \left( \alpha \left [ \begin{array}{c@{\quad}c@{\quad}c} a_{11} & a_{12} & a_{13} \\ a_{21} & a_{22} & a_{23} \\ a_{31} & a_{32} & a_{33}\\ \end{array} \right ]\right. {}\times\left. \left [ \begin{array}{c@{\quad}c@{\quad}c} r_{awb} & 0 & 0 \\ 0 & g_{awb} & 0 \\ 0 & 0 & b_{awb} \\ \end{array} \right ] \left [ \begin{array}{c} R \\ G \\ B \\ \end{array} \right ]_{in} \right)^{\gamma} \end{aligned}
\end{equation}
where\textbf{~$RGB_{in}$}  is raw images from biopsy and~\textbf{~$RGB_{out}$} is results for CNN input. In the following, a more compact version of Equation~\ref{eq_RGB_IN_OUT} is used:
\begin{equation}
    RGB_{out} = (\alpha AI_w . RGB_{in})^\gamma
\end{equation}

where~$\alpha$ is exposure compensation gain, $I_w$ refers the diagonal matrix for the illuminant compensation and~$A$ indicates the color matrix transformation. 

Figure~\ref{fig:CB1} shows the results of color balancing for three classes~(Celiac Disease (CD), Normal and Environmental Enteropathy (EE)) with different color balancing percentages between~$0.01$ and~$50$.

\section{Method}\label{sec:Method_3}
\begin{figure}[t]
    \centering
    \includegraphics[width=\textwidth]{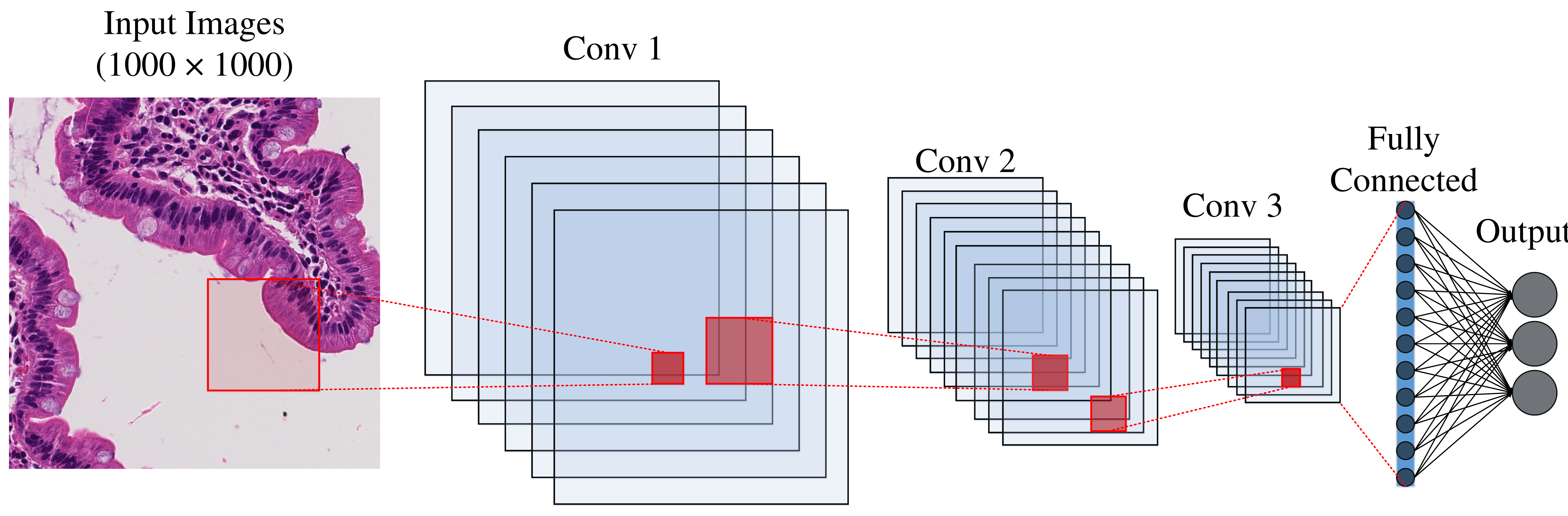}
    \caption{Structure of Convolutional Neural Net using multiple 2D feature detectors and 2D max-pooling} \label{cnn_fig}
\end{figure}

In this section, we describe Convolutional Neural Networks~(CNN) including the convolutional layers, pooling layers, activation functions, and optimizer. Then, we discuss our network architecture for the diagnosis of Celiac Disease and Environmental Enteropathy. As shown in figure~\ref{cnn_fig}, the input layers starts with image patches~($1000\times 1000$) and is connected to the convolutional layer~(\textit{Conv~$1$}). Conv~$1$ is connected to the pooling layer~(\textit{MaxPooling}), and then connected to \textit{Conv~$2$}. Finally, the last convolutional layer~(\textit{Conv~$3$}) is flattened and connected to a fully connected perception layer. The output layer contains three nodes in which each node represents one class.

\subsection{Convolutional Layer}
CNN is a deep learning architecture that can be employed for hierarchical image classification. Originally, CNNs were built for image processing with an architecture similar to the visual cortex. CNNs have been used effectively for medical image processing. In a basic CNN used for image processing, an image tensor is convolved with a set of kernels of size~$d \times d$. These convolution layers are called feature maps and can be stacked to provide multiple filters on the input. The element~(feature) of input and output matrices can be different~\cite{li2014medical}. The process to compute a single output matrix is defined as follows:

\begin{equation}
    A_{j}=f\left(\sum_{i=1}^{N}I_{i}\ast K_{i,j}+B_{j}\right)
\end{equation}
Each input matrix~$I-i$ is convolved with a corresponding kernel matrix~$K_{i,j}$, and summed with a bias value~$B_j$ at each element. Finally, a non-linear activation function~(See Section~\ref{Sec:Activation_3}) is applied to each element~\cite{li2014medical}.

In general, during the back propagation step of a CNN, the weights and biases are adjusted to create effective feature detection filters . The filters in the convolution layer are applied across all three 'channels' or $\Sigma$~(size of the color space)~\cite{Heidarysafa2018RMDL}. 

\subsection{Pooling Layer}
To reduce the computational complexity, CNNs utilize the concept of pooling to reduce the size of the output from one layer to the next in the network. Different pooling techniques are used to reduce outputs while preserving important features ~\cite{scherer2010evaluation}. The most common pooling method is max pooling where the maximum element is selected in the pooling window.

In order to feed the pooled output from stacked featured maps to the next layer, the maps are flattened into one column. The final layers in a CNN are typically fully connected~\cite{kowsari2018rmdl}.

\subsection{Neuron Activation}\label{Sec:Activation_3}
 The implementation of CNN is a discriminative trained model that uses standard back-propagation algorithm using a sigmoid (Equation~\ref{sigmoid}), (Rectified Linear Units (ReLU)~\cite{nair2010rectified}~(Equation~\ref{relu}) as activation function. The output layer for multi-class classification includes a $Softmax$ function~(as shown in Equation~\ref{Softmax}).
 \begin{align}
f(x) &= \frac{1}{1+e^{-x}}\in (0,1)\label{sigmoid}\\
f(x) &= \max(0,x)\label{relu}\\
\sigma(z)_j &= \frac{e^{z_j}}{\sum_{k=1}^K e^{z_k}}\label{Softmax}\\ 
&\forall   ~j \in \{1,\hdots, K\} \nonumber
\end{align}

\subsection{Optimizer}\label{sec:optimizer_3}
For this CNN architecture, the $Adam$ Optimizer~\cite{kingma2014adam} which is a stochastic gradient optimizer that uses only the average of the first two moments of gradient~($v$ and $m$, shown in Equation~\ref{adam}, \ref{adam1}, \ref{adam2}, and \ref{adam3}). It can handle non-stationary of the objective function as in RMSProp, while overcoming the sparse gradient issue limitation of RMSProp~\cite{kingma2014adam}.

\begin{equation}
\theta  \leftarrow \theta - \frac{\alpha}{\sqrt{\hat{v}}+\epsilon} \hat{m}\label{adam}
\end{equation}
\begin{equation}
g_{i,t} =  \nabla_\theta J(\theta_i , x_i,y_i) \label{adam1}
\end{equation}
\begin{equation}
m_t = \beta_1 m_{t-1} + (1-\beta_1)g_{i,t}\label{adam2}
\end{equation}
\begin{equation}
m_t = \beta_2 v_{t-1} + (1-\beta_2)g_{i,t}^2\label{adam3}
\end{equation}
where $m_t$ is the first moment and $v_t$ indicates second moment that both are estimated. $\hat{m_t}=\frac{m_t}{1-\beta_1^t}$ and $\hat{v_t}=\frac{v_t}{1-\beta_2^t}$.\\ 
\subsection{Network Architecture}
\begin{figure}[!t]
    \centering
    \includegraphics[width=\textwidth]{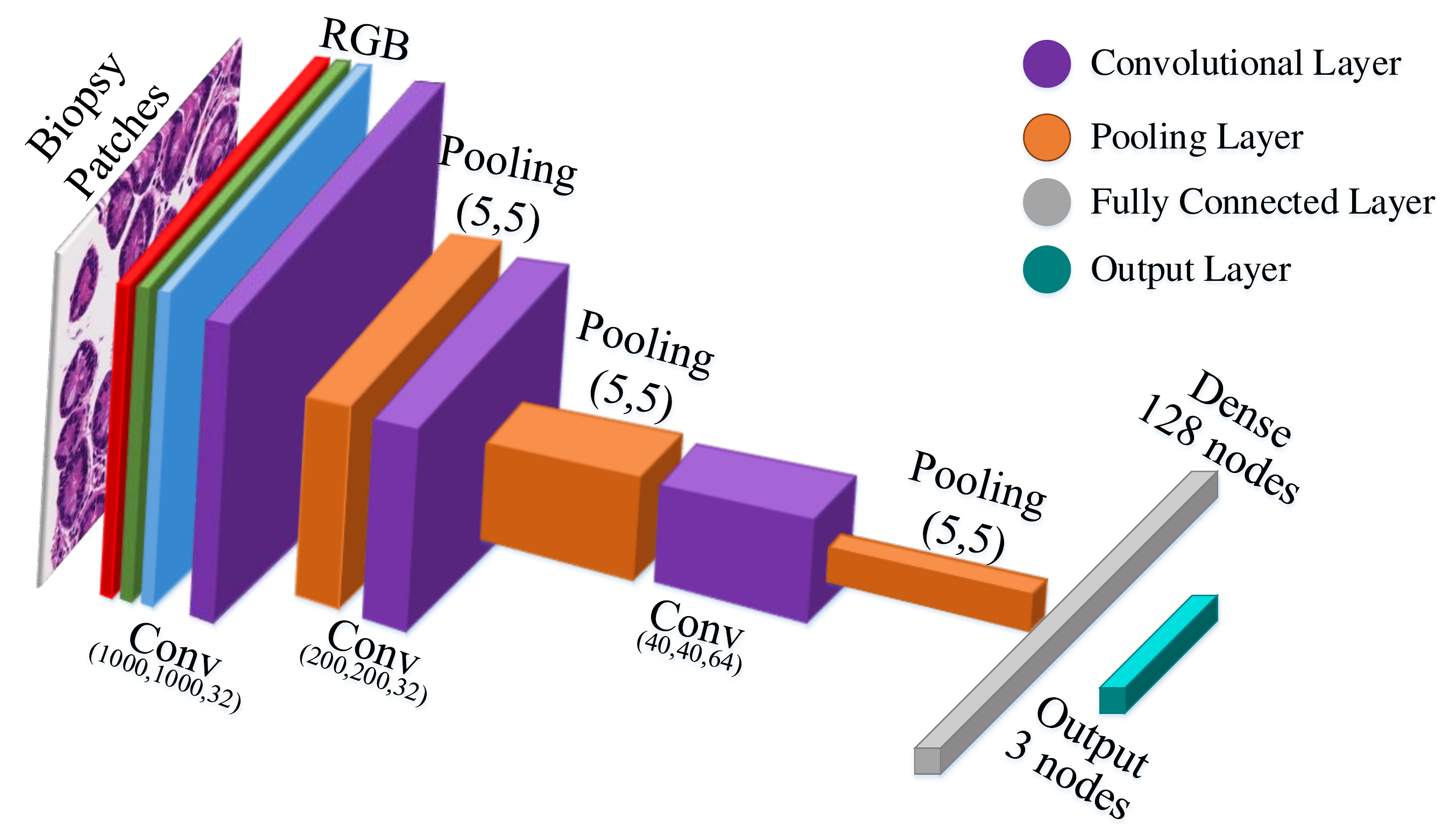}
    \caption{Our Convolutional Neural Networks' Architecture} \label{fig:cnn_Ar}
\end{figure}

As shown in Table~\ref{tb:CNN} and Figure~\ref{fig:cnn_Ar}, our CNN architecture consists of three convolution layer each followed by a pooling layer. This model receives RGB image patches with dimensions of ~$(1000\times 1000)$ as input. 

The first convolutional layer has~$32$ filters with kernel size of~$(3, 3)$. Then we have Pooling layer with size of~$(5,5)$ which reduce the feature maps from~$(1000\times 1000)$ to~$(200 \times 200)$. 

The second convolutional layers with~$32$ filters with kernel size of~$(3, 3)$. Then Pooling layer~(MaxPooling~$2D$) with size of~$(5,5)$ reduces the feature maps from~$(200\times 200)$ to~$(40 \times 40)$. The third convolutional layer has~$64$ filters with kernel size of~$(3, 3)$, and final pooling layer~(MaxPooling~$2D$) is scaled down to~$(8 \times 8)$. The feature maps as shown in Table~\ref{tb:CNN} is flatten and connected to fully connected layer with~$128$ nodes. The output layer with three nodes to represent the three classes: ~(Environmental Enteropathy, Celiac Disease, and Normal).

The optimizer used is Adam~(See Section~\ref{sec:optimizer_3}) with a learning rate of~$0.001$, $\beta_1=0.9$, $\beta_2=0.999$ and the loss considered is sparse categorical crossentropy~\cite{chollet2015keras}. Also for all layers, we use Rectified linear unit~(ReLU) as activation function except output layer which we use~$Softmax$~(See Section~\ref{Sec:Activation_3}).  

\begin{table}[h]
\centering
\caption{CNN Architecture for Diagnosis of Diseased Duodenal on Biopsy Images}\label{tb:CNN}
\begin{tabular}{|c|c|c|c|}
\hline
   & Layer (type)        & Output Shape     & \begin{tabular}[c]{@{}c@{}}Trainable \\ Parameters\end{tabular} \\ \hline
1  & Convolutional Layer & $(1000, 1000, 32)$ & $869$                                                             \\ \hline
2  & Max Pooling        & $(200, 200, 32)$   & $0$                                                               \\ \hline
3  & Convolutional Layer & $(200, 200, 32)$   & $9,248$                                                            \\ \hline
4  & Max Pooling        & $(40, 40, 32)$     & $0$                                                               \\ \hline
5  & Convolutional Layer & $(40, 40, 64)$     & $18,496$                                                           \\ \hline
6  & Max Pooling        & $(8, 8, 64)$       & $0$                                                               \\ \hline
7  & dense               & $128$              & $524,416$                                                          \\ \hline
8 & Output              & $3$                & $387$                                                             \\ \hline
\end{tabular}
\end{table}

\section{Empirical Results}\label{sec:Empirical_Results_3}

\subsection{Evaluation Setup}\label{sec:Evaluation_}
In the research community, comparable and shareable performance measures to evaluate algorithms are preferable. However, in reality such measures may only exist for a handful of methods. The major problem when evaluating image classification methods is the absence of standard data collection protocols. Even if a common collection method existed, simply choosing different training and test sets can introduce inconsistencies in model performance~\cite{yang1999evaluation}. Another challenge with respect to method evaluation is being able to compare different performance measures used in separate experiments. Performance measures generally evaluate specific aspects of classification task performance, and thus do not always present identical information. In this section, we discuss evaluation metrics and performance measures and highlight ways in which the performance of classifiers can be compared.

\subsection{Experimental Setup}

The following results were obtained using a combination of central processing units~(CPUs) and graphical processing units~(GPUs). The processing was done on a $Xeon~E5-2640~ (2.6 GHz)$ with $32$ cores and $64 GB$ memory, and the GPU cards were two $Nvidia~Titan~Xp$ and a $Nvidia~Tesla~K20c$. We implemented our approaches in Python using the Compute Unified Device Architecture~(CUDA), which is a parallel computing platform and Application Programming Interface~(API) model created by $Nvidia$. We also used Keras and TensorFlow libraries for creating the neural networks~\cite{abadi2016tensorflow,chollet2015keras}. 

\subsection{Experimental Results}
In this section we show that CNN with color balancing can improve the robustness of  medical image classification. The results for the model trained on $4$ different color balancing values are shown in Table~\ref{tb:1}. The results shown in Table~\ref{tb:2} are also based on the trained model using the same color balancing values. Although in Table~\ref{tb:2}, the test set is based on a different set of color balancing values: ~$0.5, 1.0, 1.5$~and~$2.0$. By testing on a different set of color balancing, these results show that this technique can solve the issue of multiple stain variations during histological image analysis.

As shown in Table~\ref{tb:1}, the f1-score of three classes~(Environmental Enteropathy~(EE), Celiac Disease~(CD), and Normal) are $0.98$, $0.94$, and $0.91$ respectively. In Table~\ref{tb:2}, the f1-score is reduced, but not by a significant amount. The three classes~(Environmental Enteropathy~(EE), Celiac Disease~(CD), and Normal) f1-scores are $0.94$, $0.92$, and $0.87$ respectively. The result is very similar to \textit{MA. Boni et.al}~\cite{Mohammad_al_boni} which achieved 90.59\% of accuracy in their mode, but without using the color balancing technique to allow differently stained images.

\begin{table}[h]
\centering
\caption{F1-score for train on a set with color balancing of 0.001, 0.01, 0.1, and 1.0. Then, we evaluate test set with same color balancing}\label{tb:1}
\begin{tabular}{|c|c|c|c|c|}
\hline
   & precision & recall & f1-score & support \\ \hline
Celiac Disease (CD)  & $0.89$      & $0.99$   & $0.94$     & $22,196$   \\ \hline
Normal  & $0.99$      & $0.83$   & $0.91$     & $22,194$   \\ \hline
\begin{tabular}[c]{@{}c@{}}Environmental  Enteropathy \\ (EE) \end{tabular} & $0.96$      & $1.00$   & $0.98$     & $22,198$   \\ \hline
\end{tabular}
\end{table}

\begin{table}[h]
\centering
\caption{F1-score for train with color balancing of 0.001, 0.01, 0.1, and 1.0 and test with color balancing of 0.5, 1.0, 1.5 and 2.0}\label{tb:2}
\begin{tabular}{|c|c|c|c|c|}
\hline
   & precision & recall & f1-score & support \\ \hline
Celiac Disease (CD)  & $0.90$      & $0.94$   & $0.92$     & $22,196$   \\ \hline
Normal  & $0.96$      & $0.80$   & $0.87$     & $22,194$   \\ \hline
\begin{tabular}[c]{@{}c@{}}Environmental Enteropathy  \\ (EE) \end{tabular} & $0.89$      & $1.00$   & $0.94$     & $22,198$   \\ \hline
\end{tabular}
\end{table}

\begin{figure}[!hbt]
    \centering
    \includegraphics[width=\textwidth]{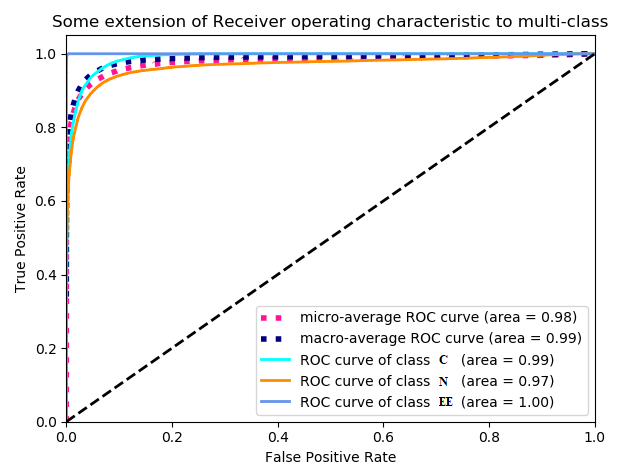}
    \caption{Receiver operating characteristics~(ROC) curves for three classes also the figure shows micro-average and macro-average of our classifier} \label{fig:ROC_3}
\end{figure}

In Figure~\ref{fig:ROC_3}, Receiver operating characteristics~(ROC) curves are valuable graphical tools for evaluating classifiers. However, class imbalances (i.e. differences in prior class probabilities) can cause ROC curves to poorly represent the classifier performance. ROC curve plots true positive rate~(TPR) and false positive rate~(FPR). The ROC shows that AUC of Environmental Enteropathy~(EE) is~$1.00$, Celiac Disease~(CD) is 0.99, and Normal is 0.97.

\begin{table}[!hbt]
\caption{Comparison accuracy with different baseline methods}\label{tb:compare}
\begin{tabular}{lccc}
\hline
Method                    & ~~\begin{tabular}[c]{@{}c@{}}Solve Color\\  Staining Problem\end{tabular}~~~ & ~\begin{tabular}[c]{@{}c@{}}Model\\ Architecture\end{tabular}~ & ~Accuracy~       \\ \hline
Shifting and Reflections~\cite{Mohammad_al_boni}  & No                     & CNN                & 85.13\%        \\ 
Gamma~\cite{Mohammad_al_boni}                     & No                     & CNN                & 90.59\%        \\ 
CLAHE~\cite{Mohammad_al_boni}                     & No                     & CNN                & 86.79\%        \\ 
Gamma-CLAHE~\cite{Mohammad_al_boni} & No                     & CNN                & 86.72\%        \\ 

Fine-tuned ALEXNET~\cite{nawaz2018classification}          & \textbf{Yes}           & ALEXNET            & 89.95\%     \\     
Ours                      & \textbf{Yes}            & CNN                & \textbf{93.39\% } \\ \hline
\end{tabular}
\end{table}

As shown in Table~\ref{tb:compare}, our model performs better compared to some other models in terms of accuracy. Among the compared models, only the fine-tuned ALEXNET~\cite{nawaz2018classification} has considered the color staining problem. This model proposes a transfer learning based approach for the classification of stained histology images. They also applied stain normalization before using images for fine tuning the model.

\section{Conclusion}\label{sec:Conclusion_3}
In this dissertation section, we proposed a data driven model for diagnosis of diseased duodenal architecture on biopsy images using color balancing on convolutional neural networks. Validation results show that this model is very robust and accurate in comparing with our baselines. Furthermore, color consistency is an issue in digital histology images and different imaging systems reproduced the colors of a histological slide differently. Our results demonstrate that application of the color balancing technique can attenuate effect of this issue in image classification.  

The methods described here can be improved in multiple ways. Additional training and testing with other color balancing techniques on data sets will continue to identify architectures that work best for these problems. Also, it is possible to extend the model to more than four different color balance percentages to capture more of the complexity in the medical image classification.
\chapter{Random Multimodel Deep Learning~(RMDL) for Medical Image Classification}\label{chpt:RMDL}

The continually increasing number of complex datasets each year necessitates ever improving machine learning methods for robust and accurate categorization of these data. This paper introduces Random Multimodel Deep Learning~(RMDL): a new ensemble, deep learning approach for classification. Deep learning models have achieved state-of-the-art results across many domains.  RMDL solves the problem of finding the best deep learning structure and architecture while simultaneously improving robustness and accuracy through ensembles of deep learning architectures. RDML can accept as input a variety data to include text, video, images, and symbolic. This chapter describes RMDL and shows test results for medical images. For testing RMDL, we resize all images to $100\times 100$ due to time complexity of this algorithm is very high.

\section{Introduction}
Categorization and classification with complex data such as medical images, images, and video are central challenges in the data science community. Recently, there has been an increasing body of work using deep learning structures and architectures for such problems. However, the majority of these deep architectures are designed for a specific type of data or domain. There is a need to develop more general information processing methods for classification and categorization across a broad range of data types. 

While many researchers have successfully used deep learning for classification problems (\textit{e.g.,} see~\cite{kowsari2017hdltex,lecun2015deep,lee2009convolutional,chung2014empirical,turan2017deep,kowsari2018web}), the central problem remains as to which deep learning architecture~(Multilayer perceptron, CNN, or RNN) and structure~(how many nodes~(units) and hidden layers) is more efficient for different types of data and applications. The favored approach to this problem is trial and error for the specific application and dataset.
This section describes an approach to this challenge using ensembles of deep learning architectures. This approach, called Random Multimodel Deep Learning~(RMDL), uses three different deep learning architectures: Deep Neural Networks~(DNN or Multilayer Perceptron), Convolutional Neural Networks~(CNN), and Recurrent Neural Networks~(RNN). Test results with a variety of data types demonstrate that this new approach is highly accurate, robust and efficient.
The three basic deep learning architectures use different feature space methods as input layers. RDML searches across randomly generated hyperparameters for the number of hidden layers and nodes~(density) in each hidden layer in the Multilayer Perceptron. CNN has been well designed for image classification. RMDL finds choices for hyperparameters in CNN using random feature maps and random numbers of hidden layers. CNN can be used for more than image data. The structures for CNN used by RMDL contains 2D for images.
RNN architectures are mostly used primarily for text classification, but it could be used for image classification~\cite{mou2017deep}. RMDL uses two specific RNN structures: Gated Recurrent Units~(GRUs) and Long Short-Term Memory~(LSTM). The number of GRU or LSTM units and hidden layers used by the RDML are also the results of search over randomly generated hyperparameters.

The main contributions of this work are as follows:~\RNum{1}) Description of an ensemble approach to deep learning which makes the final model more robust and accurate.~\RNum{2}) Use of different optimization techniques in training the models to stabilize the classification task.~\RNum{3}) Use of dropout in each individual RDL to address over-fitting.~\RNum{4}) Use of majority voting among the ~$n$ RDL models. This majority vote from the ensemble of RDL models improves the accuracy and robustness of results. Specifically, if~$k$ number of RDL models produce inaccuracies or over-fitting classifications and $n > k$, the overall system is robust and accurate~\RNum{5}) Finally, the RMDL has ability to process a variety of data types such as images and videos.

\section{Method}\label{method}\label{sec:method_4}

The novelty of this work is in using multi random deep learning models including Multilayer Perceptron, RNN, and CNN techniques for text and image classification. The method section of this paper is organized as follows: first we describe RMDL and we discuss three techniques of deep learning architectures~(Multilayer Perceptron, RNN, and CNN) which are trained in parallel. Next, we talk about multi optimizer techniques that are used in different random models. 
\subsection{Feature Extraction and Data Pre-processing}\label{subsect:Feature}
The feature extraction is divided into two main parts for RMDL~(Text and image). Text and sequential datasets are unstructured data, while the feature space is structured for image datasets.
\subsubsection{Image and 3D Object Feature Extraction}
Image features are the followings:~$h\times~w\times~c$ where~$h$ denotes the height of the image,~$w$ represents the width of image, and~$c$ is the color that has 3 dimensions~(RGB). For gray scale datasets such as $MNIST$ dataset, the feature space is~$h\times~w$. A 3D object in space contains~$n$ cloud points in space and each cloud point has~$6$ features which are~(\textit{x, y, z, R, G, and B}). The 3D object is unstructured due to number of cloud points since one object could be different with others. However, we could use simple instance down/up sampling to generate the structured datasets.
\begin{figure*}[t]
\centering
\includegraphics[width=\textwidth]{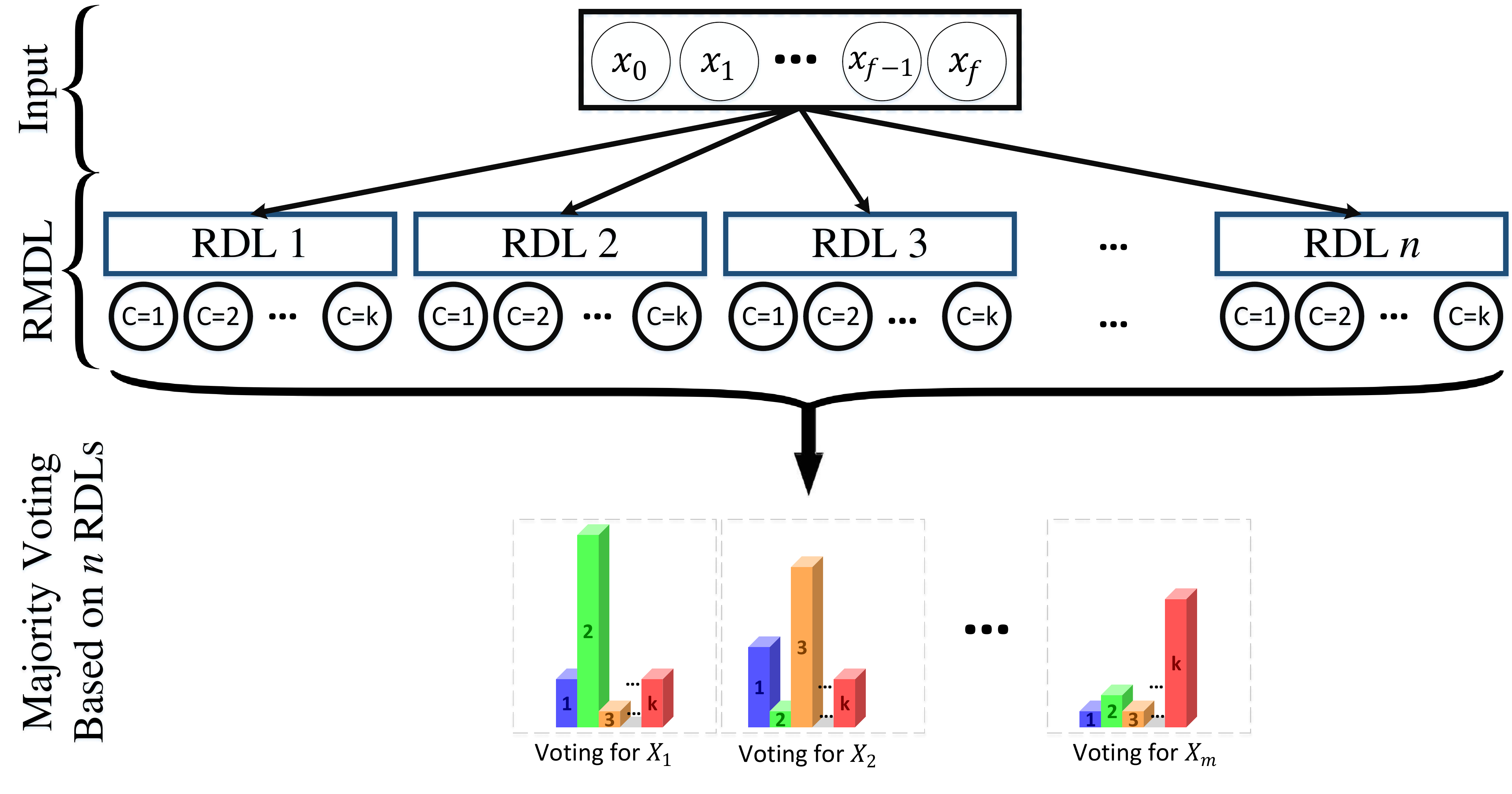}

\caption{Overview of RDML: \underline{R}andom \underline{M}ultimodel \underline{D}eep \underline{L}earning for classification that includes $n$ Random models which are $d$ random model of the Multilayer Perceptron classifiers, $c$ models of CNN classifiers, and $r$ RNN classifiers where~$r+c+d=n$.}\label{Fig_RMDL}

\end{figure*}

\subsection{Random Multimodel Deep Learning}\label{subsect:RMDL}
Random Multimodel Deep Learning is a novel technique that we can use in any kind of dataset for classification. An overview of this technique is shown in Figure~\ref{Fig_RMDL} which contains multi Deep Neural Networks~(DNN or Multilayer Perceptron), Deep Convolutional Neural Networks~(CNN), and Deep Recurrent Neural Networks~(RNN). The number of layers and nodes for all of these Deep learning multi models are generated randomly~(\textit{e.g.}~9 Random Models in RMDL constructed of~$3$ CNNs,~$3$ RNNs, and~$3$ Multilayer Perceptrons, all of them are unique due to randomly creation). 
\begin{align}
\label{eq:majority}
M(y_{i1},y_{i2},...,y_{in}) =& \bigg\lfloor \frac{1}{2}+ \frac{(\sum_{j=1}^n y_{ij}) - \frac{1}{2}}{n}\bigg\rfloor
\end{align}
Where $n$ is the number of random models, and $y_{ij}$ is the output prediction of model for data point $i$ in  model $j$~(Equation~\ref{eq:majority} is used for binary classification, $k\in\{0~ \text{or}~1\}$). Output space uses majority vote for final $\hat{y_i}$. Therefore,~$\hat{y_i}$ is given as follows:
\begin{equation}
\hat{y_i} =  
\begin{bmatrix}
\hat{y}_{i1} ~
\hdots ~
\hat{y}_{ij}~
\hdots~
\hat{y}_{in}~
\end{bmatrix}^T
\end{equation} 

Where $n$ is number of random model, and $\hat{y}_{ij}$ shows the prediction of label of document or data point of $D_i \in \{x_i,y_i\}$ for model $j$ and $\hat{y}_{i,j}$ is defined as follows:
\begin{equation}
\hat{y}_{i,j} = arg \max_{k} [ softmax(y_{i,j}^*)]
\end{equation}
After all RDL models~(RMDL) are trained, the final prediction is calculated using majority vote of these models.

\subsection{Deep Learning in RMDL}\label{subsec:Deep_learning}
The RMDL model structure~(section~\ref{subsect:RMDL}) includes three basic architectures of deep learning in parallel. We describe each individual model separately. The final model contains~$d$ random Multilayer Perceptrons~(Section~\ref{subsubsec:DNN}),~$r$ RNNs~(Section~\ref{subsubsec:RNN}), and~$c$ CNNs models~(Section~\ref{subsubsec:CNN}).
\begin{figure*}[t]
\centering
\includegraphics[width=\textwidth]{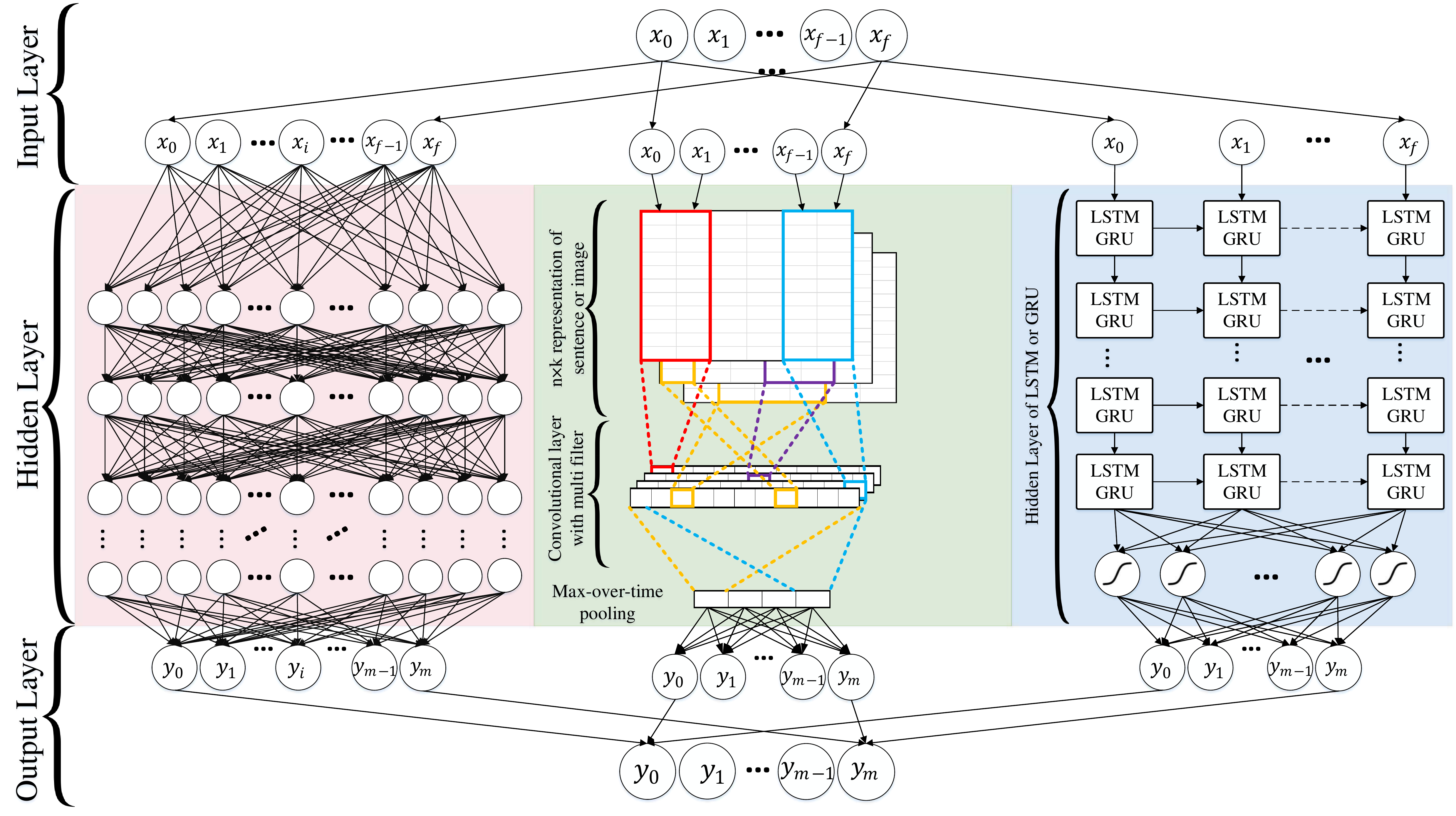}
\caption{\underline{R}andom \underline{M}ultimodel \underline{D}eep \underline{L}earning~(RDML) architecture for classification which includes~$3$ Random models, a Multilayer Perceptron classifier at left, a Deep CNN classifier at middle, and a Deep RNN classifier at right~(each unit could be LSTM or GRU).}

\end{figure*}
\subsubsection{Deep Neural Networks}\label{subsubsec:DNN}
Deep Neural Networks' structure is designed to learn by multi connection of layers that each layer only receives connection from previous and provides connections only to the next layer in hidden part. The input is a connection of feature space with first hidden layer for all random models. The output layer is number of classes for multi-class classification and only one output for binary classification. But our main contribution of this paper is that we have many training Multilayer Perceptrons for different purposes. In our techniques, we have multi-classes Multilayer Perceptrons where each learning models is generated randomly~(number of nodes in each layer and also number of layers are completely random assigned). Our implementation of Deep Neural Networks~(DNN) is discriminative trained model that uses standard back-propagation algorithm using sigmoid~(equation~\ref{sigmoid}), ReLU~\cite{nair2010rectified}~(equation~\ref{relu}) as activation function. The output layer for multi-class classification, should use $Softmax$ equation~\ref{Softmax}.

\subsubsection{Recurrent Neural Networks~(RNN)}\label{subsubsec:RNN}
Another neural network architecture that contributes in RMDL is Recurrent Neural Networks~(RNN). RNN assigns more weights to the previous data points of sequence. Therefore, this technique is a powerful method for text, string and sequential data classification but also  could be used for image classification as we did in this work. In RNN the neural net considers the information of previous nodes in a very sophisticated method which allows for better semantic analysis of structures of dataset. General formulation of this concept is given in Equation~\ref{rnn_gen} where $x_t$ is the state at time $t$ and $\boldsymbol{u_t}$ refers to the input at step t.

\begin{equation}
\label{rnn_gen}
x_{t}=F(x_{t-1},\boldsymbol{u_t},\theta)
\end{equation}

More specifically, we can use weights to formulate the Equation~\ref{rnn_gen} with  specified parameters in Equation~\ref{rnn_spec}

\begin{equation}\label{rnn_spec}
x_{t}=\mathbf{W_{rec}}\sigma(x_{t-1})+\mathbf{W_{in}}\mathbf{u_t}+\mathbf{b}
\end{equation}
Where $\mathbf{W_{rec}}$ refers to recurrent matrix weight, $\mathbf{W_{in}}$ refers to input weights, $\mathbf{b}$ is the bias and $\sigma$ denotes an element-wise function. 

Again, we have modified the basic architecture for use RMDL. Figure \ref{Fig_RMDL} left side shows this extended RNN architecture. Several problems arise from RNN when the error of the gradient descent algorithm is back propagated through the network: vanishing gradient and exploding gradient  \cite{bengio1994learning}. \\

\paragraph{Long Short-Term Memory~(LSTM)}

\begin{figure}[!t]
\centering
\includegraphics[width=\columnwidth]{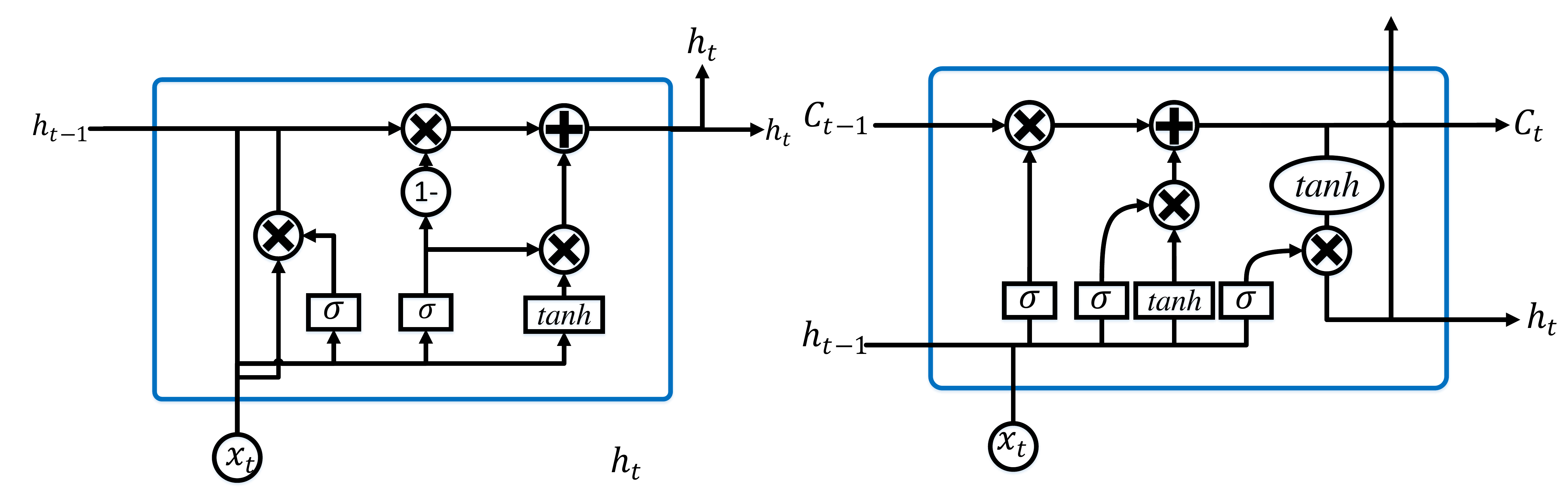}
\caption{left Figure is a cell of GRU, and right Figure is a cell of LSTM}\label{fig:LSTM}

\end{figure}

To deal with these problems Long Short-Term Memory~(LSTM) is a special type of RNN that preserve long term dependency in a more effective way in comparison to the basic RNN. This is particularly useful to overcome vanishing gradient problem~\cite{pascanu2013difficulty}. Although LSTM has a chain-like structure similar to RNN,
LSTM uses multiple gates to carefully regulate the amount of information that will be allowed into each node state. Figure~\ref{fig:LSTM} shows the basic cell of a LSTM model. A step by step explanation of a LSTM cell is as following:

\begin{align}
    &&i_{t}=&\sigma(W_{i}[x_{t},h_{t-1}]+b_{i}),&& \label{eq:lstm1}\\
    &&\tilde{C_{t}}=&\tanh(W_{c}[x_{t},h_{t-1}]+b_{c}),&& \label{eq:lstm2} \\
    &&f_{t}=&\sigma(W_{f}[x_{t},h_{t-1}]+b_{f}),&& \label{eq:lstm3}\\
    &&C_{t}=&  i_{t}* \tilde{C_{t}}+f_{t} C_{t-1},&& \label{eq:lstm4}\\
    &&o_{t}=& \sigma(W_{o}[x_{t},h_{t-1}]+b_{o}),&& \label{eq:lstm5}\\
    &&h_{t}=&o_{t}\tanh(C_{t}),&&\label{eq:lstm6}
\end{align}

Where equation~\ref{eq:lstm1} is input gate, Equation~\ref{eq:lstm2} shows candid memory cell value, Equation~\ref{eq:lstm3} is forget gate activation, Equation~\ref{eq:lstm4} is new memory cell value, and  Equation~\ref{eq:lstm5} and~\ref{eq:lstm6} show output gate value. In the above description all $b$ represents bias vectors and all $W$ represent weight matrices and $x_{t}$ is used as input to the memory cell at time~$t$. Also,~$i,c,f,o$ indices refer to input, cell memory, forget and output gates respectively.
 Figure~\ref{fig:LSTM} shows the structure of these gates with a graphical representation.\\
An RNN can be biased when later words are more influential than the earlier ones. To overcome this bias Convolutional Neural Network~(CNN)  models~(discussed in Subsection~\ref{subsubsec:CNN} were introduced which deploys a max-pooling layer to determine discriminative phrases in a text~\cite{lai2015recurrent}.\\

\paragraph{Gated Recurrent Unit~(GRU):}\label{subsec:GRU}

Gated Recurrent Unit~(GRU) is a gating mechanism for RNN which was introduced by~\cite{chung2014empirical} and~\cite{cho2014learning}. GRU is a simplified variant of the
LSTM architecture, but there are differences as follows: GRU contains two gates, a GRU does not possess internal memory (the $C_{t-1}$ in Figure~\ref{fig:LSTM}); and finally, a second non-linearity is not applied~(tanh in Figure~\ref{fig:LSTM}). A step by step explanation of a GRU cell is as following:
\begin{equation}
z_{t}=\sigma_g(W_{z}x_{t}+U_zh_{t-1}+b_{z}), \label{eq:gru1}
\end{equation}
Where~$z_t$ refers to update gate vector of~$t$,~$x_t$ stands for input vector,~$W$, $U$ and~$b$ are parameter matrices and vector, $\sigma_g$ is activation function that could be sigmoid or ReLU.
\begin{equation}
    \tilde{r_{t}}=\sigma_g(W_{r}x_{t}+U_rh_{t-1}+b_{r}), \label{eq:gru2}
\end{equation}
\begin{equation}
   h_t =  z_t \circ h_{t-1} + (1-z_t) \circ \sigma_h(W_{h} x_t + U_{h} (r_t \circ h_{t-1}) + b_h)\label{eq:gru6}
\end{equation}
Where~$h_t$ is output vector of~$t$, $r_t$ stands for reset gate vector of~$t$, $z_t$ is update gate vector of~$t$, $\sigma_h$ indicates the hyperbolic tangent function.

\subsubsection{Convolutional Neural Networks~(CNN)}\label{subsubsec:CNN}

The final deep learning approach which contributes in RMDL is  Convolutional Neural Networks~(CNN) that is employed for  document or image classification. Although originally built for image processing with architecture similar to the visual cortex, CNN have also been effectively used for  other datasets classification ~\cite{lecun1998gradient}; thus, in RMDL, this technique is used in all datasets. \\ In the basic CNN for image processing an image tensor is convolved with a set of kernels of size $d \times d$. These convolution layers are called feature maps and can be stacked to provide multiple filters on the input. To reduce the computational complexity CNN use pooling which reduces the size of the output from one layer to the next in the network. Different pooling techniques are used to reduce outputs while preserving important features ~\cite{scherer2010evaluation}. The most common pooling method is max pooling where the maximum element is selected in the pooling window.\\ In order to feed the pooled output from stacked featured maps to the final layer, the maps are flattened into one column. The final layers in a CNN are typically fully connected.\\
In general, during the back propagation step of a convolutional neural network not only the weights are adjusted but also the feature detector filters. A potential problem of CNN used for other datasets are the number of 'channels', $\Sigma$~(size of the feature space). This might be very large~(\textit{e.g.} 10K) but for images this is less of a problem~(\textit{e.g.} only 3 channels of RGB)~\cite{johnson2014effective}.

\begin{figure}[H]
\centering

\includegraphics[width=0.85\columnwidth]{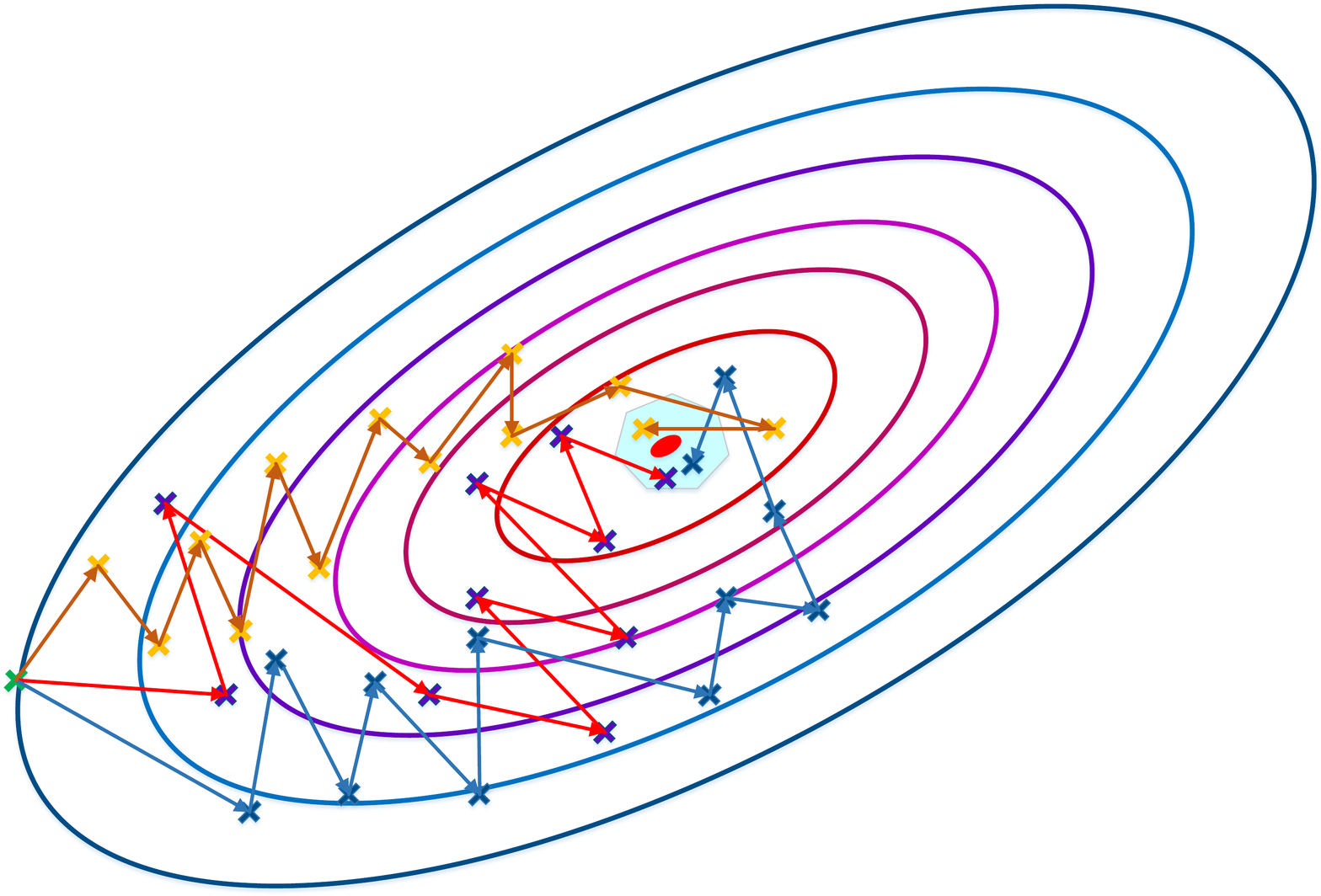}
\caption{This figure Shows multi SGD optimizer }\label{Optimizer}
\end{figure}

\subsection{Optimization}\label{subsec:Optimization}
In this paper we use two types of stochastic gradient optimizer in our neural networks implementation which are RMSProp  and Adam optimizer:
\subsubsection{Stochastic Gradient Descent~(SGD) Optimizer}
SGD has been used as one of our optimizers that is shown in equation~\ref{SGD}. It uses a momentum on re-scaled gradient which is shown in equation~\ref{momentum} for updating parameters. The other technique of optimizer that is used is RMSProp which does not do bias correction. This will be a significant problem while dealing with sparse gradient. 
\begin{align}
\label{SGD}
\theta &\leftarrow \theta - \alpha \nabla_\theta J(\theta , x_i,y_i)\\
\label{momentum}
\theta &\leftarrow \theta -\big( \gamma \theta + \alpha \nabla_\theta J(\theta , x_i,y_i)\big)
\end{align}
\subsubsection{Adam Optimizer}
Adam is another stochastic gradient optimizer which uses only the first two moments of gradient~($v$ and $m$ that are shown in equation~\ref{adam},   \ref{adam1}, \ref{adam2}, and \ref{adam3}) and average over them. It can handle non-stationary of objective function as in RMSProp while overcoming the sparse gradient issue that was a drawback in RMSProp~\cite{kingma2014adam}.

\subsubsection{Multi Optimization rule}
The main idea of using multi model with different optimizers is that if one optimizer does not provide a good fit for a specific datasets, the RMDL model with~$n$ random models~(some of them might use different optimizers) could ignore~$k$ models which are not efficient if and only if $n>k$. The Figure~\ref{Optimizer} provides a visual insight on how three optimizers work better in the concept of majority voting. Using multi techniques of optimizers such as SGD, adam, RMSProp, Adagrad, Adamax, and so on helps the RMDL model to be more stable for any type of datasets. In this research, we only used two optimizers~(Adam and RMSProp) for evaluating our model, but the RMDL model has the capability to use any kind of optimizer.


\section{Limitation and Discussion}

Model interpretability of deep learning (DL), especially RMDL, has always been a limiting factor for use cases requiring explanations of the features involved in modelling and such is the case for many healthcare problems. This problem is due to scientists preferring to use traditional techniques such as  linear models, SVM, decision trees, etc. for their works. The weights in a neural network are a measure of how strong each connection is between each neuron to find the important feature space. The more accurate model, the interpretability is lower which means the complex algorithms such as deep learning is hard to understand. This problem is even worse if we used ensemble deep learning which is used in this chapter.  Deep learning~(DL) is one of the most powerful techniques in artificial intelligence~(AI), and many researchers and scientists focus on deep learning architectures to improve the robustness and computational power of this tool.  However, deep learning architectures also have some disadvantages and limitations when applied to classification tasks. One of the main problems of this model is that DL does not facilitate a comprehensive theoretical understanding of learning~\cite{shwartz2017opening}. A well-known disadvantage of DL methods is their \qu{black box} nature~\cite{gray1996alternatives,shrikumar2017learning}. That is, the method by which DL methods come up with the convolved output is not readily understandable. Another limitation of DL is that it usually requires much more data than traditional machine learning algorithms, which means that this technique cannot be applied to classification tasks over small data sets~\cite{anthes2013deep,lampinen2017one}. Additionally, the massive amount of data needed for DL classification algorithms further exacerbates the computational complexity during the training step~\cite{severyn2015learning}. RMDL for medical image classification takes more time in comparing with other techniques which for training RMDL with 15 model (we already resize our dataset) takes 78 hours to be trained in two GPUs. 

\begin{table}[!b]
\centering
\caption{Results of RMDL per Class for 3, 9 and 15 Random models models}\label{ya:RMDL_per_class}

\begin{tabular}{|c|c c c c c|}
\hline
RMDL                 & Class & Accuracy & Precision & Recall & F1-Score \\ \hline
\multirow{3}{*}{3 RDLs}  & N     &    82.87$\pm$1.20      &    83.85$\pm$1.17      &    82.87$\pm$1.20    &     83.36$\pm$1.18   \\ \cline{2-6} 
                    & EE    &    94.71$\pm$0.72     &    98.33$\pm$0.41       &   92.94$\pm$0.82    &    93.82$\pm$0.77      \\ \cline{2-6} 
                    & C     &    82.91$\pm$1.2      &    82.95$\pm$1.21       &    82.36$\pm$1.23    &   82.66$\pm$1.22       \\ \hline
\multirow{3}{*}{9 RDLs}  & N     &    90.84$\pm$0.92      &    89.51$\pm$ 0.97      &   90.84$\pm$0.92     &     90.17$\pm$0.95     \\ \cline{2-6} 
                    & EE    &    94.71$\pm$0.71      &    95.49$\pm$0.67      &    94.71$\pm$0.72    &     95.10$\pm$0.69     \\ \cline{2-6} 
                    & C     &    87.80$\pm$1.05      &    88.42$\pm$1.03       &    87.80$\pm$1.05    &     88.11$\pm$1.04     \\ \hline
\multirow{3}{*}{15 RDLs} & N     &    88.32 $\pm$ 1.02     &   90.50$\pm$ 0.93        &   88.32$\pm$1.02     &     89.40$\pm$0.98     \\ \cline{2-6} 
                    & EE    &  97.88$\pm$0.46        &    95.65$\pm$0.65       &    97.88$\pm$0.46    &    96.76$\pm$0.57      \\ \cline{2-6} 
                    & C     &    83.72$\pm$1.19      &    88.41$\pm$1.03       &    88.53$\pm$1.02    &     88.47$\pm$1.03     \\ \hline
\end{tabular}
\end{table}

\section{Results of RMDL for Medical Images}

\subsection{Data}

\subsubsection{Data Source} As shown in Table~\ref{ta:population}, The total population of this dataset is 150 children participated in this study with a median (interquartile range) age of 37.5 (19.0 to 121.5) months and a roughly equal sex distribution; 77 males (51.3\%). The slides were converted into~$461$ whole slide images, and labeled as either EE, CD, or normal. The biopsy slides for EE patients were from the Aga Khan University Hospital~(AKUH) in Karachi, Pakistan~($n = 29$ slides from~$10$ patients) and the University of Zambia Medical Center in Lusaka, Zambia ($n = 16$). The slides for CD patients ($n = 34$) and normal ($n = 63$) were retrieved from archives at the University of Virginia~(UVa). The CD and normal slides were converted into whole slide images at~$40$x magnification using the Leica SCN~$400$ slide scanner (Meyer Instruments, Houston, TX) at UVa, and the digitized EE slides were of 20x magnification and shared via the Environmental Enteric Dysfunction Biopsy Investigators~(EEDBI) Consortium shared WUPAX server. Characteristics of our patient population are as follows: the median~($Q1$, $Q3$) age of our entire study population was~$37.5$~($19.0$, $121.5$) months, and we had a roughly equal distribution of males~($52$\%, $n = 53$) and females~($48$\%, $n = 49$). The majority of our study population were histologically normal controls~$(37.7\%)$, followed by CD patients~$(51.8\%)$, and EE patients~$(10.05\%)$.

\subsubsection{Pre-Processing}
In this section, we cover all of the pre-processing steps which include image patching, image clustering, and resizing images. The biopsy images are unstructured~(varying image sizes) and too large to process with deep neural networks; thus, requiring that images are split into multiple smaller images. After executing the split, some of the images do not contain much useful information. For instance, some only contain the mostly blank border region of the original image. In the image clustering section, the process to select useful images is described. Finally, color balancing is used to correct for varying color stains which is a common issue in histological image processing.

\paragraph{Image Patching}\label{subsec:Image_Patching4}
Although the effectiveness of CNNs in image classification has been shown in various studies in different domains, training on high-resolution Whole Slide Tissue Images (WSI) is not commonly preferred due to a high computational cost. However, applying CNNs on WSI enables losing a large amount of discriminative information due to extensive down-sampling~\cite{hou2016patch}. Due to a cellular level difference between Celiac, Environmental Enteropathy and normal cases, a trained classifier on image patches are likely to perform as well as or even better than a trained WSI-level classifier. Many researchers in pathology image analysis have considered classification or feature extraction on image patches~\cite{hou2016patch}.
In this project, after generating patches from each image, labels were applied to each patch according to its associated original image. CNN was trained to generate predictions on each individual patch. As you have shown in Figure~\ref{fig:Patch}, each biopsy whole image is divided into many patches.

\paragraph{Patch Clustering} As we discussed in Section~\ref{subsec:lower:Clustering}, Clustering is organizing objects in a such way that objects within a group or cluster in some way are more similar to each other compared to objects in other groups. There is a wide variety of algorithms for data clustering and K-means clustering is one of the easiest ones~\cite{jain2010data}.

\subsection{Empirical results}
As shown in Table~\ref{ta:results_per_class}, results per class is represent that if the more RDLs is used, that accuracy, precision, recall and F1-score are improved. Based on this table accuracy of Normal for RMDL with 3, 9, ad 15 RDLs is 82.87$\pm$1.20 and 90.84$\pm$0.92 and 88.32$\pm$1.02 respectively. The F1-Score of Normal for RMDL with 3, 9, ad 15 RDLs is 83.36$\pm$1.18 and 90.17$\pm$0.95 and  89.40$\pm$0.98 respectively. The results per class is represent that if the more RDLs is used, that accuracy, precision, recall and F1-score are improved. Based on this table accuracy of CD for RMDL with 3, 9, ad 15 RDLs is 82.91$\pm$1.2 and 87.80$\pm$0.1.05 and 83.72$\pm$1.19 respectively. The F1-Score of CD for RMDL with 3, 9, ad 15 RDLs is 82.66$\pm$0.77 and 88.11$\pm$1.04 and  88.47$\pm$1.03 respectively. Based on this table accuracy of EE for RMDL with 3, 9, ad 15 RDLs is 94.71$\pm$0.72 and 94.71$\pm$0.71 and 97.88$\pm$0.46 respectively. The F1-Score of CD for RMDL with 3, 9, ad 15 RDLs is 93.82$\pm$0.77 and 95.1$\pm$0.69 and  96.86$\pm$0.57 respectively.

\begin{figure}[!t]
    \centering
    \includegraphics[width=\textwidth]{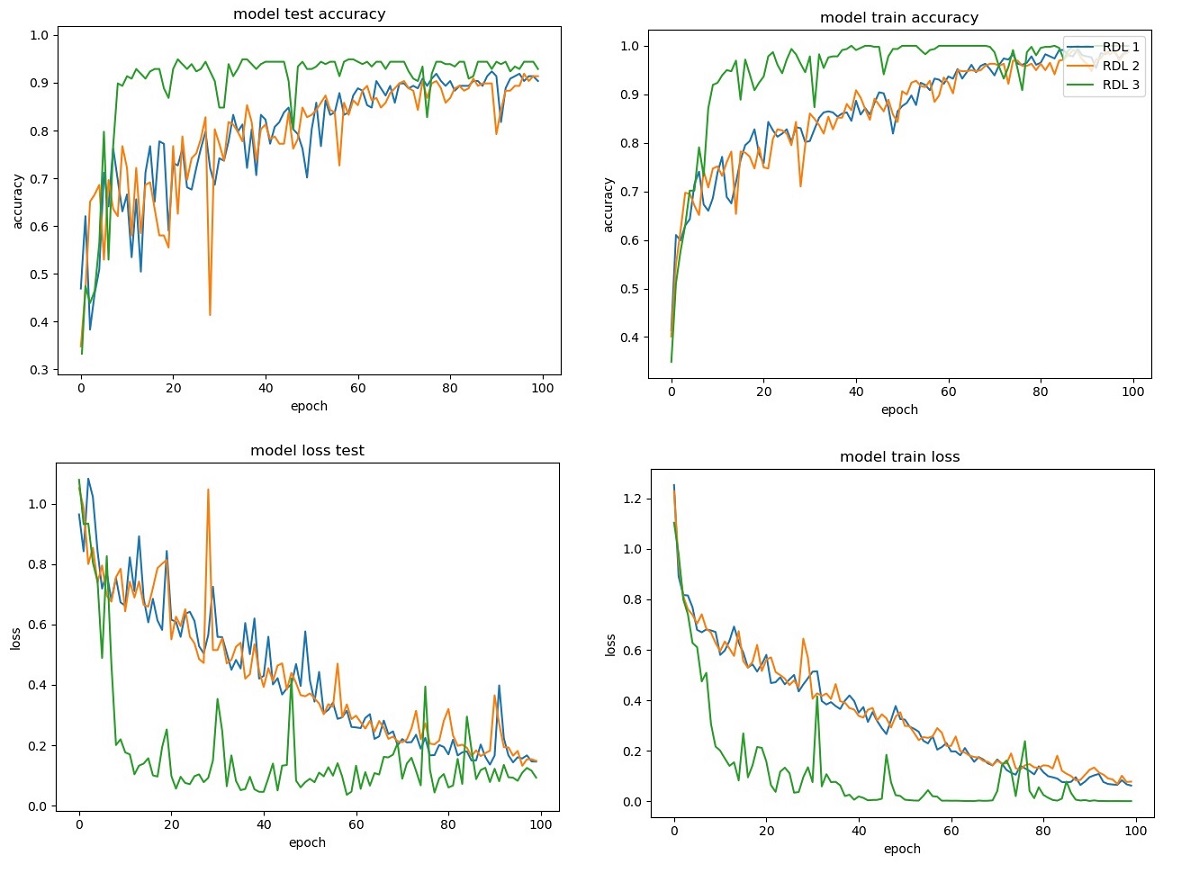}
    \caption{This Figure indicates loss function and accuracy for 3 Random Deep Learning~(RDL) model for 100 epochs.}
    \label{fig:RMDL_3_4}
\end{figure}

As shown in Table~\ref{ta:RMDL_Baseline}, the overall results is represent that if the more RDLs is used, that accuracy, precision, recall and F1-score are improved. We used three baseline which are CNN  that is shallow CNN is used with 3 Convolutional Layer with kernel size of $3\times3$ and 64 filter. this CNN contains Maxpooling with size of $2\time2$ and dropout of 0.25. The other Baseline we used is Deep neural network or Deep multi-layer perceptron algorithm which is very popular these days. In a Multilayer Perceptron, 4 hidden layers are used which each layer contains 1000 nodes. And third Baseline is DCNN which contains 16 Convolutional Layer with kernel size of $3\times3$ and 64 filter it has only 3 Maxpooling layer with size of $2\time2$  and used dropout of 0.25. The result shown as for a Multilayer Perceptron as  Accuracy is  75.74$\pm$ 0.79 Precision is equal to 88.67$\pm$0.58 Recall is 83.85$\pm$0.68 and F1-Score is 86.19$\pm$0.64. For CNN 
Accuracy is  83.01$\pm$0.69  Precision is equal to 91.62$\pm$0.51 Recall is 89.83$\pm$0.56 F1-Score is 90.72$\pm$0.53. The result of DCNN is as follows Accuracy is  89.71$\pm$0.56 Precision is equal to 96.44$\pm$0.34 Recall is 92.79$\pm$0.48 and finally F1-Score is 94.58$\pm$0.42. The interesting part of this study shows that 3 RDLs in this experiment is not good as DCNN; Thus, if we want to get a robust results we should have more than 3 RDLs, especially, when we used random deep learning architectures. 

The RMDL results with using three RDLs   Accuracy is 86.62$\pm$0.63 Precision is equal to 93.01$\pm$0.47 Recall is 92.66$\pm$0.48 F1-Score is 92.83$\pm$0.47 . The RMDL using 9 RDLs is as follows  Accuracy is 91.12$\pm$0.52  Precision is equal to 95.08$\pm$0.40 Recall is 95.63$\pm$0.37 F1-Score is 95.35$\pm$0.39 and finally, the results of RMDL with 15 RDLs is as follows: Accuracy is 91.61$\pm$0.51 Precision is equal to 95.85$\pm$0.37 Recall is 95.40$\pm$0.39 F1-Score is 95.62$\pm$0.38 .
\begin{table}[]
\centering
\caption{Results of RMDL in comparison with our baselines}\label{ta:RMDL_Baseline}
\begin{tabular}{|c|c c c c c|}
\hline
\multicolumn{2}{|c}{Model}      & Accuracy & Precision & Recall & F1-Score \\ \hline
\multirow{3}{*}{Baseline} & DNN  &     75.74$\pm$ 0.79    &     88.67$\pm$0.58       &   83.85$\pm$0.68      &     86.19$\pm$0.64      \\ \cline{2-6} 
                          & CNN  &     83.01$\pm$0.69      &    91.62$\pm$0.51        &  89.83$\pm$0.56       &    90.72$\pm$0.53     \\ \cline{2-6} 
                          & DCNN &     89.71$\pm$0.56      &    96.44$\pm$0.34        &  92.79$\pm$0.48       &    94.58$\pm$0.42      \\ \hline
\multirow{3}{*}{RMDL}     & 3 RDLs    &     86.62$\pm$0.63      &    93.01$\pm$0.47        &   92.66$\pm$0.48      &    92.83$\pm$0.47       \\ \cline{2-6} 
                          & 9 RDLs   &     91.12$\pm$0.52      &    95.08$\pm$0.40        &   95.63$\pm$0.37      &    95.35$\pm$0.39       \\ \cline{2-6} 
                          & 15 RDLs  &     91.61$\pm$0.51      &    95.85$\pm$0.37        &   95.40$\pm$0.39      &    95.62$\pm$0.38       \\ \hline
\end{tabular}
\end{table}

\begin{figure}[H]
    \centering
    \includegraphics[width=\textwidth]{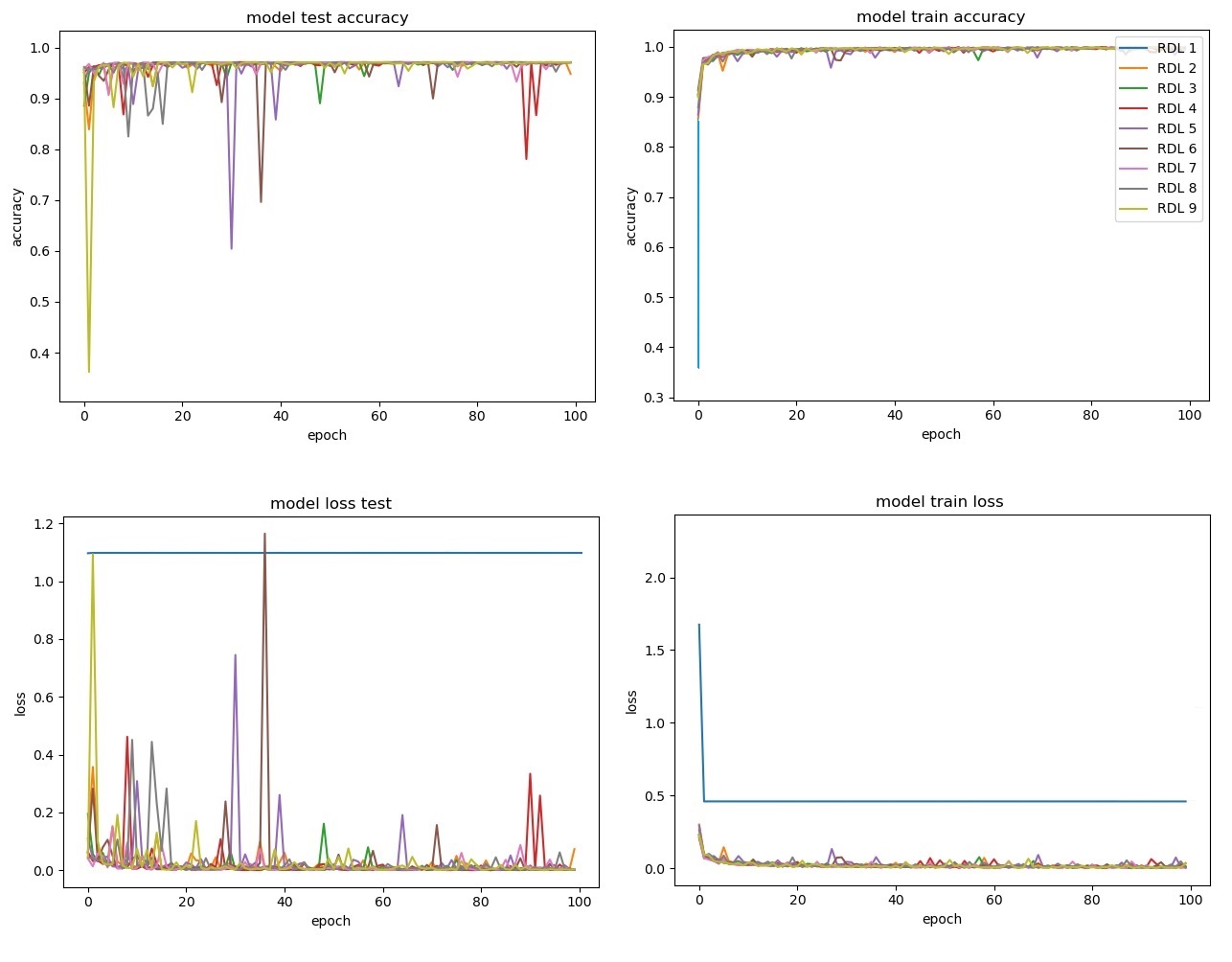}
    \caption{This Figure indicates loss function and accuracy for 9 Random Deep Learning~(RDL) model for 100 epoch}
    \label{fig:RMDL_9_4}
\end{figure}

\begin{figure}[H]
    \centering
    \includegraphics[width=\textwidth]{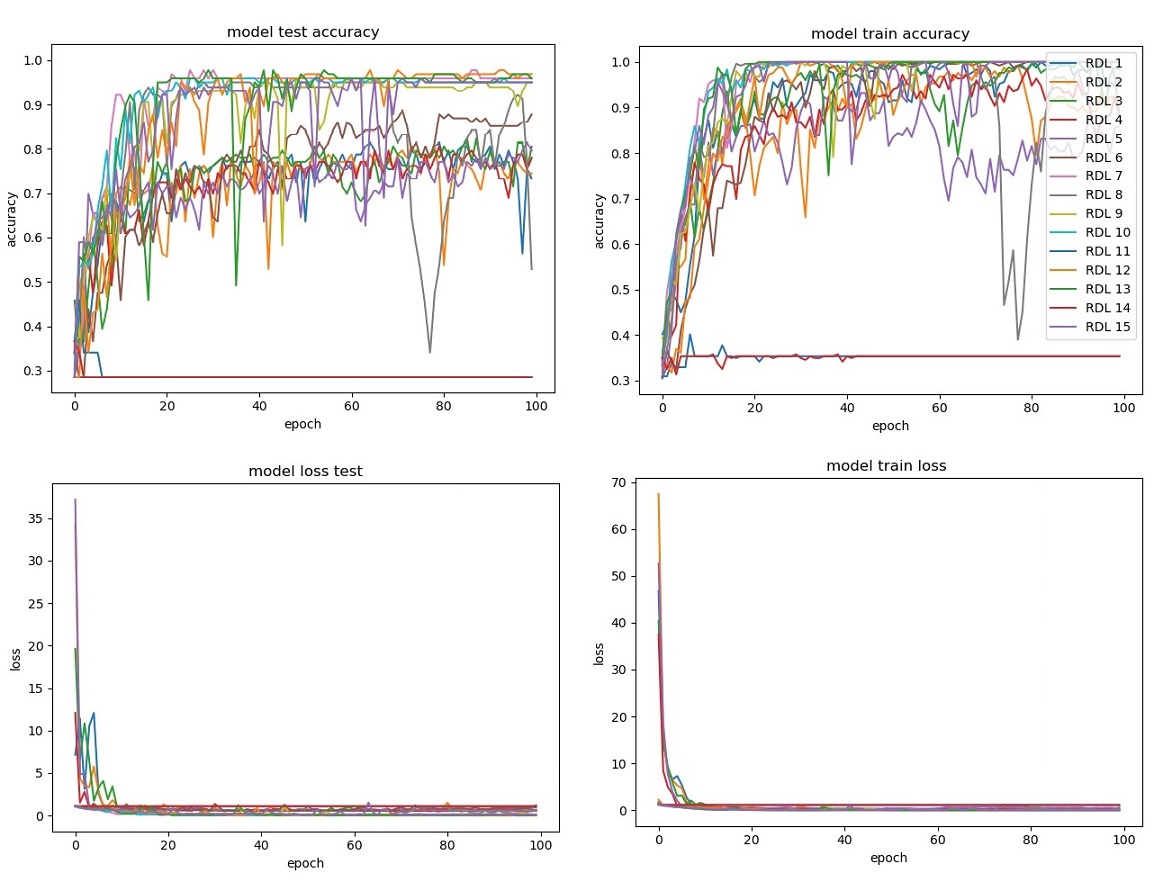}
    \caption{This Figure indicates loss function and accuracy for 15 Random Deep Learning~(RDL) model for 100 epoch}
    \label{fig:RMDL_15_4}
\end{figure}

Figure~\ref{fig:RMDL_3_4} indicates loss function and accuracy for 3 Random Deep Learning~(RDL) model for 100 epochs which as it clear one of the RDL is converged much faster that two other models. Figure~\ref{fig:RMDL_9_4} indicates loss function and accuracy for 9 Random Deep Learning~(RDL) model for 100 epochs which as it clear two of the RDL is  never converged but due to overall results is based on majority votes this RDLs not effect on the final results. Figure~\ref{fig:RMDL_15_4} indicates loss function and accuracy for 15 Random Deep Learning~(RDL) model for 100 epochs which as the 2 RDL are converged but their accuracy about 30 percent and one model after 70 epochs the accuracy is drooped to less than 40 percent and it backs to around 80 percent; but finally these models are not effect to overall results is based on majority votes this RDLs not effect on the final results.

\subsection{Hardware and Framework}
The processing units that has been used through this experiment was intel on \textit{Xeon~E5-2640~ (2.6 GHz)} with~\textit{12 cores} and \textit{64~GB} memory~(DDR3). Also, graphical card on our machine is \textit{Nvidia Quadro~K620} and \textit{Nvidia Tesla~K20c}. This work is implemented in Python using Compute Unified Device Architecture~(CUDA) which is a parallel computing platform and Application Programming Interface ~(API) model created by $Nvidia$. We used $TensorFelow$~\cite{abadi2016tensorflow} and $Keras$~\cite{chollet2015keras} library for creating the neural networks. 

\section{Conclusion}

The classification task is an important problem to address in machine learning, given the growing number and size of datasets that need sophisticated classification.  We propose a novel technique to solve the problem of choosing best technique and method out of many possible structures and architectures in deep learning. This paper introduces a new approach called RMDL~(Random Multimodel Deep Learning) for the medical image classification that combines multi deep learning approaches to produce random classification models. Our evaluation on datasets obtained from the biopsy images shows that combinations of Multilayer Perceptrons, RNNs and CNNs with the parallel learning architecture, has consistently higher accuracy. These results show that deep learning methods can provide improvements for classification and that they provide flexibility to classify datasets by using majority vote. The proposed approach has the ability to improve accuracy and efficiency of models and can be use across a wide range of other data types and applications.

\chapter{Celiac Disease Severity Diagnosis on Duodenal Histopathological Images Using Deep Learning}\label{chpt:CeliacNet}

Celiac Disease (CD) is a chronic autoimmune disease that affects the small intestine in genetically predisposed children and adults. Gluten exposure triggers an inflammatory cascade which leads to compromised intestinal barrier function. If this enteropathy is unrecognized, this can lead to anemia, decreased bone density, and, in longstanding cases, intestinal cancer.  The prevalence of the disorder is \textbf{$1\%$} in the United States. An intestinal (duodenal) biopsy is considered the “gold standard” for diagnosis. The mild CD might go unnoticed due to non-specific clinical symptoms or mild histologic features. In our current work, we trained two model based on shallow convolutional neural network~(CNN) and deep residual networks to diagnose CD severity using  a histological scoring system called the modified Marsh score. The proposed model was evaluated using an independent set of \textbf{$120$} whole slide images from~\textbf{$15$}~CD patients and achieved an AUC greater than~$0.96$ for ResNet and  in all classes. These results demonstrate the capability of the proposed model for CD severity classification using histopathological images.

\section{Introduction}

Celiac disease (CD) is an inability to normally process dietary gluten (present in foods such as wheat, rye, and barley) and is present in $1$\% of the US population. Gluten consumption by people with CD can cause diarrhea, abdominal pain, bloating, and weight loss. If unrecognized, it can lead to anemia, decreased bone density, and, in longstanding cases, intestinal cancer~\cite{fasano2003prevalence, parzanese2017celiac}. An intestinal (duodenal) biopsy, obtained via endoscopic evaluation, is considered the “gold standard” for diagnosis of CD.  Due to unclear clinical symptoms and/or obscure histopathological features~(based on biopsy images),~CD is often undiagnosed~\cite{corazza2007comparison}. There has been major clinical interest towards developing new and innovative methods to automate and enhance the detection of morphological features of CD on biopsy images.

Among various architectures of CNNs, Residual Networks~(ResNet) have received special attention due to their considerably superior performance in the analysis of histopathological images for disease detection, diagnosis and prognosis prediction to complement the opinion of a human pathologist. Multiple groups have published on the use of the ResNet architecture for classification of Hematoxylin and Eosin~(H\&E) stained biopsy images including breast and prostate cancer~\cite{gandomkar2018mudern,rakhlin2018deep,motlagh2018breast,chougrad2018deep,schaumberg2018h} and colorectal polyps~\cite{korbar2017deep}. Similarly, impressive results for CD diagnosis based on whole slide biopsy images have been noted in published literature~\cite{wei2019automated}. Herein we explore the performance of deep residual networks in severity diagnosis of CD on duodenal biopsy images. 

This dissertation section is organized as follows: In Section~\ref{sec:severitytypes}, disease severity classes of CD are presented. In Section~\ref{sec:Data_Source_5}, we describe the  data used in this study. Section~\ref{sec:Pre-Processing_5} presents the data pre-processing steps. The methodology is explained in Section~\ref{sec:Method_5}. Empirical results are elaborated in Section~\ref{sec:Empirical_Results_5}. Finally, Section~\ref{sec:Conclusion_5} concludes the dissertation section along with outlining future directions.

\section{Severity Classes of Celiac Disease}\label{sec:severitytypes}
\begin{figure}[!t]
    \centering
    \includegraphics[width=\columnwidth]{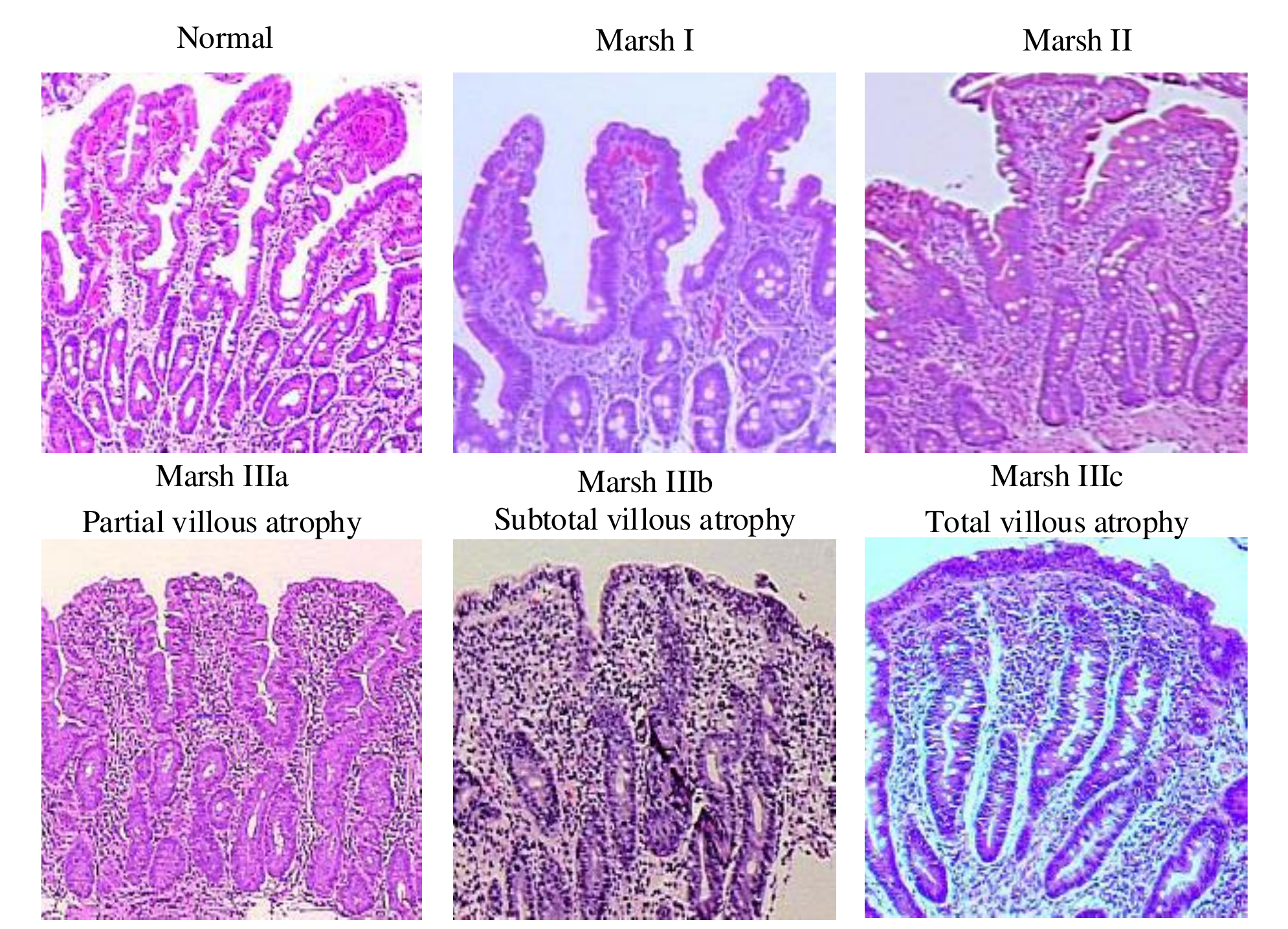}
    \caption{CD severity classification based on modified Marsh score~\cite{fasano2001current}} \label{fig:MarshScore}
\end{figure}

~Modified Marsh Score Classification was developed to classify the severity of CD based on microscopic histological morphological features~(Figure~\ref{fig:MarshScore}). It takes into account the architecture of the duodenum as having finger-like projections~(called \qu{villi}) which are lined by cells called epithelial cells. Between the villi are crevices called crypts that contain regenerating epithelial cells. 
The normal ratio of the length of a typical healthy villus to the depth of a representative health crypt  should be between 3:1 and 5:1. In the normal, healthy duodenum (first part of the small intestine), there should be no more than 30 immune cells known as lymphocytes interspersed per 100 epithelial cells in the top layer of the villus. Marsh~I histology comprises of normal villus architecture with an increase in the number of intraepithelial lymphocytes. Marsh~II includes increased intraepithelial lymphocytes along with a finding known as crypt hypertrophy in which the crypts appear enlarged. This is usually rare since patients typically rapidly progress from Marsh I to IIIa. Marsh~III is sub-divided into~IIIa~(partial villus atrophy), Marsh IIIb~(subtotal villus atrophy) and Marsh IIIc~(total villus atrophy) to explain the spectrum of villus atrophy along with crypt hypertrophy and increased intra-epithelial lymphocytes. Finally, in Marsh~IV, villi are completely atrophied. This is called “hypoplastic” or complete villus atrophy and describes the microscopic histology of duodenal tissue from patients at the extreme end of gluten sensitivity.

\section{Data Source}\label{sec:Data_Source_5} 
~$162$ H\&E stained duodenal biopsy slides were obtained from the archival biopsies of $34$ CD patients from the University of Virginia~(UVa) in Charlottesville, VA, United States. Each slide contained multiple biopsies per patient resulting in $336$ whole slide images at 40x magnification using the Leica SCN 400 slide scanner~(Meyer Instruments, Houston, TX) at the Biorepository and Tissue Research Facility at UVa. Characteristics of our patient population were as follows: the median $(Q1, Q3)$ age was $130~(92.5, 175.5)$ months. we had a roughly equal  distribution  of males  $(47.1\% ,n=16)$  and  females  $(52.9\%, n=18)$. Biopsy images for our study population were scored by two medical professionals and validated with reads from a pathologist specialized in gastroenterology. Our biopsy image dataset ranged from Marsh I to IIIc with no biopsy images present in Marsh II.

\section{Data Pre-processing}\label{sec:Pre-Processing_5}~ Since whole slide images (WSIs) were digitized at high resolutions, these were large files with notable color variability apparent on visual inspection. Therefore, we pre-processed these before any computational analyses were conducted. This section describes all pre-processing steps including image patching, patch clustering and color normalization. 
\subsection{Image Patching}\label{subsec:Patching}~The effectiveness of CNNs in image classification has been shown in various studies across different domains~\cite{kowsari2018rmdl,hu2018deep,Heidarysafa2018RMDL}. However, the training of a CNN on high resolution WSIs that are at a gigapixel level is not often feasible due to high computational cost. Also, the application of CNNs on WSIs further contributes to the loss of a large part of discriminatory information due to extensive down-sampling which is needed in such images~\cite{hou2016patch}. We hypothesized that since there were cellular level morphological differences between different CD severity classes given the spectrum of pathology, a trained classifier on image patches would likely perform as well or better than a trained WSI-level classifier.
A sliding window method was applied to each high-resolution WSI to generate patches of size $500\times500$ pixels with~$50\%$ overlapping area. After generating patches from each image, we labelled each patch based on its associated image.

\begin{figure}[!b]
    \centering
    \includegraphics[width=\columnwidth]{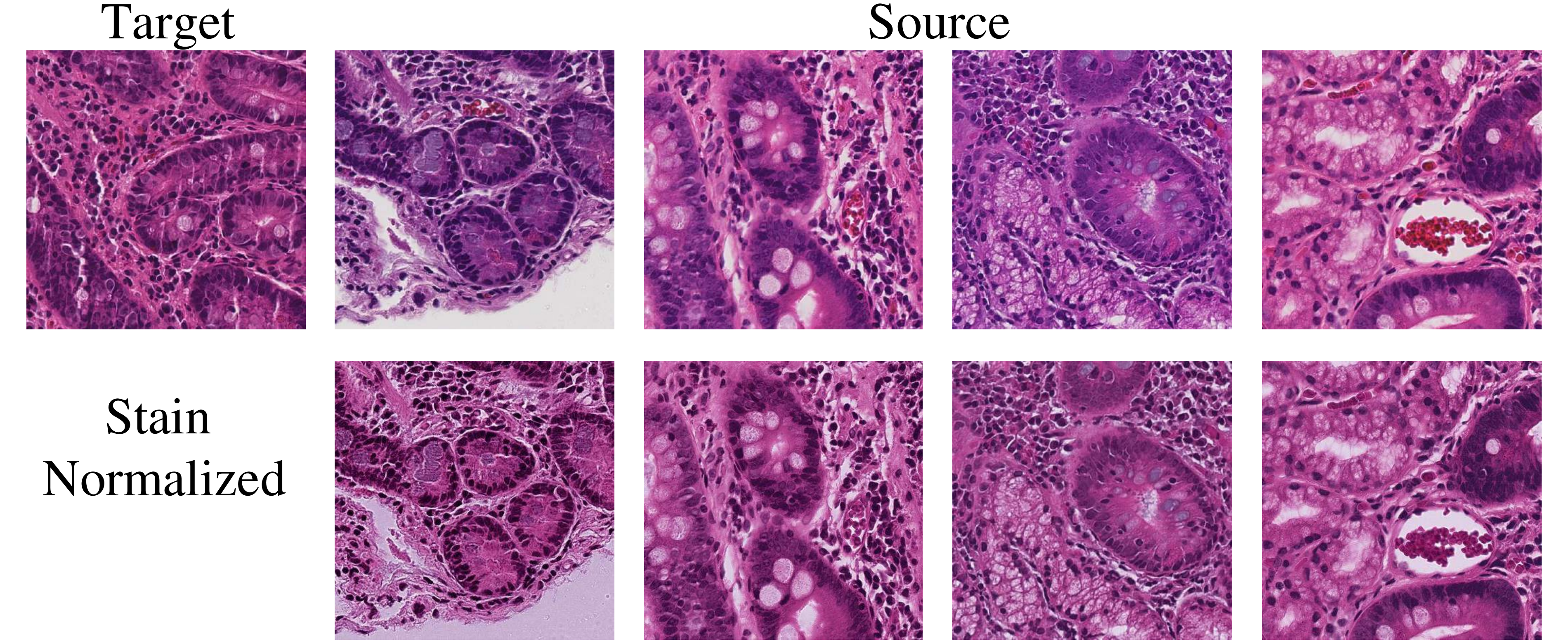}
    \caption{Color normalization artifacts when using the method proposed by Vahadane et al.~\cite{vahadane2016structure}. Images in the first row represent the target image and some source images. Their associated normalized images are in second row} \label{fig:StainNormalization}
\end{figure}

\subsection{Patch Clustering}\label{subsec:Clustering_5}~Clustering is organizing objects in a such way that objects within a group or cluster in some way are more similar to each other compared to objects in other groups. There is a wide variety of algorithms for data clustering and K-means clustering is one of the easiest ones~\cite{jain2010data}. Finding the optimal solution to the k-means clustering problem is NP-hard in general Euclidean space even for 2 clusters. Clustering of $n$ $d$-dimension entities in $k$ clusters can be exactly solved in time of $O(n^{dk+1})$~\cite{aloise2009np}. Obviously, reduction of dimension $d$ will result in significant improvement of the K-means clustering algorithm in term of time complexity. To address the problem of dimensionality reduction, a convolutional auto-encoder~\cite{goodfellow2016deep} was used to learn embedded features of each patch. These auto-encoders have been reported in the literature as having had great success as a dimensionality reduction method via the powerful reprehensibility of neural networks~\cite{wang2014generalized}.

In our work, a two-step clustering process was applied to identify useless patches which had mostly been created from the background of the WSIs. All or a large part of these patches were blank or did not  contain any useful biopsy information. Through the first step, a convolutional autoencoder was used to learn the embedded features of each patch and in the second step k-means clustering algorithm was applied to cluster embedded features into two clusters: useful and not useful. Some results of patch clustering have been shown in Figure~\ref{fig_Clustering}.

\subsection{Stain Normalization}\label{subsec:StainNormalization}

Histological images have substantial color variation that adds bias while training the model~\cite{sali2019celiacnet}. This arises due to a wide variety of factors such as differences in raw materials and manufacturing techniques of stain vendors, staining protocols of labs, and color responses of digital scanners~\cite{vahadane2016structure,sali2019celiacnet}. To avoid any bias, unwanted color variations are neutralized by conducting color normalization as an essential pre-processing step prior to any analyses. Various color normalization approaches have been proposed in the published literature. In this study, we used the approach proposed by Vahadane et al.~\cite{vahadane2016structure} for child level due to that all CD images are collected from one center. This approach preserves biological structure information by basing color mixture modeling on sparse non-negative matrix factorization. Figure~\ref{fig:StainNormalization} shows an example of the result of applying this technique on representative biopsy patches.

\section{Methodology}\label{sec:Method_5}
\subsection{Model Development}\label{subsec:Model}
\begin{figure}[!t]
    \centering
    \includegraphics[width=\columnwidth]{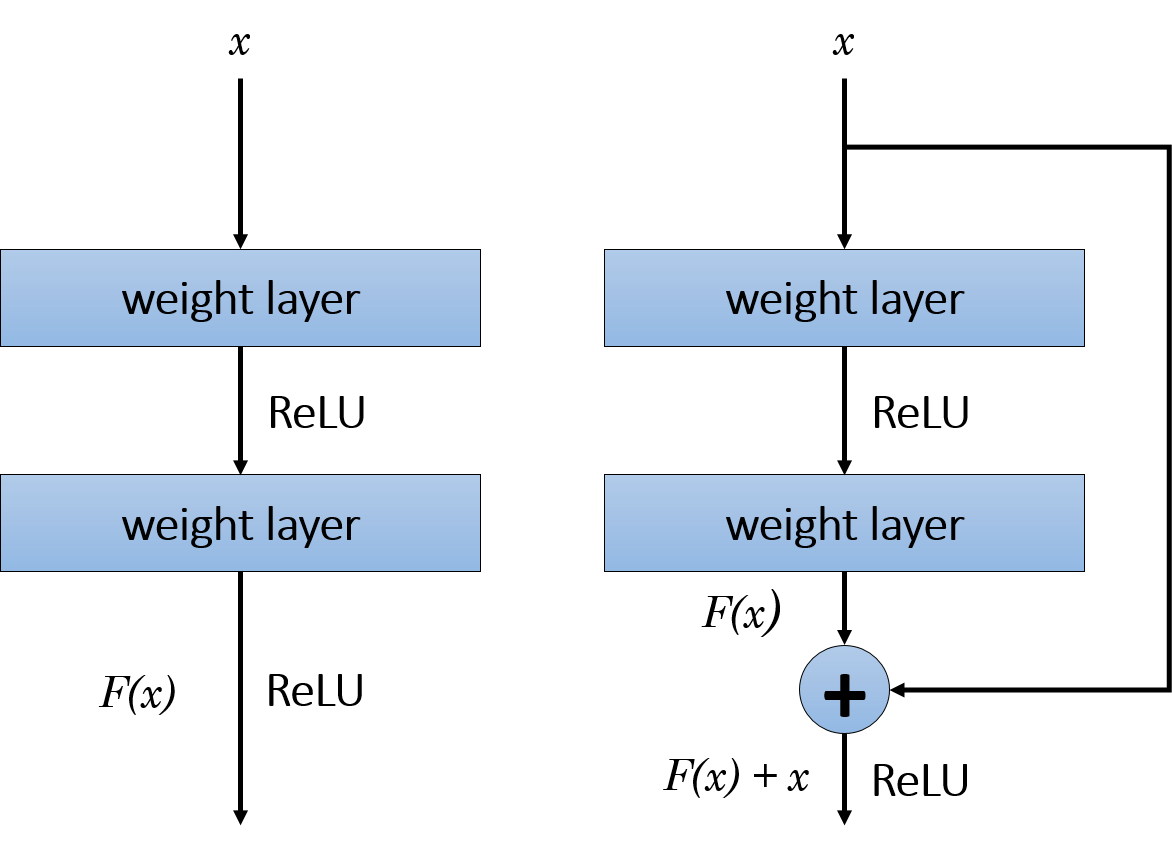}
    \caption{Building blocks of~(left) a traditional CNN,~(right) a ResNet } \label{fig:buildin blocks}
\end{figure}

~ CNNs have demonstrated promising performance in image classification tasks. There are many different architectures of CNNs in the literature, with associated advantages and drawbacks. In the current study, we used the deep residual network~(ResNet)\cite{he2016deep}, a model which has shown great performance in image classification problems including medical image analysis\cite{gandomkar2018mudern,wen2018deep}. Although it has been shown that CNNs with more convolutional layers achieve the most accurate results, simply stacking more convolutional layers will not lead to better performance. When the deep network reaches a certain depth, its performance tends to be saturated and even begins to rapidly decline. In such cases, the models involve a large number of parameters and are computationally expensive to train through whole parameters. This is called the degradation problem and ResNet was originally proposed to tackle this issue. The core idea of ResNet is introducing a skip connection that skips one or more layers and bypasses the input from the previous layer to the next layer without any modification. Since these added shortcut connections perform identity mapping, extra parameters are not added to the model. Such architecture enables deployment of deeper networks without problem of degeneracy. The building block of the ResNet is compared to the building block of the traditional network in Figure \ref{fig:buildin blocks}. In the traditional networks, the mapping from input to output can be represented by the nonlinear function $H(x)$. In residual learning blocks, $F(x) = H(x) - x$ is used as mapping function~\cite{he2016deep}. In essence, as part of traditional CNNs, the input $x$ is mapped to $F(x)$ which is a completely new representation that does not keep any information about the original input, while ResNet blocks compute a slight change to the original input $x$ to get a slightly altered representation. ResNet was the Winner of the~ILSVRC~2015 in image classification, detection, and localization, as well as the Winner of the MS COCO~2015 detection and segmentation.

\begin{table}[!t]

\caption{Architecture of the model}\label{tb:model}
\centering
\begin{tabular}{cccc}
\hline
\textbf{Class} & \textbf{Layer Type}  & \textbf{Output Shape}  & \textbf{Number of Parameters}  \\ \hline

 1 & Model& $(7,7,2048)$  & $2,3587,712$ \\
 
 2 & Flatten & $100352$  & $0$ \\
 
 3 & Dense & $1024$  & $102,761,472$ \\
 
 4 & Dropout & $1024$  & $0$ \\
 
 5 & Dense & $4$  & $4,100$ \\\hline
\end{tabular}
\end{table}

Different variants of ResNet models such as ResNet50, ResNet101, and ResNet152 were trained on the ImageNet dataset~\cite{guillaumin2012large}. We customized the Resnet50 by removing fully connected layers and keeping only the ResNet backbone as a feature extractor. Then we added one fully connected layer with $1024$ neurons that received the flattened output of the feature extractor. Finally, the output layer was added such that it represented a  prediction  probability  for  each  of  the  four Marsh score categories: I, IIIa, IIIb and IIIc. We used dropout on the fully-connected layers with $p=0.5$ as the regulizer. This model has been summarized in Table~\ref{tb:model}.

\begin{table}[!t]

\caption{Patch-level performance of model for celiac disease severity diagnosis Based on the two Deep learning techniques which are ResNet and CNN. The model performances for I and IIIa ResNet is more robust but for IIIb and IIIc}\label{table:patch-level_results}
\centering
\begin{tabular}{|c| c|  c  c c |}
\hline
\multicolumn{2}{|c|}{}                                                                                                  & Precision (\%)                                                   &Recall~(\%)                                                    & F1-measure (\%)                                               \\ \hline
\multirow{8}{*}{ResNet } & \begin{tabular}[c]{@{}c@{}}I\\  (n = 6988)\end{tabular}    & \begin{tabular}[c]{@{}c@{}}93.30 \\ (92.71, 93.89)\end{tabular}  & \begin{tabular}[c]{@{}c@{}}89.54 \\ (88.82, 90.26)\end{tabular} & \begin{tabular}[c]{@{}c@{}}91.38 \\ (90.72, 92.04)\end{tabular}  \\ \cline{2-5} 
                        & \begin{tabular}[c]{@{}c@{}}IIIa \\ (n = 6615)\end{tabular} & \begin{tabular}[c]{@{}c@{}}94.16 \\ (93.59, 94.73)\end{tabular}  & \begin{tabular}[c]{@{}c@{}}84.75 \\ (83.88, 85.62)\end{tabular} & \begin{tabular}[c]{@{}c@{}}89.20 \\ (88.45, 89.95)\end{tabular}  \\ \cline{2-5} 
                        & \begin{tabular}[c]{@{}c@{}}IIIb \\ (n = 7695)\end{tabular}  & \begin{tabular}[c]{@{}c@{}}83.94\\  (83.12, 84.76)\end{tabular}  & \begin{tabular}[c]{@{}c@{}}89.45 \\ (88.76, 90.14)\end{tabular} & \begin{tabular}[c]{@{}c@{}}86.61 \\ (85.85, 87.37)\end{tabular}  \\ \cline{2-5} 
                        & \begin{tabular}[c]{@{}c@{}}IIIc \\ (n = 7369)\end{tabular}   & \begin{tabular}[c]{@{}c@{}}85.53 \\ (84.73, 86.33)\end{tabular}  & \begin{tabular}[c]{@{}c@{}}90.61 \\ (89.94, 91.28)\end{tabular} & \begin{tabular}[c]{@{}c@{}}87.99 \\ (87.25, 88.73)\end{tabular}  \\ \hline
\multirow{8}{*}{CNN}    & \begin{tabular}[c]{@{}c@{}}I \\ (n = 2137)\end{tabular}    & \begin{tabular}[c]{@{}c@{}}88.73\\  (88.10 , 89.36)\end{tabular} & \begin{tabular}[c]{@{}c@{}}85.07 \\ (84.36, 85.78)\end{tabular} & \begin{tabular}[c]{@{}c@{}}86.86 \\ (86.19, 87.53 )\end{tabular} \\ \cline{2-5} 
                        & \begin{tabular}[c]{@{}c@{}}IIIa \\ (n = 2052)\end{tabular}  & \begin{tabular}[c]{@{}c@{}}81.19\\ (80.37, 82.01)\end{tabular}  & \begin{tabular}[c]{@{}c@{}}83.72 \\ (82.95, 84.49)\end{tabular} & \begin{tabular}[c]{@{}c@{}}82.44 \\ (81.64, 83.23)\end{tabular}  \\ \cline{2-5} 
                        & \begin{tabular}[c]{@{}c@{}}IIIb \\ (n = 2436)\end{tabular}  & \begin{tabular}[c]{@{}c@{}}90.51 \\ (90.18, 90.81)\end{tabular}  & \begin{tabular}[c]{@{}c@{}}90.48 \\ (89.31, 91.64)\end{tabular} & \begin{tabular}[c]{@{}c@{}}90.49\\  (89.25, 91.74)\end{tabular}  \\ \cline{2-5} 
                        & \begin{tabular}[c]{@{}c@{}}IIIc \\ (n = 2433)\end{tabular}   & \begin{tabular}[c]{@{}c@{}}89.26 \\ (88.03, 90.49)\end{tabular}  & \begin{tabular}[c]{@{}c@{}}90.18 \\ (88.99, 91.36)\end{tabular} & \begin{tabular}[c]{@{}c@{}}89.72 \\ (88.51, 90.92)\end{tabular}  \\ \hline
\end{tabular}
\end{table}

\begin{figure}[!t]
    \centering
    \includegraphics[width=\columnwidth]{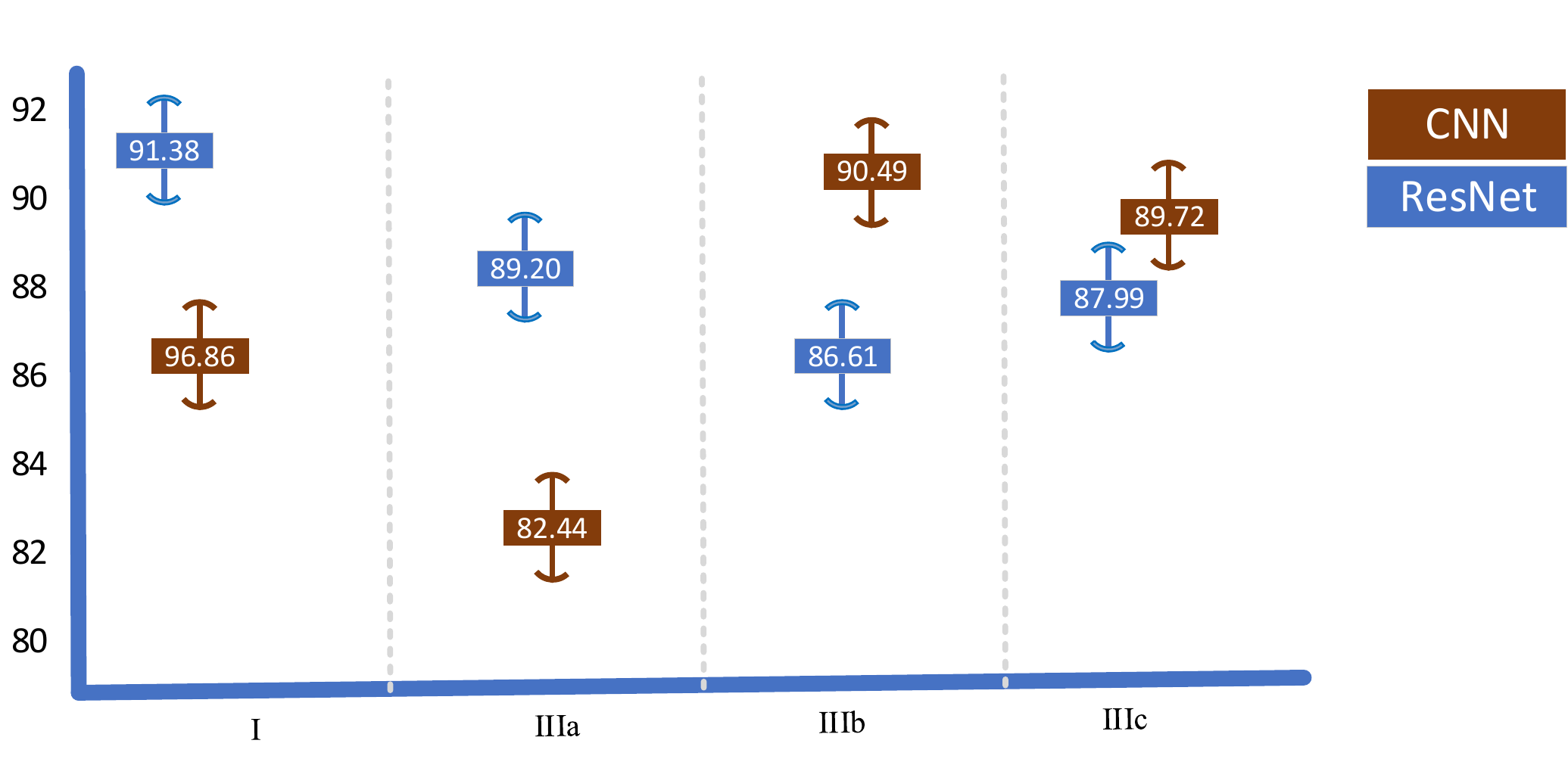}
    \caption{Comparison of F1-score results for four classes between ResNet and CNN\textbf{}} \label{fig:t-test}
\end{figure}

We resized pre-processed patches into~$224\times224$ pixels and used them to train our model. Both horizontal and vertical random rotations were performed as part of our data augmentation. The model was trained on around~$50,000$ patches for each of four classes. Optimization was performed using RMSprop optimization with
no momentum, a base learning rate of \num{1e-5} and a multiclass cross entropy loss function. As shown in Table~\ref{table:patch-level_results}, for classes I and IIIa, ResNet with mean of F1-measure of 91.38 and 89.20 is more accurate than CNN. But as regards to IIIb and IIIc classes, CNN is more robust with mean F1-measure of 90.49 and 89.72. The overall F1-measure of CNN is 87.37 and ResNet is 88.59. on T -test~\cite{gad1982statistics} which the The overall F1-measure of ResNet with 1.22 is higher than CNN with standard error of 0.294, and $p=0.00134$. As shown in Figure~\ref{fig:t-test}, the result of ResNet for class I and IIIa is absolutely better than CNN. The result of CNN in class IIIb is better than ResNet. The result for class IIIc shows that none of these models is absolutely better than the other but the mean of CNN model is higher than ResNet.

\subsection{Whole slide classification}\label{sec:WSIClassification}
\begin{figure}[!b]
    \centering
    \includegraphics[width=\columnwidth]{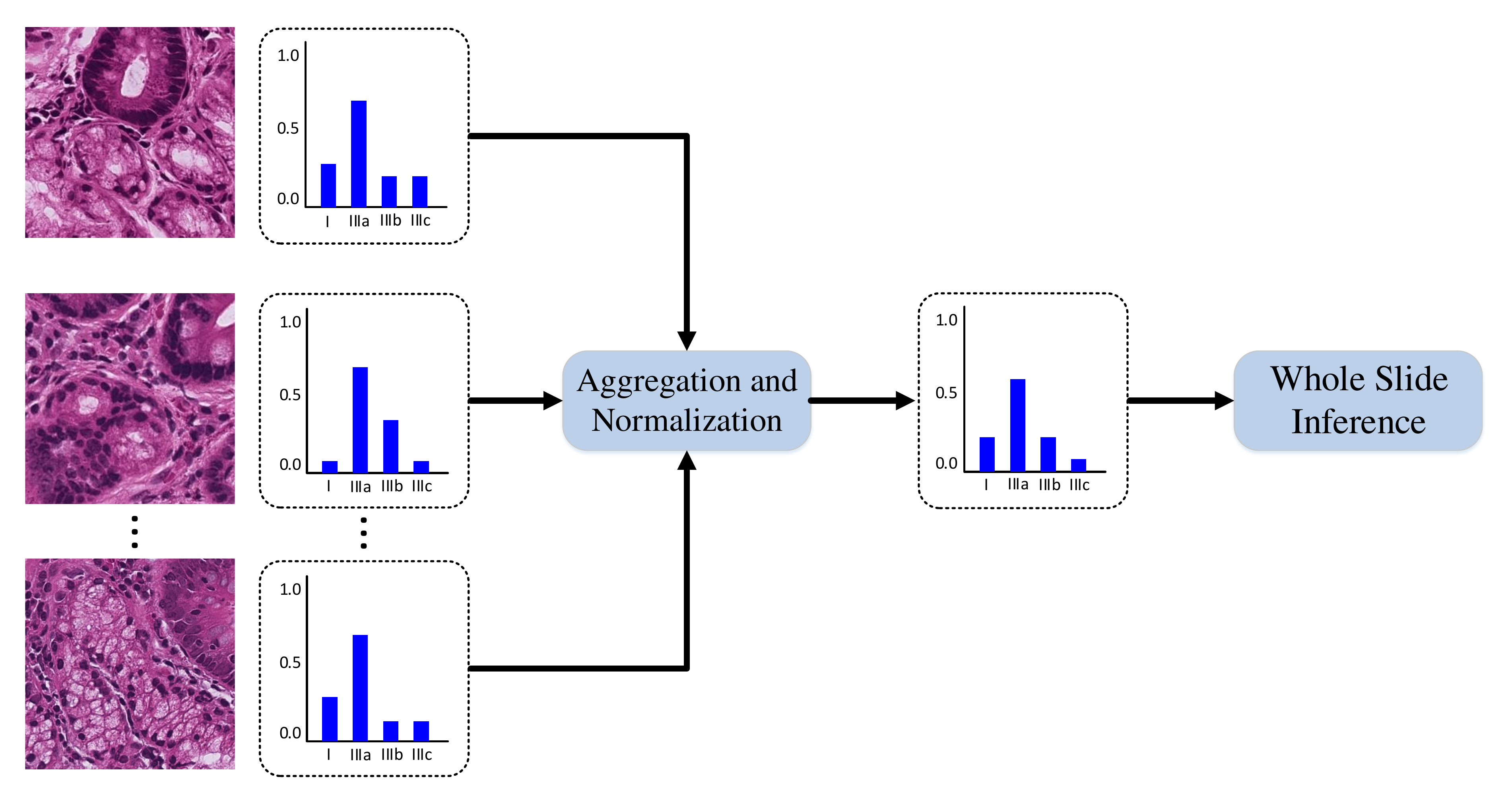}
    \caption{Overview of whole-slide inference process using aggregation of patch-level classifications} \label{fig:WSIClassification}
\end{figure}

~Our goal was to classify WSIs based on severity assessed via the modified Marsh score. The model used was trained to classify small patches rather than WSIs. To achieve this goal, a heuristic method was developed which aggregated crop classifications and translated them to whole-slide inferences. Each WSI in the test set was initially patched, those patches which did not contain any information were filtered out and finally stain normalization was performed. After these pre-processing steps our trained model was applied with the goal of image classification. We denoted the probability distribution over possible labels, given the crop $x$ and training set~$D$ by~$p(y\vert x,D)$. In general, this represented a vector of length $C$, where $C$ is number of classes. In our notation, the probability is conditional on the test patch $x$, as well as the training set~$D$. For each crop, the model gives an output of a vector composed of four components showing probabilities for each one of the four classes of CD severity. Given a probabilistic output, the patch~$j$ in slide $i$ is assigned to the most probable class label~$\hat{y}_{ij}$ which is shown in Equation~\ref{eq:patchClass}.

\begin{equation}\label{eq:patchClass}
\hat{y}_{ij} = \argmax_{c \in \{1,2,3,...,C\}} p(y_{ij} = c\vert x_{ij},D)
\end{equation}

where $\hat{y}$ is called maximum a posteriori~(MAP) . Summation over these vectors and normalizing the resultant vector, created a vector that had components showing the probability of CD severity for the associated WSI. Equation~\ref{eq:slideClass}, shows how the class of WSI was predicted. 

\begin{equation}\label{eq:slideClass}
\hat{y}_i = \argmax_{c \in \{1,2,3,...,C\}} \sum_{j=1}^{N_i}p(y_{ij} = c\vert x_{ij},D)
\end{equation}

where $N_i$ is number of patches in slide $i$. Figure~\ref{fig:WSIClassification} depicts overview of the whole-slide inference process.

\begin{figure}[!b]
    \centering
    \includegraphics[width=\columnwidth]{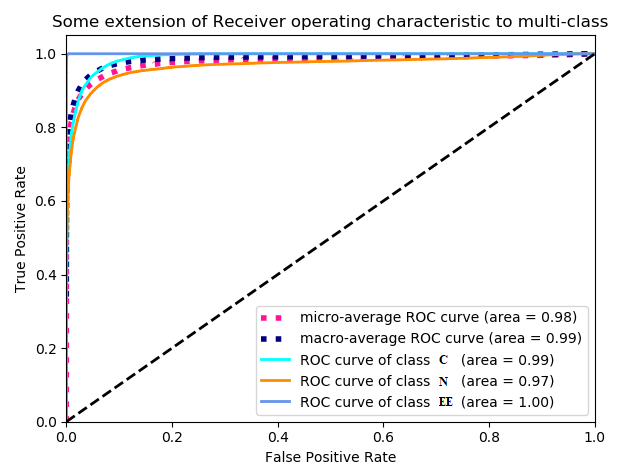}
    \caption{Patch-level ROC and AUC for different classes} \label{fig:ROC}
\end{figure}

\begin{figure*}[!htb]
    \centering
    \includegraphics[width=\textwidth]{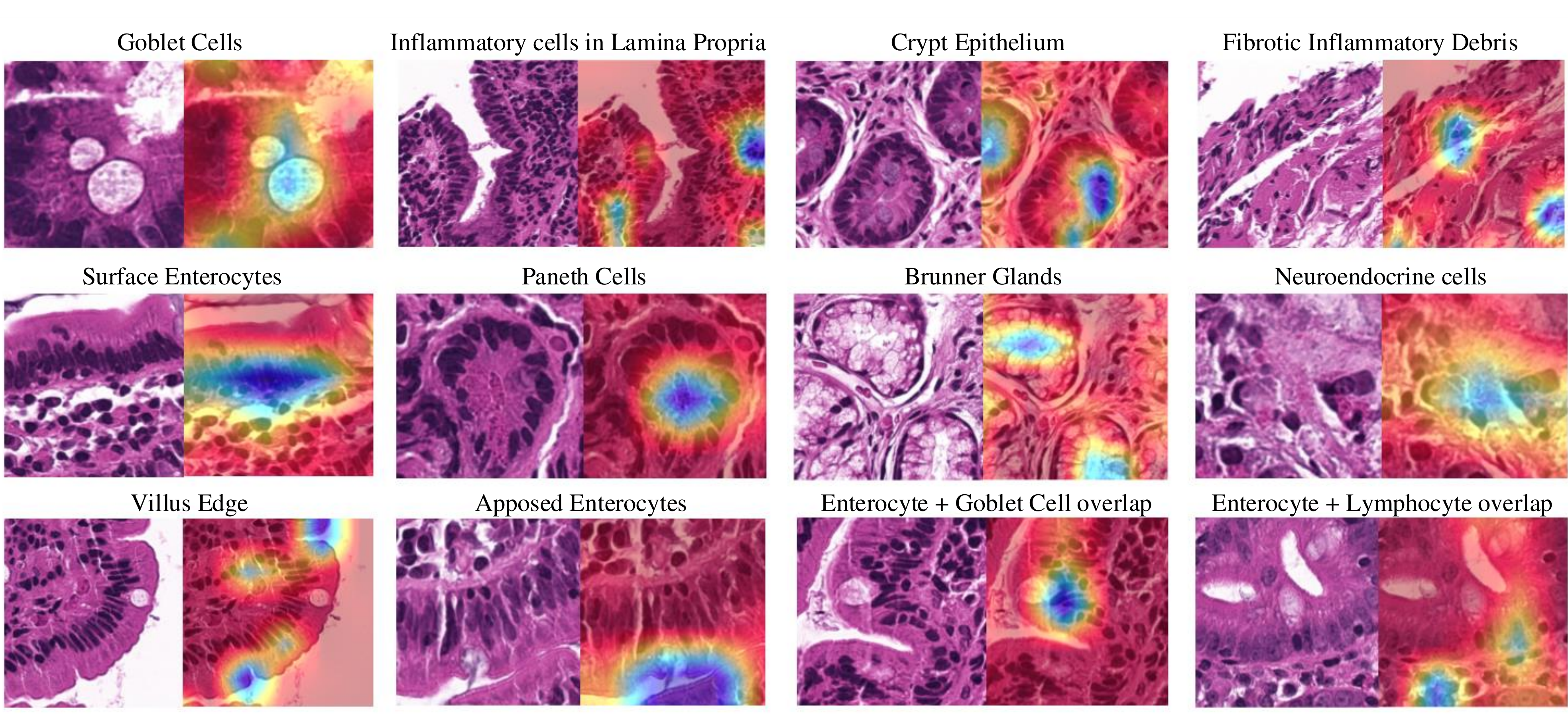}
    \caption{ Class activation mapping heat maps highlighting the most informative regions of patches relevant to different categories including goblet cells, inflammatory cells in lamina propria, crypt epithelium, fibrotic inflammatory debris, surface enterocytes, Paneth cells, Brunner's glands, neuroendocrine cells, villus edge and apposed enterocytes. Area of attention is shown in blue color.
} \label{fig:gradcam_results_5}
\end{figure*}

\section{Experimental results}\label{sec:Empirical_Results_5}
\subsection{Patch-level performance}~To evaluate the effectiveness of our proposed model, we used an independent test set including~$120$ WSIs. After application of a sliding window for patching these whole slides and doing the aforementioned pre-processing steps,~$28,667$ crops remained to be used for our model  evaluation. Performance of our model on this set is shown
in Table \ref{table:patch-level_results}, which includes accuracy, precision, recall, and
the F1 score with~$95\%$ confidence intervals. Also patch-level ROC curves and AUC for each class are shown in Figure~\ref{fig:ROC}. As shown AUC for all classes is greater than $0.96$.

\subsection{Slide-level performance}
After classification of the test patches, their results were aggregated based on the method described in section~\ref{sec:WSIClassification} to make an inference about each test slide. By applying this method, all slides in the test set were classified correctly and the accuracy of the model in all the classes was $100\%$. In the four classes of I, IIIa, IIIb and IIIc there were $20$, $21$, $44$ and $35$ slides, respectively. This means that CD severity was correctly diagnosed.

\subsection{Class Activation Mapping}~We used the Grad-CAM approach to obtain visual explanation microscopic feature heat-maps for WSI patch areas predictive of CD severity. Grad-CAM visualizations were obtained for~$350$ images~($95$ Marsh I, $75$ Marsh IIIa,~$100$ Marsh~IIIb,~$80$ Marsh IIIc). Qualitatively, the Grad-CAM images of our model localized microscopic morphological features such as different cell types and tissue structures that corresponded to the disease pathology. Quantitatively, our Grad-CAM heat-maps were reviewed by two medical professionals. These heat-maps were broadly categorized into~10 groups that are as follows:  goblet cells, inflammatory cells in the lamina propria, crypt epithelium, fibrotic inflammatory debris, surface enterocytes, Paneth cells, Brunner's glands, neuroendocrine cells, villus edge and apposed enterocytes. Visualization of these different categories on individual patches are shown in Fig~\ref{fig:gradcam_results_5}. Most images depicted an overlap of heat-map for enterocytes and goblet cells or enterocytes and lymphocytes that are known to be representative of CD~\cite{oberhuber1999histopathology}~(Fig~\ref{fig:gradcam_results_5}).

\subsection{Hardware and Framework}
The processing units that has been used through this experiment was intel on \textit{Xeon~E5-2640~ (2.6 GHz)} with~\textit{12 cores} and \textit{64~GB} memory~(DDR3). Also, graphical card on our machine is \textit{Nvidia Quadro~K620} and \textit{Nvidia Tesla~K20c}. This work is implemented in Python using Compute Unified Device Architecture~(CUDA) which is a parallel computing platform and Application Programming Interface ~(API) model created by $Nvidia$. We used $TensorFelow$~\cite{abadi2016tensorflow} and $Keras$~\cite{chollet2015keras} library for creating the neural networks.

\section{Conclusion}\label{sec:Conclusion_5}~In this dissertation section, we investigated CD severity using CNNs applied to histopathological images. A state-of-the-art deep residual neural network architecture was used to categorize patients based on H\&E stained duodenal histopathological images into four classes, representing different CD severity based on a histological classification called the modified Marsh score. Our model was trained to classify different patches of WSIs. In addition, we provided a heuristic to aggregate results of patch classification and make inference about the WSIs. Our model was tested on~$28,667$ crops derived from an independent test set $120$ WSIs from $15$ CD patients. It achieved AUC greater than $0.96$ in all classes. At the WSI level classification, the proposed model correctly classified all WSIs. Validation results were highly promising and showed that our model has great potential to be utilized by pathologists to support their CD severity decision based on a histological assessment. We  also used the  Grad-CAM approach to obtain  visual explanation of microscopic features predictive of CD severity. These heat-maps were broadly categorized into~$10$ groups including goblet cells, inflammatory cells in the lamina propria, crypt epithelium, fibrotic inflammatory debris, surface enterocytes, Paneth cells, Brunner's glands, neuroendocrine cells, villus edge and apposed enterocytes.

Albeit achieving promising results, this study has a number of limitations. Firstly, healthy cases were not included this study. This is an avenue for future work. In addition, all biopsy images used in this study were collected from a single medical center and scanned with the same equipment, thus our data may not be representative of the entire range of histopathologic patterns in patients worldwide. Furthermore, the target image for stain normalization was selected manually based on the opinion of a pathologist. Selecting a different image as the target image could affect the appearance of stain normalized images. It is known that some variability exists in this selection, which is then propagated through the framework. Finally, in this study we applied a single method of stain normalization and the use of other methods may lead to different results. Therefore, investigating the effect of different stain normalization techniques can be another potential area of future work.

\chapter{Hierarchical Medical Image Classification~(HMIC)}\label{chpt:HMIC}

Image classification is central for the big data revolution in medicine. Improved information processing methods for diagnosis and classification of digital medical images has shown to be successful via deep learning approaches. As this field is explored, there are limitations to the performance of traditional supervised classifiers. This paper outlines an approach which is different from the current medical image classification methods that view the problem as a multi-class classification. We performed a hierarchical classification using our Hierarchical Medical Image classification~(HMIC) approach. HMIC employs stacks of deep learning architectures to provide specialized understanding at each level of the medical image hierarchy.








\section{Introduction and Related Works}

Automatic diagnosis of diseases based on medical image classification has become increasingly challenging over the last several years~\cite{sali2019celiacnet,kowsari2019diagnosis}. Areas of research involving deep learning architectures for image analysis have grown in the past few years with an increasing interest in their exploration and understanding of the domain application~\cite{Heidarysafa2018RMDL,kowsari2017hdltex,kowsari2018rmdl,info10040150,litjens2017survey,nobles2018identification,zhai2016doubly}. Deep learning models achieved state-of-the-art results in a wide variety of fundamental tasks such as  image classification  in the medical domain~\cite{hegde2019comparison,zhang2018patient2vec}. This growth has raised questions regarding classification of sub-types of disease across a range of disciplines including Cancer (e.g., stage of cancer), Celiac Disease (e.g., Marsh Score Severity Class), and Chronic Kidney Disease (e.g., Stage 1-5) among others~\cite{pavik2013secreted}. Therefore, it is important to not just label medical images based specialized areas, but to also organize them within an overall field (i.e. name of disease) with the accompanying sub-field (i.e. sub-type of disease) which we have done in this paper via Hierarchical Medical Image Classification~(HMIC). Hierarchical models also combat the problem of unbalanced medical image datasets for training the model as have been successful for other domains~\cite{kowsari2017hdltex,dumais2000hierarchical}.

Under-nutrition is the underlying cause of approximately~$45$\% of the~$5$ million under~$5$-year-old childhood deaths annually in low and middle-income countries~(LMICs)~\cite{WHO.Children} and is a major cause of mortality in this population. Linear growth failure (or stunting) is a major complication of under-nutrition, and is associated with irreversible
physical and cognitive deficits, with profound developmental implications~\cite{syed2016environmental}. A common cause of stunting in
LMICs is EE, for which there are no universally accepted, clear diagnostic algorithms or non-invasive
biomarkers for accurate diagnosis~\cite{syed2016environmental}, making this a critical priority~\cite{naylor2015environmental}. EE has been described to be caused by 
chronic exposure to enteropathogens which results in a vicious cycle of constant
mucosal inflammation, villous blunting, and a damaged epithelium~\cite{syed2016environmental}. These deficiencies contribute to a markedly reduced nutrient absorption and thus under-nutrition and stunting~\cite{syed2016environmental}. Interestingly, CD, a common cause of stunting in
the United States, with an estimated~$1$\% prevalence, is an autoimmune disorder caused by a gluten sensitivity~\cite{husby2012european} and has many shared histological features with EE~(such as increased inflammatory cells and villous blunting)~\cite{syed2016environmental}. This resemblance has led to the major challenge of differentiating clinical biopsy images for these similar but distinct diseases. 
Therefore, there is a major clinical interest towards developing new, innovative methods to automate and enhance the detection of morphological features of
EE versus CD, and to differentiate between diseased and healthy small intestinal tissue~\cite{bejnordi2017diagnostic}.

In this dissertation's section, we propose a CNN-based model for classification of biopsy images. In recent years, Deep Learning architectures have received great attention after achieving state-of-the-art results in a wide variety of fundamental tasks such classification~\cite{Heidarysafa2018RMDL,kowsari2017hdltex,kowsari2018rmdl,info10040150,litjens2017survey,nobles2018identification,zhai2016doubly} or other medical domains~\cite{hegde2019comparison,zhang2018patient2vec}. CNNs in particular have proven to be very effective in medical image processing. CNNs preserve local image relations, while reducing dimensionality and for this reason are the most popular machine learning algorithm in image recognition and visual learning tasks~\cite{ker2018deep}. CNNs have been extensively used for classification task and also it used for segmentation in any kind of medical applications such as histopathological
images of breast tissues, lung images, etc.~\cite{gulshan2016development,litjens2017survey}. Researchers and scientists produced a promising result on duodenal biopsies classification using CNNs~\cite{Mohammad_al_boni}, but those models are only robust to a single type of image staining. Many scientists try to apply a stain normalization technique as part of the pre-processing step to both the training and validation datasets~\cite{nawaz2018classification}. In this dissertation's section, varying levels of color balancing were applied during image pre-processing in parent level which order to account for multiple stain variations. 

\begin{figure}
    \centering
    \includegraphics[width=\columnwidth]{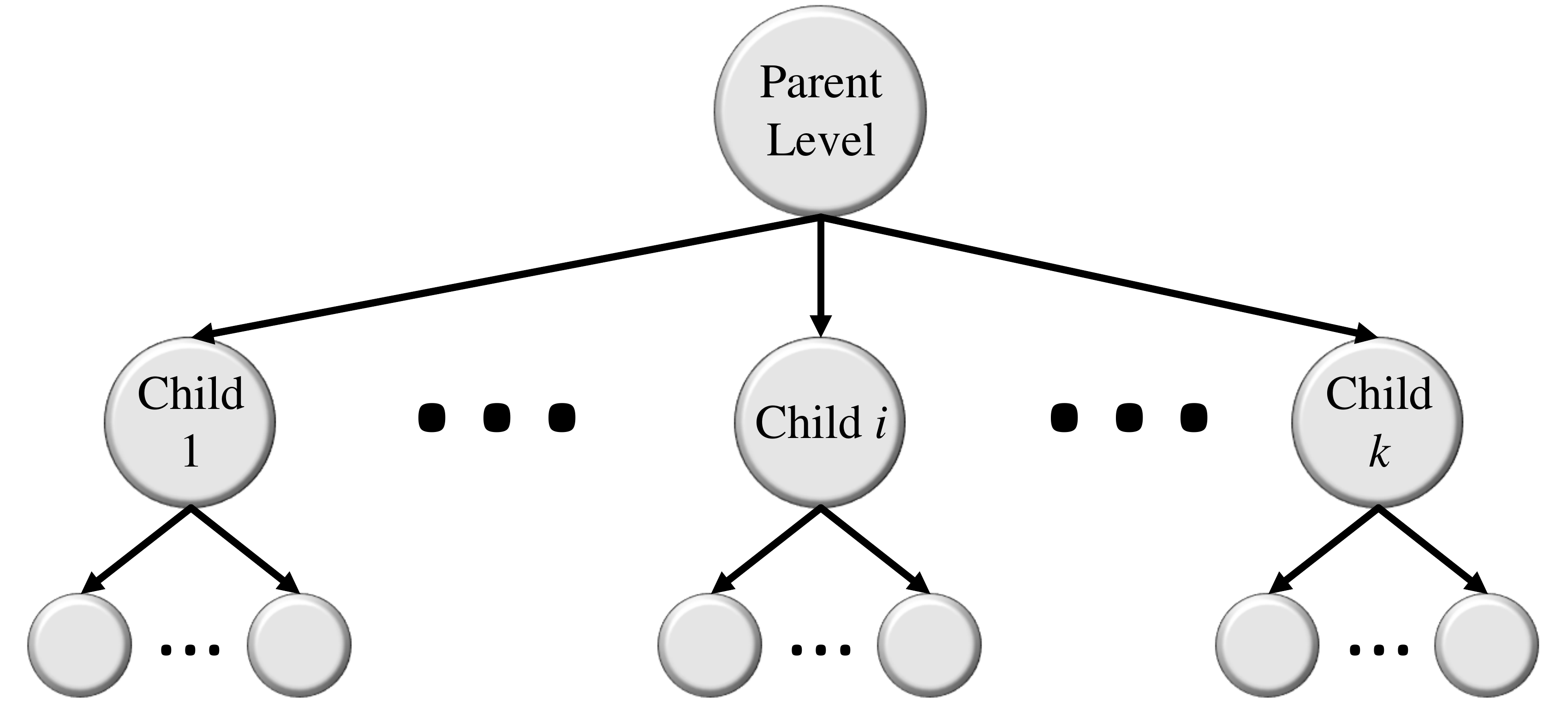}
    \caption{Hierarchical Medical Image Classification} \label{fig:HMIC}
\end{figure}

As shown in Figure~\ref{fig:HMIC}, This technique is data level Hierarchical Medical Image Classification~(HMIC). The parent level is a model to trained based on high level of data. For example, the biopsy images in this dissertation section could be classified into Normal, CD, and EE, but CD could be in the different stages such as I, IIIa, IIIb, and IIIc; thus, the other model could bge trained to diagnosis different type of this disease as child level.

The rest of this dissertation section is organized as follows: In Section~\ref{sec:Data_Source_6}, we describe the different data sets used in this work, as well as, the required pre-processing steps. The architecture of the model is explained in Section~\ref{sec:Method}. Empirical results are elaborated in Section~\ref{sec:Empirical_Results}. Finally, Section~\ref{sec:Conclusion_6} concludes the dissertation section along with outlining future directions.

\section{Data Source}\label{sec:Data_Source_6}

As shown in Table~\ref{ta:population}, The biopsies had already been obtained from 150 children in this study with a median (interquartile range) age of 37.5 (19.0 to 121.5) months and a roughly equal sex distribution; 77 males (51.3\%). Duodenal biopsy slides were converted into~$461$ whole slide images, and labeled as either EE, CD, or normal. The biopsy slides for EE patients were from the Aga Khan University Hospital~(AKUH) in Karachi, Pakistan~($n = 29$ slides from~$10$ patients) and the University of Zambia Medical Center in Lusaka, Zambia ($n = 16$). The slides for CD patients ($n = 34$) and normal ($n = 63$) were retrieved from archives at the University of Virginia~(UVa). CD and normal slides were converted into whole slide images at~$40$x magnification using the Leica SCN~$400$ slide scanner (Meyer Instruments, Houston, TX) at UVa, and the digitized EE slides were of $20$x magnification and shared via the Environmental Enteric Dysfunction Biopsy Investigators~(EEDBI) Consortium shared WUPAX server. Characteristics of our patient population are as follows: the median~($Q1$, $Q3$) age of our entire study population was~$37.5$~($19.0$, $121.5$) months, and we had a roughly equal distribution of males~($52$\%, $n = 53$) and females~($48$\%, $n = 49$). The majority of our study population were histologically normal controls~$(37.7\%)$, followed by CD patients~$(51.8\%)$, and EE patients~$(10.05\%)$.

\begin{table}[]
\centering
\caption{Population results of biopsies dataset}\label{ta:population}
\begin{tabular}{ c  c  c  c  c c }
\hline
                                                                        & Total                                                                                & Pakistan                                                                                        & Zambia                                                                                          & \multicolumn{2}{c}{US}                                                                                                                                                                   \\ \hline
Data                                                                    & 150                                                                                             & \begin{tabular}[c]{@{}c@{}}EE \\(n = 10)  \end{tabular}                                                                             & \begin{tabular}[c]{@{}c@{}}EE\\ (n = 16)   \end{tabular}                                                                                        & \begin{tabular}[c]{@{}c@{}}Celiac \\(n = 63)    \end{tabular}                                                                         & \begin{tabular}[c]{@{}c@{}}Normal\\(n = 61) \end{tabular}                                                                            \\ \hline
\begin{tabular}[c]{@{}c@{}}Biopsy  \\ Imagesa\end{tabular}              & 461                                                                                             & 29                                                                                        & 19                                                                                              & 239                                                                                         & 174                                                                                         \\ \hline
Age & \begin{tabular}[c]{@{}c@{}}37.5 \\ (19 to 121)\end{tabular}                                 & \begin{tabular}[c]{@{}c@{}}22 \\ (20 to 23)\end{tabular}                            & \begin{tabular}[c]{@{}c@{}}16 \\ (9 to 21)\end{tabular}                                   & \begin{tabular}[c]{@{}c@{}}130 \\ (85 to 176)\end{tabular}                            & \begin{tabular}[c]{@{}c@{}}25 \\ (16 to 41)\end{tabular}                              \\ \hline
Gender     & \multicolumn{1}{c}{\begin{tabular}[c]{@{}l@{}}M = 77\\ F  = 73\end{tabular}} & \multicolumn{1}{c}{\begin{tabular}[c]{@{}l@{}}M = 5\\ F  = 5 \end{tabular}} & \multicolumn{1}{c}{\begin{tabular}[c]{@{}l@{}}M = 10 \\ F  = 6 \end{tabular}} & \multicolumn{1}{c}{\begin{tabular}[c]{@{}l@{}}M = 29\\ F  = 34 \end{tabular}} & \multicolumn{1}{c}{\begin{tabular}[c]{@{}l@{}}M = 33\\ F  = 28 \end{tabular}} \\ \hline
\begin{tabular}[c]{@{}c@{}}LAZ/\\  HAZ\end{tabular}      & \begin{tabular}[c]{@{}c@{}}-0.6 \\ (-1.9 to 0.4)\end{tabular}                                   & \begin{tabular}[c]{@{}c@{}}-2.8\\  (-3.6 to -2.3)\end{tabular}                            & \begin{tabular}[c]{@{}c@{}}-3.1 \\ (-4.1 to -2.2)\end{tabular}                                  & \begin{tabular}[c]{@{}c@{}}-0.3 \\ (-0.8 to 0.7)\end{tabular}                               & \begin{tabular}[c]{@{}c@{}}-0.2 \\ (-1.3 to 0.5)\end{tabular}                               \\ \hline
\end{tabular}
\end{table}

~$239$ H\&E stained duodenal biopsy slides were obtained from the archival biopsies of $63$ CD patients from the University of Virginia~(UVa) in Charlottesville, VA, United States. Each slide contained multiple biopsies per patient resulting in whole slide images at 40x magnification using the Leica SCN 400 slide scanner~(Meyer Instruments, Houston, TX) at the Biorepository and Tissue Research Facility at UVa. Characteristics of our patient population were as follows: the median $(Q1, Q3)$ age was $130~(85.0, 176.0)$ months. we had a roughly equal  distribution  of males  $(46\% ,n=29)$  and  females  $(54\%, n=54)$. Biopsy images for our study population were scored by two medical professionals and validated with reads from a pathologist specialized in gastroenterology. Our biopsy image dataset ranged from Marsh I to IIIc with no biopsy images present in Marsh II.

Based on Table~\ref{ta:data}, the biopsy images are patched to~$91,899$ total images which contain $32,393$ normal patches, $29,308$ EE patches, and $30,198$ CD patches. In the child level of the medical biopsy patches, CD contains 4 severities of disease~(Type I, IIIa, IIIb and IIIc) which has $7,125$ Type I patches, $6,842$ Type IIIa patches, $8,120$ Type IIIb patches,  and $8,111$ Type IIIb patches. The training of normal and EE contains $22,676$ and $20,516$ patches, respectively, and for testing $9,717$ and $8,792$ patches, respectively. For CD, we have two sets of training and testing which one belongs to the parent model and the other belongs to child level. The parent set contains $21,140$ patches for training and $9,058$ image patches for testing with the common label of CD for all. In the CD child dataset, we have four types of this disease (I, IIIa, IIIb, and IIIc). Type I of CD  contains $4,988$ patches in the training set and $2,137$ patches in the test set. Type IIIa of CD contains $4,790$ patches in the training set and $2,052$ patches in the test set. Type IIIb of CD contains $5,684$ patches in the training set and $2,436$ patches in the test set. Finally,  IIIc of CD contains $5,678$ patches in the training set and $2,137$ patches in the test set.

\begin{figure}[!b]
    \centering
    \includegraphics[width=\textwidth]{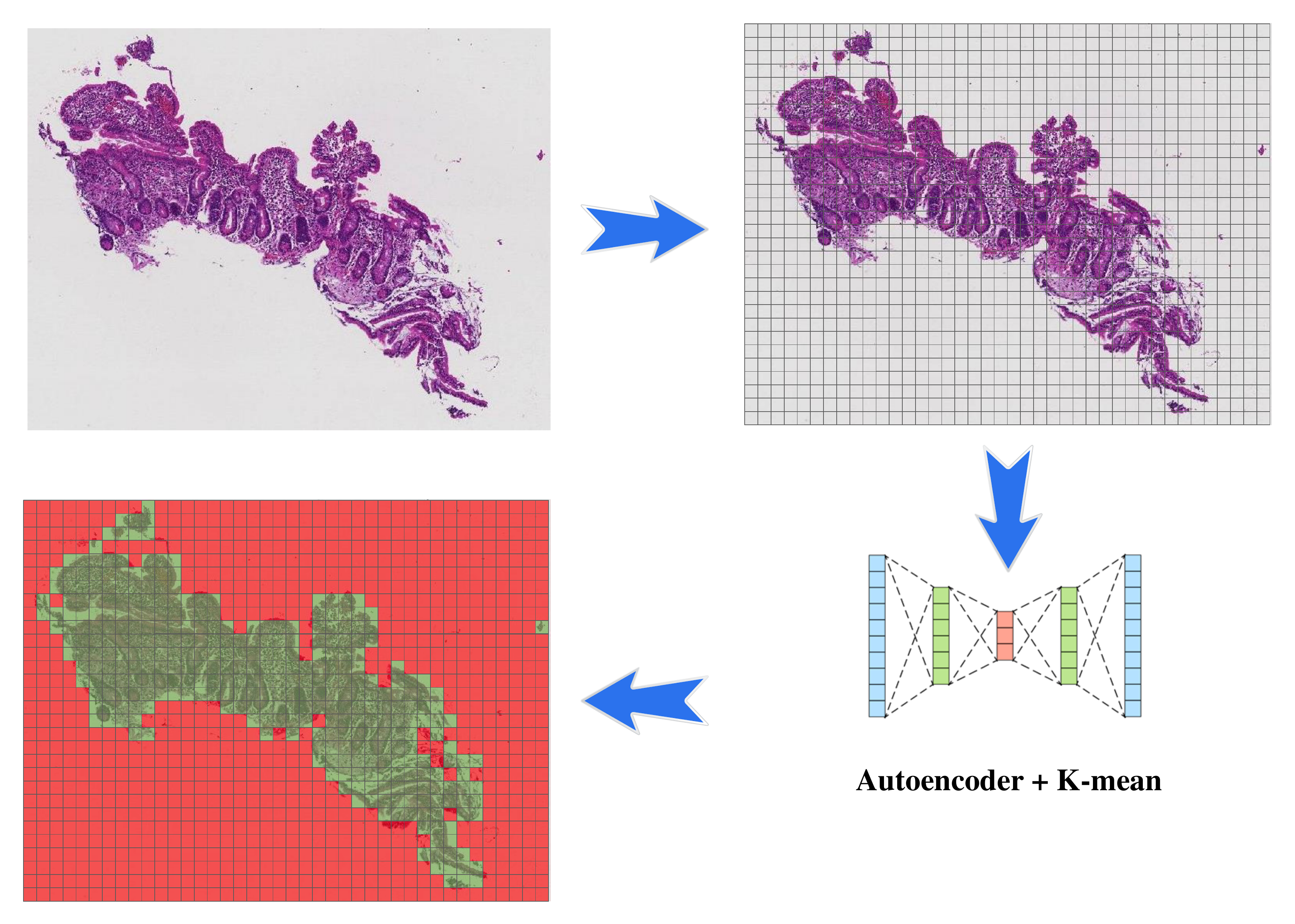}
    \caption{Pipeline of patching and autoencoder to find useful patches for training model. The biopsy images are very large, so we need to divide into smaller patches to be used in the machine learning model. As you can see in the image, many of these patches do not contain any useful medical information. After using an autoencoder, we can apply a clustering algorithm to discard useless patches (green patches contain useful information, while red patches do not).} \label{fig:Patch}
\end{figure}

\section{Pre-Processing}\label{sec:Pre-Processing}
In this section, we cover all of the pre-processing steps which include image patching, image clustering, and color balancing. The biopsy images are unstructured~(varying image sizes) and too large to process with deep neural networks; thus, requiring that images are split into multiple smaller images. After executing the split, some of the images do not contain much useful information. For instance, some only contain the mostly blank border region of the original image. In the image clustering section, the process to select useful images is described. Finally, color balancing is used to correct for varying color stains which is a common issue in histological image processing.


\begin{figure}[!b]
    \centering
    \includegraphics[width=\columnwidth]{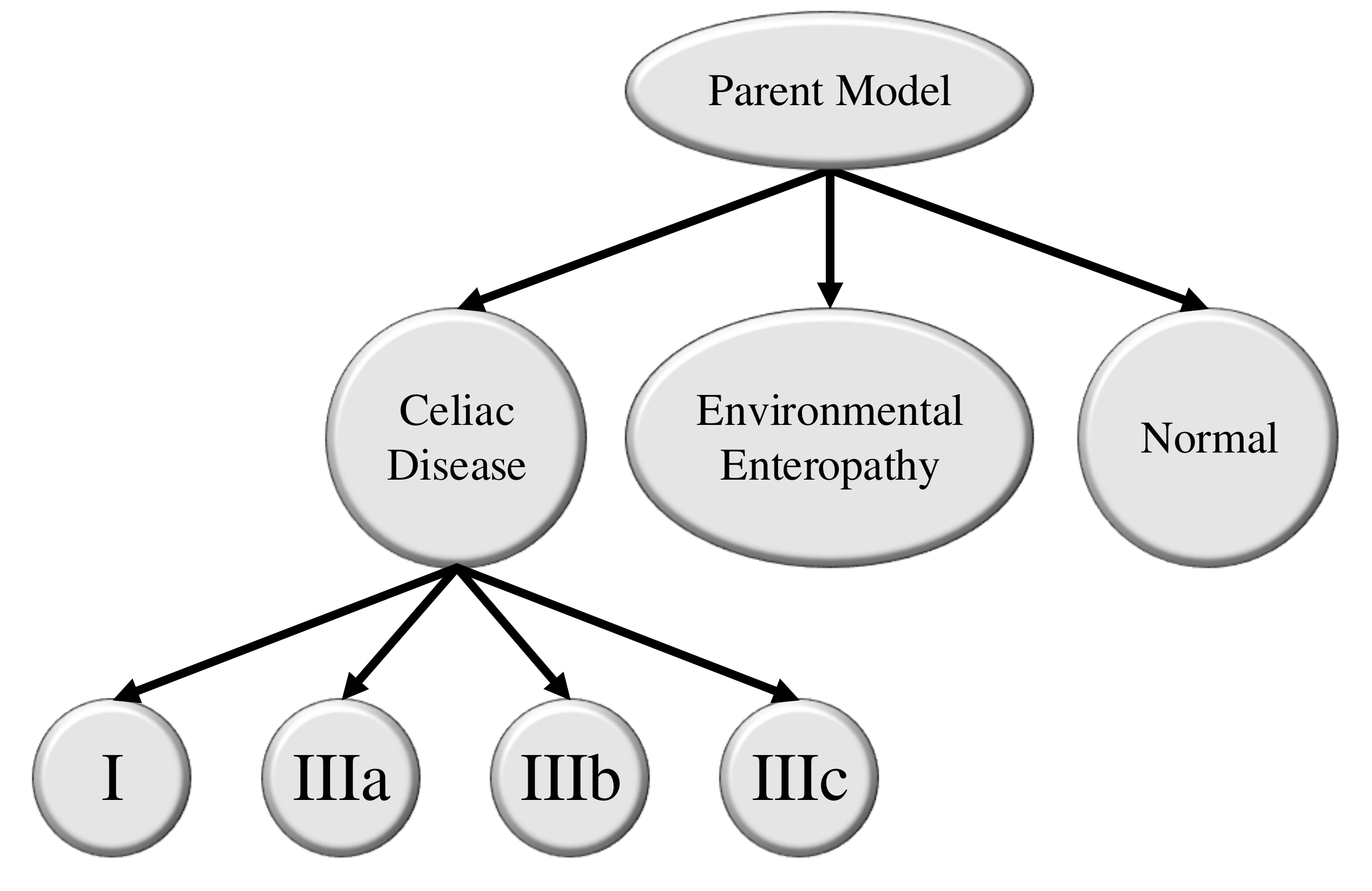}
    \caption{Hierarchical Medical Image classification~(HMIC) to diagnosis Celiac Disease and
Environmental Enteropathy and child level of Celiac Disease Severity Diagnosis
on Duodenal Histopathological Images} \label{fig:HC4Biopsy}
\end{figure}

\subsection{Image Patching}\label{subsec:Image_Patching6}
Although effectiveness of CNNs in image classification has been shown in various studies in different domains, training on high resolution Whole Slide Tissue Images (WSI) is not commonly preferred due to a high computational cost. However, applying CNNs on WSI enables losing a large amount of discriminative information due to extensive downsampling~\cite{hou2016patch}. Due to a cellular level differences between Celiac disease, environmental entropathy, and normal cases, a trained classifier on image patches is likely to perform as well as or even better than a trained WSI-level classifier. Many researchers in pathology image analysis have considered classification or feature extraction on image patches~\cite{hou2016patch}.
In this project, after generating patches from WSI, labels were applied to each patch according to its associated original WSI. A CNN was trained to generate predictions on each individual patch. As you shown in Figure~\ref{fig:Patch}, each biopsy whole image is divided into many patches.

\begin{table}[]
\caption{Dataset is used for Hierarchical Medical Image Classification (HMIC) }\label{ta:data}
\centering
\begin{tabular}{|c|c|c|c|c|c|c|c|}
\hline
\multicolumn{2}{|c|}{Data}                           & \multicolumn{2}{c|}{Train}      & \multicolumn{2}{c|}{Test}      & \multicolumn{2}{c|}{Total}      \\ \hline
\multicolumn{2}{|c|}{Normal}                         & \multicolumn{2}{c|}{22,676}     & \multicolumn{2}{c|}{9,717}     & \multicolumn{2}{c|}{32,393}     \\ \hline
\multicolumn{2}{|c|}{\begin{tabular}[c]{@{}c@{}}Environmental\\ Enteropathy (EE) \end{tabular}}  & \multicolumn{2}{c|}{20,516}     & \multicolumn{2}{c|}{8,792}     & \multicolumn{2}{c|}{29,308}     \\ \hline
\multirow{5}{*}{Celiac Disease (CD)}      &          & Parent                  & Child & Parent                 & Child & Parent                  & Child \\ \cline{2-8} 
                                          & I        & \multirow{4}{*}{21,140} & 4,988 & \multirow{4}{*}{9,058} & 2,137 & \multirow{4}{*}{30,198} & 7,125 \\ \cline{2-2} \cline{4-4} \cline{6-6} \cline{8-8} 
                                          & IIIa     &                         & 4,790 &                        & 2,052 &                         & 6,842 \\ \cline{2-2} \cline{4-4} \cline{6-6} \cline{8-8} 
                                          & IIIb     &                         & 5,684 &                        & 2,436 &                         & 8,120 \\ \cline{2-2} \cline{4-4} \cline{6-6} \cline{8-8} 
                                          & IIIc     &                         & 5,678 &                        & 2,433 &                         & 8,111 \\ \hline
\end{tabular}
\end{table}

\subsection{Clustering}\label{subsec:Clustering_6}~\\
As you shown in Figure~\ref{fig:Patch}, after each biopsy whole image is divided into many patches, many of these images are useless; thus, autoencoder help us to select only useful patches. In this study, after image patching, some of created patches do not contain any useful information regarding biopsies and should be removed from the data~\cite{kowsari2019diagnosis}. These patches have been created from mostly background parts of WSIs. A two-step clustering process was applied to identify the unimportant patches. For the first step, a convolutional autoencoder was used to learn embedded features of each patch and in the second step we used k-means to cluster embedded features into two clusters: useful and not useful. In Figure~\ref{fig:AE}, the pipeline of our clustering technique is shown which contains both the autoencoder and k-mean clustering.

\subsubsection{Autoencoder}

An autoencoder is a type of neural network that is designed to match the model's inputs to the outputs~\cite{goodfellow2016deep}. The autoencoder has achieved great success as a dimensionality reduction method via the powerful reprehensibility of neural networks~\cite{wang2014generalized}. The first version of autoencoder was introduced by~\textit{DE. Rumelhart el at.}~\cite{rumelhart1985learning} in 1985. The main idea is that one hidden layer between input and output layers has much fewer units~\cite{liang2017text} and can be used to reduce the dimensions of a feature space. For medical images which typically contain many features, using an autoencoder can help allow for faster, more efficient data processing.

A CNN-based autoencoder can be divided into two main steps~\cite{masci2011stacked}~: encoding and decoding. This equation is:

\begin{equation}
\begin{split}
    O_m(i, j) = a\bigg(&\sum_{d=1}^{D}\sum_{u=-2k-1}^{2k+1}\sum_{v=-2k -1}^{2k +1}F^{(1)}_{m_d}(u, v)I_d(i -u, j -v)\bigg) \\&\quad m = 1, \cdots, n
\end{split}
\end{equation}
Where~$F \in \{F^{(1)}_{1},F^{(1)}_{2},\hdots,F^{(1)}_{n},\}$ is a convolutional filter, with convolution among an input volume defined by~$I = \left\{I_1,\cdots, I_D\right\}$ which it learns to represent the input by combining non-linear functions:

\begin{figure}[!b]
    \centering
    \includegraphics[width=\columnwidth]{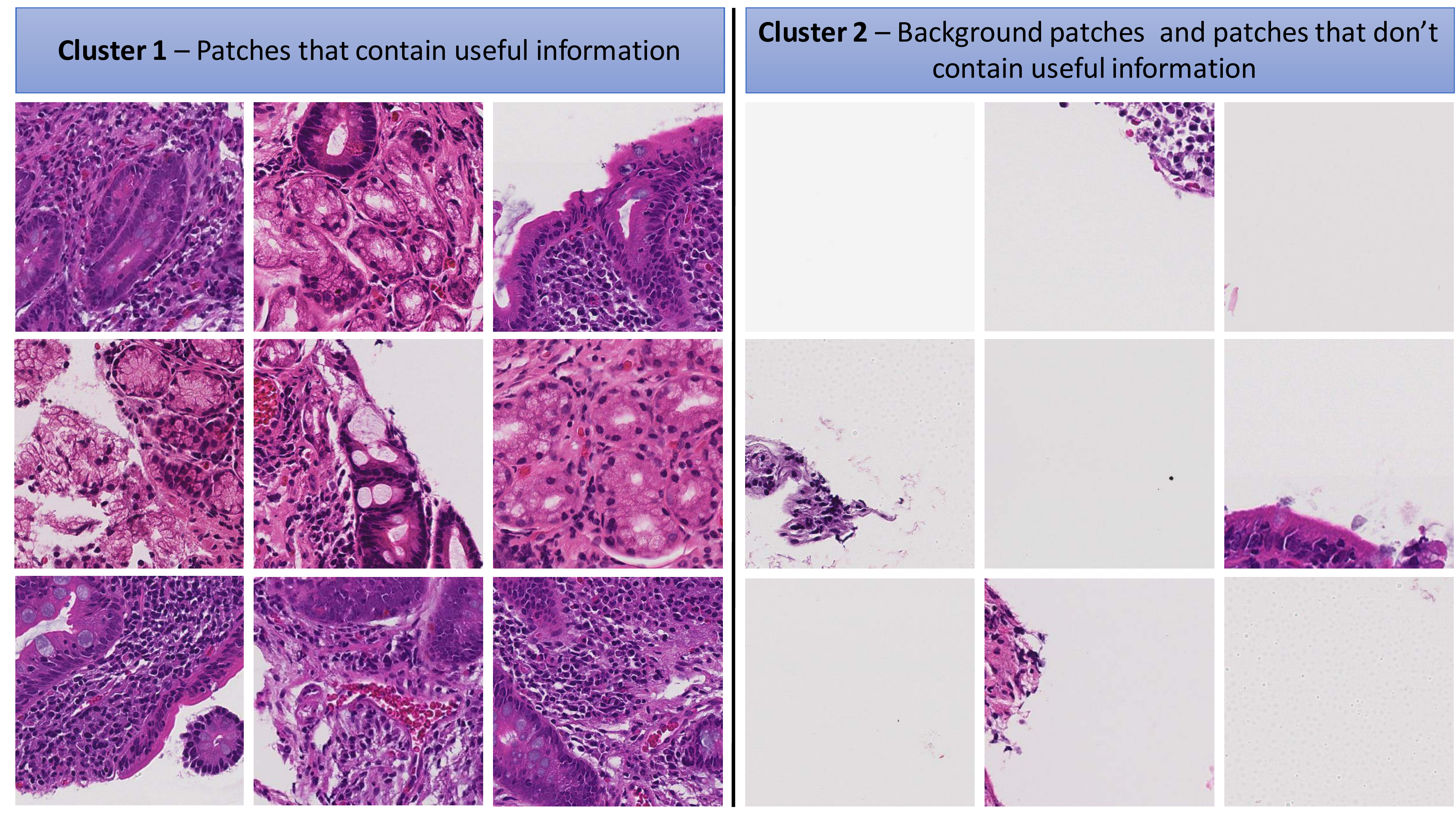}
    \caption{Some samples of clustering results - cluster 1 includes patches with useful information and cluster 2 includes patches   without useful information (mostly created from background parts of WSIs)} \label{fig_Clustering6}
\end{figure}

\begin{equation}
    z_m = O_m = a(I * F^{(1)}_{m} + b^{(1)}_m) \quad m = 1, \cdots, m
\end{equation}
where~$b^{(1)}_m$ is the bias, and the number of zeros we want to pad the input with is such that: \text{dim}(I) = \text{dim}(\text{decode}(\text{encode}(I))). Finally, the encoding convolution is equal to:
\begin{equation}
\begin{split}
     O_w = O_h &= (I_w + 2(2k +1) -2) - (2k + 1) + 1 \\&= I_w + (2k + 1) - 1
\end{split}
\end{equation}
The decoding convolution step produces~$n$ feature maps~$z_{m=1,\hdots,n}$. The reconstructed results~$\hat{I}$ is the result of the convolution between the volume of feature maps~$Z=\{z_{i=1}\}^n$ and this convolutional filters volume~$F^{(2)}$~\cite{chen2015page,geng2015high}.

\subsubsection{K-Means}
\label{sec:kmeans6}
K-means clustering is one of the most popular clustering algorithms~\cite{jain2010data} in which data is given  $D\in\{x_1,x_2,...,x_n\}$ in $d$ dimensional vectors for $x\in f^d$. The aim is to identify groups of similar data points and assign each point to one of the groups. There are many other clustering algorithms, but the K-means approach works well for this problem, because there are only two clusters and it is computationally inexpensive compared to other methods. As an unsupervised approach, ...~\cite{kowsari2016weighted}. 

If we want to find the error rate the time complexity is equal to exponential. One measure of how good clustering is would be the sum of distances to the center. Therefore, K-means wants to minimize this $\xi$~(quantity), choosing $\mu$ and a’s to minimize but it is difficult to do analytically because a’s are binary assignments. So the K-means algorithm tries to iteratively solve the minimization~(sort of greedy algorithm).

Minimize~$\xi$ with respect to~$a$ and~$\mu$ by:
\begin{align}
    \xi = \sum_{j=1}^k \sum_{x_i} ||x_i-\mu_j||^2 = \sum_{j=1}^k \sum_{i=1}^n A_{ij}||x_i-\mu_j||
\end{align}
where $\xi$ is the number of data points, and $\mu$ is the centroid of each cluster.

\begin{algorithm}[!t]
\SetKwFor{Foreach}{for}{do}{endfor}
\SetKwFor{WHILE}{while}{do}{endwhile}
\caption{K-means $n$ images for $K$ Clusters (in our expriment k=2)}

\textbf{Input: }$D= \{\overrightarrow{x_1},\overrightarrow{x_2},...,\overrightarrow{x_n}\}$

\textbf{Output: }$\mu = \overrightarrow{\mu_1},\overrightarrow{\mu_2},...,\overrightarrow{\mu_k}\} $

$\{\overrightarrow{s_1}, \overrightarrow{s_2},...,\overrightarrow{s_k}\}$ set random seeds\\ $(\{\overrightarrow{x_1},\overrightarrow{x_2},...,\overrightarrow{x_n}\},K)$

    \Foreach {$k \leftarrow 1$ to K}{
    
$\overrightarrow{\mu_k} \leftarrow \overrightarrow{s_k}$
    
    }
    \WHILE{Criterion has not been met}
    {
      \Foreach {$k \leftarrow 1$ to K}{
      
      $w_k \leftarrow \{\}$
      }
      
    \Foreach {$n \leftarrow 1$ to $N$}{
      
      $j \leftarrow arg \min_{j'} |\overrightarrow{\mu_{j'}} - \overrightarrow{\mu_{x_n}}|$

      $w_j \leftarrow w_j \bigcup ~\{\overrightarrow{x_n}\}$

     }
    \Foreach {$k \leftarrow 1$ to K}{
      
      $\mu_k \leftarrow \frac{1}{|w_k|} \sum_{\overrightarrow{x}\in w_k} \overrightarrow{x}$
      
      }
    }
\end{algorithm}

K-Means is not guaranteed to converge to the global minimum of this function. In fact, finding a global minimum is an NP-hard problem, but it is reasonable to some extent. The time complexity of K-means is equal to $K$, the number of clusters, multiplied by $I$, the number of the iterations, Although if $I$ is not available, time complexity can be exponential. K-means is also used as image and data clustering for information retrieval~\cite{jain2010data,manning20introduction}. The data points or image representations from the same cluster behave similarly with respect to relevance to datapoint's information which is extracted as feature space. The centroid $\mu$ of each cluster is calculated as follows:
\begin{equation}
    \mu (w) = \frac{1}{|w|} \sum_{\bar{x}\in w} \bar{x}
\end{equation}

 Results of patch clustering have been summarized in Table~\ref{tb:clustering}. Obviously, patches in cluster~$1$, which were deemed useful, are used for the analysis in this section.

\subsection{Staining Problem}

\subsubsection{Color Balancing}
The concept of color balancing for this paper is to convert all images to the same color space to account for variations in H\&E staining. The images can be represented with the illuminant spectral power distribution~$I(\lambda)$, the surface spectral reflectance~$S(\lambda)$, and the sensor spectral sensitivities~$C(\lambda)$~\cite{bianco2017improving,bianco2014error}. Using this notation~\cite{bianco2014error}, the sensor responses at the pixel with coordinates~$(x,y)$ can be found in Section~\ref{subsec:CB_3}

\subsubsection{Stain normalization}
For child level (only CD which are collected from one center), we are using color normalization (See Section~\ref{subsec:StainNormalization}).

\section{Baseline}

\subsection{Deep Convolutional Neural Networks}
A Convolutional Neural Network~(CNN) performs hierarchical medical image classification for each individual image. The original version of the CNN was built for image processing with an architecture similar to the visual cortex . In this basic CNN baseline for image processing, an image tensor is convolved with a set of kernels of size $d \times d$. These convolution layers are called feature maps and these can be stacked to provide multiple filters on the input. We used a flat CNN as one of our baselines.

\subsection{Deep Neural Networks}
A Deep Neural Network (DNN) or multilayer perceptron is designed to be trained by a multi-connection of layers. Each layer only receives connection from the previous layers' nodes and provides connections to the next layer. The input is a connection of flattened feature space~(RGB). The output layer is number of classes for multi-class classification~(six nodes). Our baseline implementation of DNN~(multilayer perceptron) is a discriminative trained model that uses a standard back-propagation algorithm with  sigmoid~(equation~\ref{sigmoid}) and Rectified Linear Units (ReLU)~\cite{nair2010rectified}~(equation~\ref{relu}) activation functions. The output layer for multi-class classification uses the $Softmax$ activation function shown in Equation~\ref{Softmax}.

\section{Method}\label{sec:Method}

In this section, we describe Convolutional Neural Networks~(CNN) including the convolutional layers, pooling layers, activation functions, and optimizer. Then, we discuss our network architecture for diagnosis of Celiac Disease and Environmental Enteropathy. As shown in figure~\ref{fig:cnn}, the input layers starts with image patches~($1000\times 1000$) and is connected to the convolutional layer~(\textit{Conv~$1$}). Conv~$1$ is connected to the pooling layer~(\textit{MaxPooling}), and then connected to \textit{Conv~$2$}. Finally, the last convolutional layer~(\textit{Conv~$3$}) is flattened and connected to a fully connected perception layer. The output layer contains three nodes which each node represents one class.

\subsection{Convolutional Neural Networks}
\subsubsection{Convolutional Layer}
CNN is a deep learning architecture that can be employed for hierarchical image classification. Originally, CNNs were built for image processing with an architecture similar to the visual cortex. CNNs have been used effectively for medical image processing. In a basic CNN used for image processing, an image tensor is convolved with a set of kernels of size~$d \times d$. These convolution layers are called feature maps and can be stacked to provide multiple filters on the input. The element~(feature) of input and output matrices can be different~\cite{li2014medical}. The process to compute a single output matrix is defined as follows:

\begin{equation}
    A_{j}=f\left(\sum_{i=1}^{N}I_{i}\ast K_{i,j}+B_{j}\right)
\end{equation}
Each input matrix~$I-i$ is convolved with a corresponding kernel matrix~$K_{i,j}$, and summed with a bias value~$B_j$ at each element. Finally, a non-linear activation function~(See Section~\ref{Sec:Activation_6}) is applied to each element~\cite{li2014medical}.

In general, during the back propagation step of a CNN, the weights and biases are adjusted to create effective feature detection filters. The filters in the convolution layer are applied across all three 'channels' or $\Sigma$~(size of the color space)~\cite{Heidarysafa2018RMDL}. 
\begin{figure}[t]
    \centering
    \includegraphics[width=\textwidth]{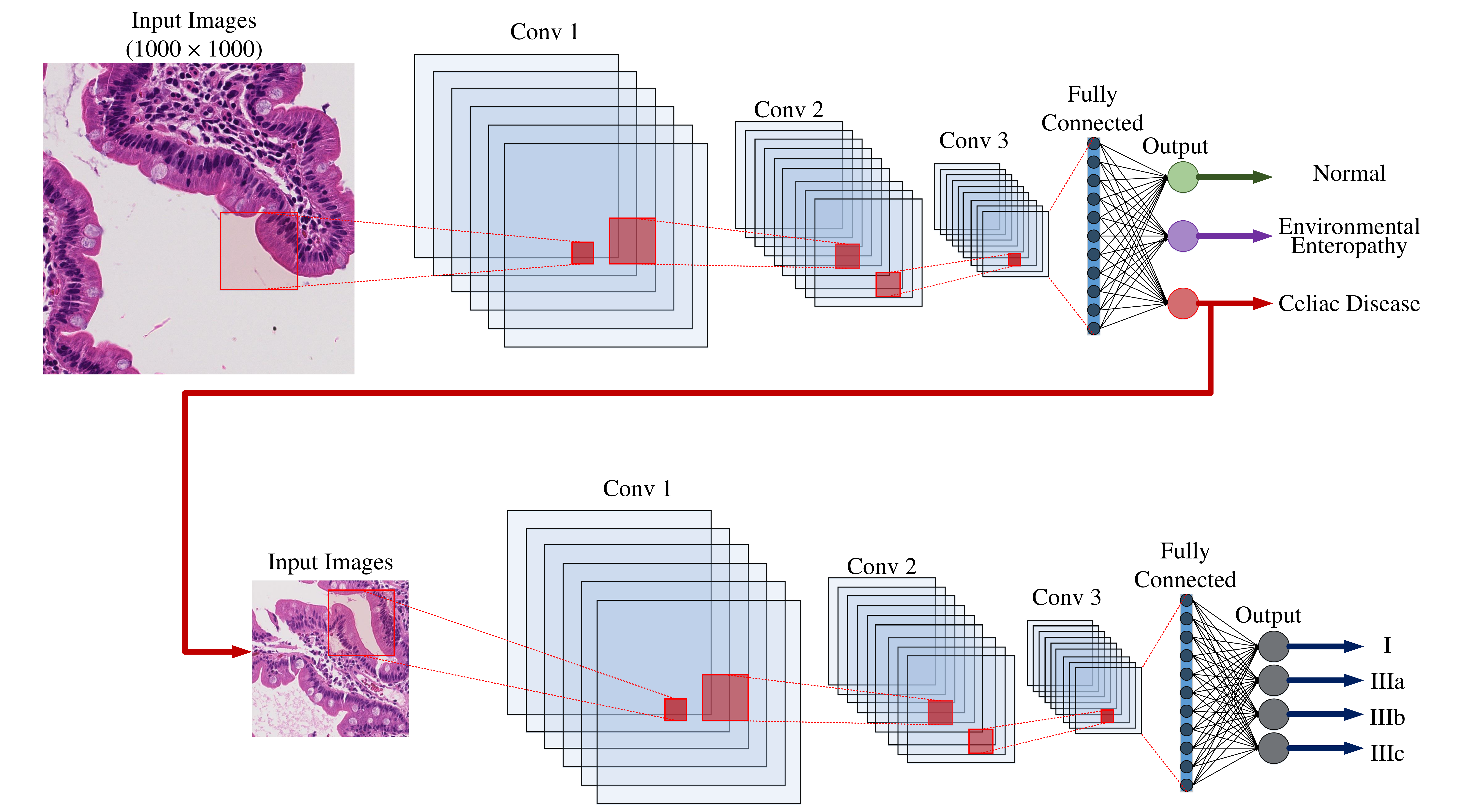}
    \caption{Structure of Convolutional Neural Net using multiple 2D feature detectors and 2D max-pooling} \label{fig:cnn}
\end{figure}

\subsubsection{Pooling Layer}
To reduce the computational complexity, CNNs utilize the concept of pooling to reduce the size of the output from one layer to the next in the network. Different pooling techniques are used to reduce outputs while preserving important features~\cite{scherer2010evaluation}. The most common pooling method is max pooling where the maximum element is selected in the pooling window.\\ In order to feed the pooled output from stacked featured maps to the next layer, the maps are flattened into one column. The final layers in a CNN are typically fully connected~\cite{kowsari2018rmdl}.\\

\subsubsection{Neuron Activation}\label{Sec:Activation_6}
 The implementation of CNN is a discriminative trained model that uses standard back-propagation algorithm using a sigmoid~(Equation~\ref{sigmoid}), (Rectified Linear Units (ReLU)~\cite{nair2010rectified}~(Equation~\ref{relu}) as activation function. The output layer for multi-class classification includes a $Softmax$ activation function~(as shown in Equation~\ref{Softmax}).

\subsubsection{Optimizer}\label{sec:optimizer_6}
For this CNN architecture, the $Adam$ Optimizer \cite{kingma2014adam} which is a stochastic gradient optimizer that uses only the average of the first two moments of gradient~($v$ and $m$, shown in Equation~\ref{adam}, \ref{adam1}, \ref{adam2}, and \ref{adam3}). It can handle non-stationary of the objective function as in RMSProp, while overcoming the sparse gradient issue limitation of RMSProp~\cite{kingma2014adam}.

\subsubsection{Network Architecture}
As shown in Figure~\ref{fig:cnn}, our CNN architecture consists of three convolution layers each followed by a pooling layer. This model receives RGB image patches with dimensions of ~$(1000\times 1000)$ as input. The first convolutional layer has~$32$ filters with kernel size of~$(3, 3)$. Then, we have a pooling layer with size of~$(5,5)$ which reduces the feature maps from~$(1000\times 1000)$ to~$(200 \times 200)$. The second convolutional layer contains~$32$ filters with kernel size of~$(3, 3)$. Then, a pooling layer~(MaxPooling~$2D$) with size of~$(5,5)$ reduces the feature maps from~$(200\times 200)$ to~$(40 \times 40)$. The third convolutional layer contains~$64$ filters with kernel size of~$(3, 3)$, and is followed by a final pooling layer~(MaxPooling~$2D$) to scale down to~$(8 \times 8)$. The feature map is flattened and connected to a fully connected layer with~$128$ nodes. The output layer contains three nodes to represent the three classes:~(Environmental Enteropathy, Celiac Disease, and Normal). The child level of this model as shown on the bottom of Figure~\ref{fig:cnn}, contains three convolutional layers. The first convolutional layer contains~$64$ filters with kernel size of~$(3, 3)$ and is connected to a pooling layer with a pooling size of~$(5,5)$ to reduce the feature maps from~$(1000\times 1000)$ to~$(200 \times 200)$. The second convolutional layers contains~$64$ filters with kernel size of~$(3, 3)$ and is followed by a pooling layer~(MaxPooling~$2D$) with a size of~$(5,5)$ to reduce the feature maps from~$(200\times 200)$ to~$(40 \times 40)$. Finally, the last convolutional layer contains~$128$ filters with a kernel size of~$(3, 3)$, and is followed by a final pooling layer~(MaxPooling~$2D$) that scales down to~$(8 \times 8)$.

The optimizer used is Adam~(See Section~\ref{sec:optimizer_6}) with a learning rate of~$0.001$, $\beta_1=0.9$, and $\beta_2=0.999$.  The loss function is sparse categorical crossentropy~\cite{chollet2015keras}. Also, for all layers, we use Rectified linear unit~(ReLU) as activation function except output layer which we use~$Softmax$~(See Section~\ref{Sec:Activation_6}).

\subsection{Whole Slide Classification}

Our goal was to classify WSIs based on severity assessed via the modified Marsh score. The model used was trained to classify small patches rather than WSIs. To achieve this goal, a heuristic method was developed which aggregated crop classifications and translated them to whole-slide inferences. Each WSI in the test set was initially patched, those patches which did not contain any information were filtered out and then stain normalization was performed on the remaining patches. After these pre-processing steps, our trained model was applied with the goal of image classification. We denoted the probability distribution over possible labels, given the crop $x$ and training set~$D$ by~$p(y\vert x,D)$. In general, this represented a vector of length $C$, where $C$ is number of classes. In our notation, the probability is conditional on the test patch $x$, as well as, the training set~$D$. For each crop, the model gives an output of a vector composed of four components showing probabilities for each one of the four classes of CD severity. Given a probabilistic output, the patch~$j$ in slide $i$ is assigned to the most probable class label~$\hat{y}_{ij}$ which is shown in Equation~\ref{eq:patchClass}.

\subsection{Hierarchical Medical Image Classification}
The primary contribution of this paper is hierarchical image classification of biopsies.  A traditional multi-class classification technique can work well for a limited number of classes, but performance drops when we have unbalanced data or an increasing number of classes. In our hierarchical deep learning model, we solve this problem by creating architectures that create specialized deep learning approaches for their level of the medical hierarchy~(e.g., see Figure~\ref{fig:cnn}).

\section{Results}\label{sec:Empirical_Results}
In this section we have two main results which are empirical results and visualizations for patches. The empirical results are mostly used for comparing our accuracy with our baseline.

\subsection{Evaluation Setup}\label{sec:Evaluation_6}
In the research community, comparable and shareable performance measures to evaluate algorithms are preferable. However, in reality, such measures may only exist for a handful of methods. The major problem when evaluating image classification methods is the absence of standard data collection protocols. Even if a common collection method existed, simply choosing different training and test sets can introduce inconsistencies in model performance~\cite{yang1999evaluation}. Another challenge with respect to method evaluation is being able to compare different performance measures used in separate experiments. Performance measures generally evaluate specific aspects of classification task performance, and thus do not always present identical information. In this section, we discuss evaluation metrics and performance measures and highlight ways in which the performance of classifiers can be compared. Some examples of these metrics include recall, precision, accuracy, and F-measure. These metrics are based on a~``confusion matrix'' that comprises true positives~(TP), false positives~(FP), false negatives~(FN) and true negatives~(TN)~\cite{lever2016points}. The significance of these four elements may vary based on the classification application. The fraction of correct predictions over all predictions is called accuracy~(Eq. \ref{eq:acc}). The proportion of correctly predicted positives to all positives is called precision,~\textit{i.e.} positive predictive value (Eq. \ref{eq:pres}).

\begin{table}[!t]
\centering
\caption{Result of parent level classifications for normal, environmental enteropathy, and  Celiac disease}\label{ta:parent}
\begin{tabular}{ccccc}
\hline
                               & Precision & Recall & F1-score  \\ \hline
Normal                         &   89.97$\pm$0.59        &   89.35$\pm$0.61     &    89.66$\pm$0.60          \\ \hline
\multicolumn{1}{c }{Environmental Enteropathy } &     94.02$\pm$0.49      &   97.30$\pm$0.33      &     95.63$\pm$0.42       \\ \hline
Celiac Disease           &      91.12$\pm$0.32     &   88.71$\pm$0.35     &     89.90$\pm$1.27        \\ \hline
\end{tabular}
\end{table}

\begin{table}[!b]
\centering
\caption{Results of HMIC with comparison with our baseline}\label{ta:results_overall}
\begin{tabular}{|c|c c c c |}
\hline
\multicolumn{2}{|c}{Model}                   & Precision                   & Recall                      & F1-score                                      \\ \hline
\multirow{3}{*}{Baseline}      & CNN          & 76.76$\pm$0.49          & 80.18$\pm$0.47          & 78.43$\pm$0.48                   \\ \cline{2-5} 
                               & DNN          &            76.19$\pm$0.50                 &             79.40$\pm$0.47                &             77.76$\pm$0.49                                 \\ \cline{2-5} 
                               &  DCNN         & 82.95$\pm$0.44          & 87.28$\pm$0.39          & 85.06$\pm$0.42                  \\ \hline
\multirow{2}{*}{HMIC} &NWS & 84.13$\pm$0.37 & 93.56$\pm$0.29 & 88.61$\pm$0.37  \\ \cline{2-5} 
& WS  & \textbf{88.01$\pm$0.38} & \textbf{93.98$\pm$0.28} & \textbf{90.89$\pm$0.38} 
                              \\ \hline
\end{tabular}
\end{table}

\subsection{Experimental Setup}
The following results were obtained using a combination of central processing units (CPUs) and graphical processing units~(GPUs). The processing was done on a $Core i7-9700F$ with $8$ cores and $128 GB$ memory, and the GPU cards were two  \emph{Nvidia~GeForce~RTX~2080 Ti}. We implemented our approaches in Python using the Compute Unified Device Architecture~(CUDA), which is a parallel computing platform and Application Programming Interface~(API) model created by $Nvidia$. We also used Keras and TensorFlow libraries for creating the neural networks~\cite{abadi2016tensorflow,chollet2015keras}.

\begin{table}[!t]
\centering
\caption{Results per-classed of HMIC with comparison with our baseline}\label{ta:results_per_class}
\begin{tabular}{|c| c |c| c| c c c |}
\hline
\multicolumn{3}{|c }{Model}                                                             &          & Precision & Recall & F1-score  \\ \hline
    \multirow{18}{*}{Baseline} & \multirow{6}{*}{CNN} & \multicolumn{2}{c|}{Normal}                    &    87.83$\pm$0.57       &    90.77$\pm$0.65    &      89.28$\pm$0.61       \\ \cline{3-7} 
                               &                      & \multicolumn{2}{c|}{EE} &     90.93$\pm$0.61      &    82.48$\pm$0.79    &    86.50$\pm$0.71       \\ \cline{3-7} 
                               &                      & \multirow{4}{*}{CD}     & I        &    68.37$\pm$1.98       &   68.62$\pm$ 1.96   &    68.50$\pm$1.96       \\ \cline{4-7} 
                               &                      &                                     & IIIa     &  56.26$\pm$1.01         &    56.26$\pm$2.21    &  59.29$\pm$1.95          \\ \cline{4-7} 
                               &                      &                                     & IIIb     &   65.28$\pm$0.97        &    98.28$\pm$2.01    &   66.64$\pm$1.87           \\ \cline{4-7} 
                               &                      &                                     & IIIc     &  62.66$\pm$1.99         &    66.83$\pm$1.99    &   64.68$\pm$2.02           \\ \cline{2-7} 
                               & \multirow{6}{*}{DNN}    & \multicolumn{2}{c|}{Normal}                    &      87.97$\pm$0.76     &    81.87$\pm$0.76    &    84.81$\pm$0.71       \\ \cline{3-7} 
                               &                      & \multicolumn{2}{c|}{EE} &     87.25$\pm$0.69      &    90.18$\pm$0.62    &     88.69$\pm$0.66       \\ \cline{3-7} 
                               &                      & \multirow{4}{*}{CD}     & I        &    57.92$\pm$2.07      &    60.74$\pm$2.07    &    59.30$\pm$2.09       \\ \cline{4-7} 
                               &                      &                                     & IIIa     &     62.58$\pm$2.09      &  62.18$\pm$2.09      &    60.89$\pm$2.11     \\ \cline{4-7} 
                               &                 &                                    & IIIb     &     65.00$\pm$1.89      &    66.09$\pm$1.87    &    65.56$\pm$1.88         \\ \cline{4-7} 
                               &                      &                                     & IIIc     &    67.97$\pm$1.85       &    74.85$\pm$1.72    &   71.24$\pm$1.78       \\ \cline{2-7} 
                               &  \multirow{6}{*}{DCNN}& \multicolumn{2}{c|}{Normal}                    &     95.14$\pm$0.42      &    94.91$\pm$0.43    &  95.14$\pm$0.42              \\ \cline{3-7} 
                               &                      & \multicolumn{2}{c|}{EE} &    92.22$\pm$0.55       &     90.62$\pm$0.60   &    91.52$\pm$0.58         \\ \cline{3-7} 
                               &                      & \multirow{4}{*}{CD}     & I        &     75.41$\pm$1.82      &   72.63$\pm$1.89      &    73.99$\pm$1.85          \\ \cline{4-7} 
                               &                      &                                     & IIIa     &      70.81$\pm$1.92     &   72.47$\pm$1.93     &   71.63$\pm$1.79          \\ \cline{4-7} 
                               &                      &                                     & IIIb     &    81.08$\pm$0.81       &    74.67$\pm$1.84    &   77.74$\pm$1.65            \\ \cline{4-7} 
                               &                      &                                     & IIIc     &     75.07$\pm$1.83      &    76.37$\pm$1.81    &   75.71$\pm$1.81          \\ \hline
\multirow{12}{*}{\textbf{HMIC}} & \multirow{6}{*}{\textbf{NWS}} & \multicolumn{2}{c|}{\textbf{Normal}}                                        & \textbf{89.97$\pm$0.59 } & \textbf{ 89.35$\pm$0.61} & \textbf{ 89.66$\pm$0.61} \\ \cline{3-7} 
                                &                               & \multicolumn{2}{c|}{\textbf{EE}}  & \textbf{94.02$\pm$ 0.49} & \textbf{97.30$\pm$0.33} & \textbf{95.63$\pm$0.33} \\ \cline{3-7} 
                                &                                   & \multirow{4}{*}{\textbf{CD}} & \textbf{I}    & \textbf{83.25$\pm$1.58} & \textbf{80.91$\pm$1.66} & \textbf{82.06$\pm$1.62} \\ \cline{4-7} 
                                &                               &                                          & \textbf{IIIa} & \textbf{80.34$\pm$1.62} & \textbf{80.46$\pm$1.71} & \textbf{80.40$\pm$1.57} \\ \cline{4-7} 
                                &                               &                                          & \textbf{IIIb} & \textbf{85.35$\pm$1.49} & \textbf{81.77$\pm$1.67} & \textbf{83.52$\pm$1.47} \\ \cline{4-7} 
                                &                               &                                          & \textbf{IIIc} & \textbf{85.54$\pm$1.49} & \textbf{82.71$\pm$1.60} & \textbf{84.10$\pm$1.55} \\ \cline{2-7} 
                                
                                & \multirow{6}{*}{\textbf{WS}} & \multicolumn{2}{c|}{\textbf{Normal}}                     & \textbf{90.64$\pm$0.57 } & \textbf{ 90.06$\pm$0.57} & \textbf{ 90.35$\pm$0.58} \\ \cline{3-7} 
                                &                               & \multicolumn{2}{c|}{\textbf{EE}} & \textbf{94.08$\pm$ 0.49} & \textbf{97.33$\pm$0.42} & \textbf{98.68$\pm$0.42}  \\ \cline{3-7} 
                                &                               & \multirow{4}{*}{\textbf{CD}} & \textbf{I}    & \textbf{88.73$\pm$1.34} & \textbf{85.07$\pm$1.51} & \textbf{86.86$\pm$1.43} \\ \cline{4-7} 
                                &                               &                                          & \textbf{IIIa} & \textbf{81.19$\pm$1.65} & \textbf{81.19$\pm$1.65} & \textbf{82.44$\pm$1.51} \\ \cline{4-7} 
                                &                               &                                          & \textbf{IIIb} & \textbf{90.51$\pm$1.24} & \textbf{90.48$\pm$1.27} & \textbf{90.49$\pm$1.16} \\ \cline{4-7} 
                                &                               &                                          & \textbf{IIIc} & \textbf{89.26$\pm$1.31} & \textbf{90.18$\pm$1.26} & \textbf{89.72$\pm$1.28}                             \\                              \hline
\end{tabular}
\end{table}

\subsection{Empirical Results}
In this sub-section as we discussed in Section~\ref{sec:Evaluation}, we report precision, recall, and F1-score. 

Table~\ref{ta:parent} shows the results of the parent level model trained for classifying between Normal, Environmental Enteropathy (EE) and Celiac Disease (CD). The precision of normal patches is $89.97\pm0.5973$ and recall is $89.35\pm0.6133$. The F1-score of normal is $89.66\pm0.6054$. For EE,  precision is $94.02\pm0.4955$, recall is  $97.30\pm0.3385$, F1-score is $95.63\pm0.4270$. The CD evaluation measure for the parent level is as follows: precision is equal to $91.12\pm0.3208$, recall is equal to $88.71\pm0.3569$, and F1-score is equal to $89.90\pm1.2778$.

Table~\ref{ta:results_overall} shows the comparison of our techniques with three different baselines. The Search Results Web results from Convolutional Neural Network~(CNN), Deep Neural Network (Multilayer perceptron), and Deep Convolutional Neural Network (DCNN) are three baselines we are using in this dissertation section. Much research has been done in this domain such as ResNet, but these novel techniques can only handle small images such as $250\times 250$. In this dataset, we create $1000\time 1000$ patches, so we could not compare our work with ResNet, AlexNet, etc. Regarding precision, the highest is HMIC whole-slide with a mean of $88.01$ percent and a confidence interval of $0.3841$ followed by HMIC none whole-slide  $84.13$ percent and confidence interval of $0.3751$. The precision of CNN is $76.76\pm0.4985$, multilayer perceptron is $76.19\pm0.5030$, and DCNN is $82.95\pm0.4439$.  Regarding recall, the highest is HMIC whole-slide with a mean of $93.98$ percent and a confidence interval of $0.2811$ followed by HMIC non whole-slide at $93.56$ percent and confidence interval of $0.29.1$. The recall of CNN is $80.18\pm0.4706$, multilayer perceptron is $79.4\pm0.471$, and DCNN is $87.28\pm0.3933$. The highest F1-score is HMIC whole-slide with a mean of $90.89$ percent and a confidence interval of $0.3804$ followed by HMIC non whole-slide with $88.61$ percent and confidence interval of $0.3751$. The recall of CNN is $78.43\pm0.4855$, multilayer perceptron is $77.76\pm0.4911$, and DCNN is $85.06\pm0.4207$.

Table~\ref{ta:results_per_class} shows the results by each class. For Normal images, the best classifier is DCNN with $95.14\pm0.42$ recall of $94.91\pm0.43$ F1-score of $95.14\pm0.42$. For EE, HMIC is the best classifier. The whole-slide images classifier for parent level is more robust in comparison with non -whole slide with precision of $94.08\pm0.49$ Recall of $97.33\pm0.42$ F1-score of $98.68\pm0.42$. Although the normal and EE are very similar to flat models such as DCNN, CD contains 4 different stages and the margin is very high. The best flat model is DCNN with mean of F1-score of 73.99 for I, 71.63 for IIIa, 77.74 for IIIb, and 75.71 IIIc.

The Table~\ref{ta:results_per_class} indicates the margin for child level is very high even for the non whole-slide level of this dataset. The best results belong to the whole-slide classifier for parent level with precision with $88.73\pm1.34$ for I, $81.19\pm1.65$ for IIIa, $90.51\pm1.24$ for IIIb, $89.26\pm1.31$ for IIIc. The whole-slide classifier for parent level with recall with $85.07\pm1.51$ for I, $81.19\pm1.65$ for IIIa, $90.48\pm1.27$ for IIIb, $90.18\pm1.26$ for IIIc. The results of whole-slide classifier for parent level for recall is $85.07\pm1.51$ for I, $83.72\pm0.78$ for IIIa, $90.48\pm0.61$ for IIIb, $90.18\pm1.26$ for IIIc. Finally, The F1-score for whole-slide classifier for parent level is equal to $86.86\pm1.43$ for I, $82.44\pm1.51$ for IIIa, $90.49\pm1.16$ for IIIb, $89.72\pm1.28$.

\begin{figure}[!t]
    \centering
    \includegraphics[width=\columnwidth]{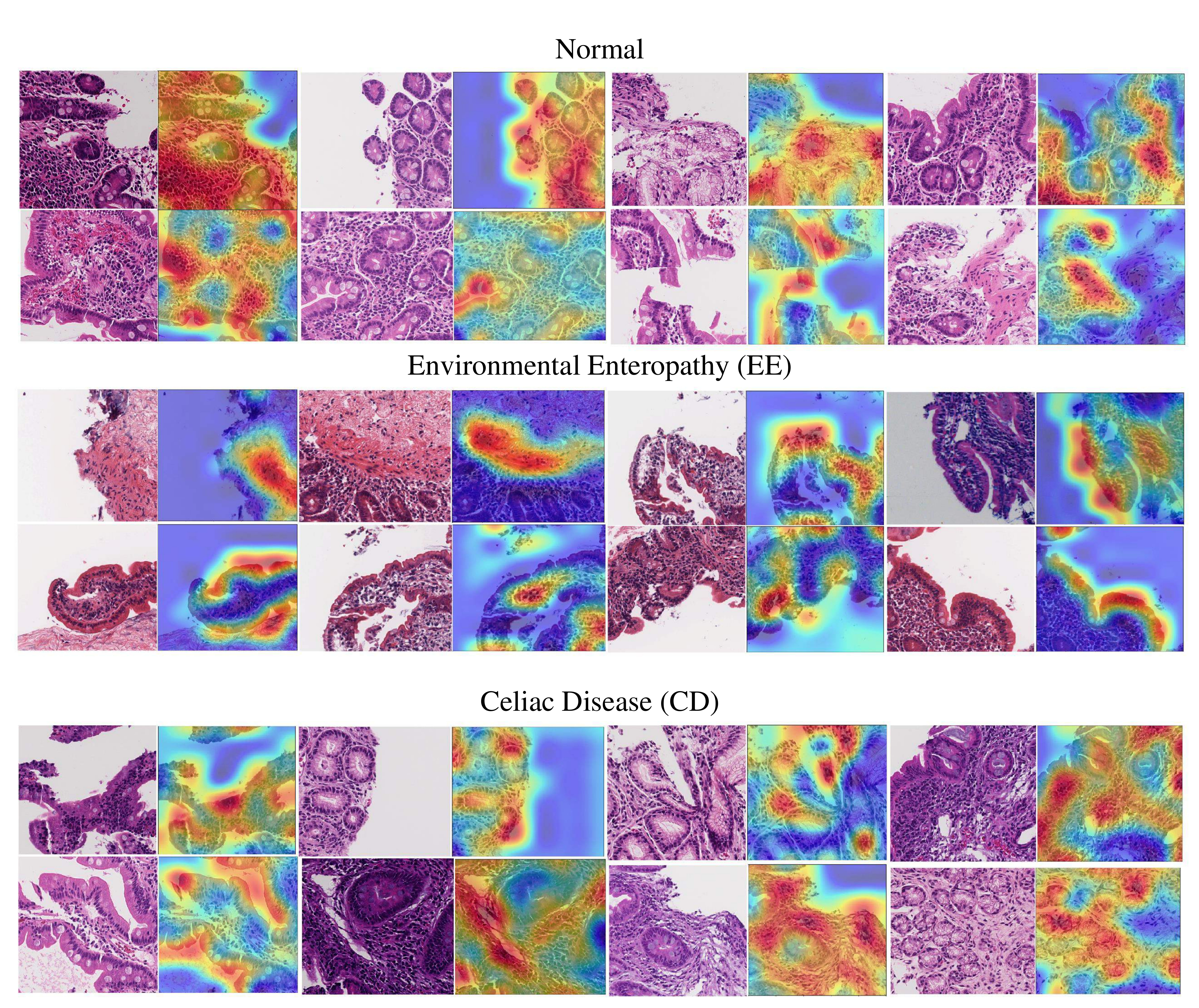}
    \caption{Results of Grad-CAM for showing feature importance.} \label{fig:V_Results}
\end{figure}

\subsection{Visualization}

As shown in Figure~\ref{fig:V_Results}, Grad-CAMs were generated for 41 patches (18 EE, 14 Celiac Disease, and 9 histologically normal duodenal controls) which mainly focused on distinct, yet medically relevant cellular features outlined below . Although, most heatmaps focused on medically relevant features, there were some patches that focused on too many features (n=8) or focused on connective tissue debris (n=10) that we were unable to categorize.
\begin{itemize}
    \item EE: surface epithelium with IELs and goblet cells was highlighted. Within the lamina propria, the heatmaps also focused on mononuclear cells. 
	\item CD: heatmaps highlighted the edge of crypt cross sections, surface epithelium with IELs and goblet cells, and areas with mononuclear cells within the lamina propria. 
	\item Histologically Normal: surface epithelium with epithelial cells containing abundant cytoplasm was highlighted. 
\end{itemize}


\section{Conclusion}\label{sec:Conclusion_6}
Medical image classification is an important problem to address, given the growing medical instrument to collect digital images. When medical images are organize hierarchically, multi-class approaches are difficult to apply using traditional supervised learning methods. This paper introduces a new approach to hierarchical medical image classification, HMIC, that combines multiple deep learning approaches to produce hierarchical classifications. Testing on a data set of of biopsy image patches shows that this technique at the higher and lower level produced accuracies consistently higher than those obtainable by conventional approaches using CNN, Multilayer perceptron, and DCNN. These results show that hierarchical deep learning methods can provide improvements for classification and that they provide flexibility to classify these data within a hierarchy. Hence, they provide extensions over current methods that only consider the multi-class problem.

The methods described here can be improved in multiple ways. Additional training and testing with other hierarchically structured medical data sets will continue to identify architectures that work best for these problems. Also, it is possible to extend the hierarchy to more than two levels to capture more of the complexity in the hierarchical classification. For example, if stage of disease are treated as ordered then the hierarchy continues down multiple levels. Scoring here could be performed on small sets using human judges.


\chapter{Conclusions \& Future Directions}\label{chpt:Conclusions}

\section{Summary of Contributions}
In this research, we approached four techniques to diagnosis of celiac disease and Environmental Enteropathy on biopsy images that fist technique Color Balancing~(CB) is used on Convolutional Neural Networks~(CNN), the second technique is using Random Multimodel Deep Learning~(RMDL) to the diagnosis of celiac disease and Environmental Enteropathy. The third technique is used to diagnosis only in the stage of Celiac Disease Severity Diagnosis on Duodenal Histopathological images using Deep Learning~(DL) which we approach two techniques shallow convolutional neural network (CNN) and deep residual networks to diagnose  CD  severity using a  histological scoring system called the modified Marsh score. Finally, we used an approach called Hierarchical Medical Image classification (HMIC) to combine these models as the hierarchical representations of theses biopsy images.
\subsection{Two approaches for diagnosing celiac disease and Environmental Enteropathy}
In this dissertation, two main ideas have been covered to the diagnosis of diseased duodenal architecture on biopsy images.  we proposed a data-driven model for diagnosis of diseased duodenal architecture on biopsy images using color balancing on convolutional neural networks. The validation results of this model show that it can be utilized by pathologists in diagnostic operations regarding CD and EE. 
The second approach is called RMDL (Random Multimodel Deep Learning) for the medical image classification that combines multi deep learning approaches to produce random classification models. Our evaluation on datasets obtained from the biopsy images shows that combinations of DNNs, RNNs, and CNNs with the parallel learning architecture, has consistently higher accuracy, but Model interpretability of deep learning (DL), especially RMDL, has always been a limiting factor for use cases requiring explanations of the features involved in modeling and such is the case for many healthcare problems.

\subsection{Two Approaches for Diagnosis of Duodenal Histopathological images Celiac Disease}
we investigated CD severity using CNNs and residual neural network architecture applied to histopathological images in which a state-of-the-art deep neural network architecture was used to categorize patients based on H\&E stained duodenal histopathological images into four classes, representing different CD severity based on a histological classification called the modified Marsh score. albeit achieving promising results, this study has a number of limitations. Firstly, healthy cases were not included in this study. This is an avenue for future work. In addition, all biopsy images used in this study were collected from a single medical center and scanned with the same equipment, thus our data may not be representative of the entire range of histopathologic patterns in patients worldwide.
\subsection{Hierarchical Medical Image Classification (HMIC)}
In this dissertation, we proposed a technique called Hierarchical Medical Image Classification (HMIC), that could combine multiple deep learning approaches to produce hierarchical classifications. Testing on a data set of these biopsy images obtained from the biopsy images shows that this technique at the parents and child level produced accuracies consistently higher than those obtainable by conventional approaches using CNN, DNN, and DCNN. The method that is presented in this dissertation described here can be improved in multiple ways. Additional training and testing with other hierarchically structured medical data sets will continue to identify architectures that work best for these problems. Also, it is possible to extend the hierarchy to more than two levels to capture more of the complexity in the hierarchical classification.

\section{Summary of Future Works}
it is possible to extend this research in multiple ways which include as following ways. As we discussed in Chapter~\ref{chpt:Celiac_color}, we could use more color balancing range to make the trained model completely color independent which needs powerful hardware and resource. As shown in Chapter~\ref{chpt:RMDL}, we could use more and more models and also, in future we need to run without resizing data from $1000\times 1000$ to $100\times100$. For Chapter~\ref{chpt:HMIC}, the hierarchy to more than two levels to capture more of the complexity in the hierarchical classification. For example, if the stage of the disease is treated as ordered then the hierarchy continues down multiple levels. Scoring here could be performed on small sets using human judges.

\bibliographystyle{ieeetr}
\bibliography{ref}

\begin{thebibliography}{100}

\bibitem{wei2019automated}
J.~W. Wei, J.~W. Wei, C.~R. Jackson, B.~Ren, A.~A. Suriawinata, and
  S.~Hassanpour, ``Automated detection of celiac disease on duodenal biopsy
  slides: A deep learning approach,'' {\em arXiv preprint arXiv:1901.11447},
  2019.

\bibitem{syed2019assessment}
S.~Syed, M.~Al-Boni, M.~N. Khan, K.~Sadiq, N.~T. Iqbal, C.~A. Moskaluk,
  P.~Kelly, B.~Amadi, S.~A. Ali, S.~R. Moore, {\em et~al.}, ``Assessment of
  machine learning detection of environmental enteropathy and celiac disease in
  children,'' {\em JAMA network open}, vol.~2, no.~6, pp.~e195822--e195822,
  2019.

\bibitem{ung2010mercury}
C.~Y. Ung, S.~H. Lam, M.~M. Hlaing, C.~L. Winata, S.~Korzh, S.~Mathavan, and
  Z.~Gong, ``Mercury-induced hepatotoxicity in zebrafish: in vivo mechanistic
  insights from transcriptome analysis, phenotype anchoring and targeted gene
  expression validation,'' {\em Bmc Genomics}, vol.~11, no.~1, p.~212, 2010.

\bibitem{fasano2001current}
A.~Fasano and C.~Catassi, ``Current approaches to diagnosis and treatment of
  celiac disease: an evolving spectrum,'' {\em Gastroenterology}, vol.~120,
  no.~3, pp.~636--651, 2001.

\bibitem{vahadane2016structure}
A.~Vahadane, T.~Peng, A.~Sethi, S.~Albarqouni, L.~Wang, M.~Baust, K.~Steiger,
  A.~M. Schlitter, I.~Esposito, and N.~Navab, ``Structure-preserving color
  normalization and sparse stain separation for histological images,'' {\em
  IEEE transactions on medical imaging}, vol.~35, no.~8, pp.~1962--1971, 2016.

\bibitem{kowsari2019diagnosis}
K.~Kowsari, R.~Sali, M.~N. Khan, W.~Adorno, S.~A. Ali, S.~R. Moore, B.~C.
  Amadi, P.~Kelly, S.~Syed, and D.~E. Brown, ``Diagnosis of celiac disease and
  environmental enteropathy on biopsy images using color balancing on
  convolutional neural networks,'' 2019.

\bibitem{sali2019celiacnet}
R.~Sali, L.~Ehsan, K.~Kowsari, M.~Khan, C.~A. Moskaluk, S.~Syed, and D.~E.
  Brown, ``Celiacnet: Celiac disease severity diagnosis on duodenal
  histopathological images using deep residual networks,'' {\em arXiv preprint
  arXiv:1910.03084}, 2019.

\bibitem{iqbal2019study}
N.~T. Iqbal, S.~Syed, K.~Sadiq, M.~N. Khan, J.~Iqbal, J.~Z. Ma, F.~Umrani,
  S.~Ahmed, E.~A. Maier, L.~A. Denson, {\em et~al.}, ``Study of environmental
  enteropathy and malnutrition (seem) in pakistan: protocols for biopsy based
  biomarker discovery and validation,'' {\em BMC pediatrics}, vol.~19, no.~1,
  p.~247, 2019.

\bibitem{gulshan2016development}
V.~Gulshan, L.~Peng, M.~Coram, M.~C. Stumpe, D.~Wu, A.~Narayanaswamy,
  S.~Venugopalan, K.~Widner, T.~Madams, J.~Cuadros, {\em et~al.}, ``Development
  and validation of a deep learning algorithm for detection of diabetic
  retinopathy in retinal fundus photographs,'' {\em Jama}, vol.~316, no.~22,
  pp.~2402--2410, 2016.

\bibitem{litjens2017survey}
G.~Litjens, T.~Kooi, B.~E. Bejnordi, A.~A.~A. Setio, F.~Ciompi, M.~Ghafoorian,
  J.~A. Van Der~Laak, B.~Van~Ginneken, and C.~I. S{\'a}nchez, ``A survey on
  deep learning in medical image analysis,'' {\em Medical image analysis},
  vol.~42, pp.~60--88, 2017.

\bibitem{Mohammad_al_boni}
M.~Al~Boni, S.~Syed, A.~Ali, S.~R. Moore, and D.~E. Brown, ``Duodenal biopsies
  classification and understanding using convolutionalneural networks,'' {\em
  American Medical Informatics Association}, 2019.

\bibitem{WHO.Children}
``Who. children: reducing mortality. fact sheet 2017.''
  \url{http://www.who.int/mediacentre/factsheets/fs178/en/}.
\newblock Accessed: 2019-1-30.

\bibitem{syed2016environmental}
S.~Syed, A.~Ali, and C.~Duggan, ``Environmental enteric dysfunction in
  children: a review,'' {\em Journal of pediatric gastroenterology and
  nutrition}, vol.~63, no.~1, p.~6, 2016.

\bibitem{naylor2015environmental}
C.~Naylor, M.~Lu, R.~Haque, D.~Mondal, E.~Buonomo, U.~Nayak, J.~C. Mychaleckyj,
  B.~Kirkpatrick, R.~Colgate, M.~Carmolli, {\em et~al.}, ``Environmental
  enteropathy, oral vaccine failure and growth faltering in infants in
  bangladesh,'' {\em EBioMedicine}, vol.~2, no.~11, pp.~1759--1766, 2015.

\bibitem{husby2012european}
S.~Husby {\em et~al.}, ``European society for pediatric gastroenterology,
  hepatology, and nutrition guidelines for the diagnosis of coeliac disease,''
  {\em Journal of pediatric gastroenterology and nutrition}, vol.~54, no.~1,
  pp.~136--160, 2012.

\bibitem{bejnordi2017diagnostic}
B.~E. Bejnordi {\em et~al.}, ``Diagnostic assessment of deep learning
  algorithms for detection of lymph node metastases in women with breast
  cancer,'' {\em Jama}, vol.~318, no.~22, pp.~2199--2210, 2017.

\bibitem{Heidarysafa2018RMDL}
M.~Heidarysafa, K.~Kowsari, D.~E. Brown, K.~Jafari~Meimandi, and L.~E. Barnes,
  ``An improvement of data classification using random multimodel deep learning
  (rmdl),'' {\em International Journal of Machine Learning and Computing
  (IJMLC)}, vol.~8, no.~4, pp.~298--310, 2018.

\bibitem{kowsari2017hdltex}
K.~Kowsari, D.~E. Brown, M.~Heidarysafa, K.~J. Meimandi, M.~S. Gerber, and
  L.~E. Barnes, ``Hdltex: Hierarchical deep learning for text classification,''
  in {\em 2017 16th IEEE international conference on machine learning and
  applications (ICMLA)}, pp.~364--371, IEEE, 2017.

\bibitem{kowsari2018rmdl}
K.~Kowsari, M.~Heidarysafa, D.~E. Brown, K.~Jafari~Meimandi, and L.~E. Barnes,
  ``Rmdl: Random multimodel deep learning for classification,'' in {\em
  Proceedings of the 2018 International Conference on Information System and
  Data Mining}, ACM, 2018.

\bibitem{heidarysafa2018analysis}
M.~Heidarysafa, K.~Kowsari, L.~E. Barnes, and D.~E. Brown, ``Analysis of
  railway accidents' narratives using deep learning,'' 2018.

\bibitem{info10040150}
K.~Kowsari, K.~Jafari~Meimandi, M.~Heidarysafa, S.~Mendu, L.~Barnes, and
  D.~Brown, ``Text classification algorithms: A survey,'' {\em Information},
  vol.~10, no.~4, 2019.

\bibitem{nobles2018identification}
A.~L. Nobles, J.~J. Glenn, K.~Kowsari, B.~A. Teachman, and L.~E. Barnes,
  ``Identification of imminent suicide risk among young adults using text
  messages,'' in {\em Proceedings of the 2018 CHI Conference on Human Factors
  in Computing Systems}, pp.~1--11, 2018.

\bibitem{zhai2016doubly}
S.~Zhai, Y.~Cheng, Z.~M. Zhang, and W.~Lu, ``Doubly convolutional neural
  networks,'' in {\em Advances in neural information processing systems},
  pp.~1082--1090, 2016.

\bibitem{kowsari2020gender}
K.~Kowsari, M.~Heidarysafa, T.~Odukoya, P.~Potter, L.~E. Barnes, and D.~E.
  Brown, ``Gender detection on social networks using ensemble deep learning,''
  {\em arXiv preprint arXiv:2004.06518}, 2020.

\bibitem{hegde2019comparison}
R.~B. Hegde, K.~Prasad, H.~Hebbar, and B.~M.~K. Singh, ``Comparison of
  traditional image processing and deep learning approaches for classification
  of white blood cells in peripheral blood smear images,'' {\em Biocybernetics
  and Biomedical Engineering}, 2019.

\bibitem{zhang2018patient2vec}
J.~Zhang, K.~Kowsari, J.~H. Harrison, J.~M. Lobo, and L.~E. Barnes,
  ``Patient2vec: A personalized interpretable deep representation of the
  longitudinal electronic health record,'' {\em IEEE Access}, vol.~6,
  pp.~65333--65346, 2018.

\bibitem{ker2018deep}
J.~Ker, L.~Wang, J.~Rao, and T.~Lim, ``Deep learning applications in medical
  image analysis,'' {\em IEEE Access}, vol.~6, pp.~9375--9389, 2018.

\bibitem{nawaz2018classification}
W.~Nawaz, S.~Ahmed, A.~Tahir, and H.~A. Khan, ``Classification of breast cancer
  histology images using alexnet,'' in {\em International Conference Image
  Analysis and Recognition}, pp.~869--876, Springer, 2018.

\bibitem{al2019duodenal}
M.~Al~Boni, S.~Syed, A.~Ali, S.~R. Moore, and D.~E. Brown, ``Duodenal biopsies
  classification and understanding using convolutional neural networks,'' {\em
  AMIA Summits on Translational Science Proceedings}, vol.~2019, p.~453, 2019.

\bibitem{ermarth2017identification}
A.~Ermarth, M.~Bryce, S.~Woodward, G.~Stoddard, L.~Book, and M.~K. Jensen,
  ``Identification of pediatric patients with celiac disease based on serology
  and a classification and regression tree analysis,'' {\em Clinical
  Gastroenterology and Hepatology}, vol.~15, no.~3, pp.~396--402, 2017.

\bibitem{lim2017failure}
H.~K. Lim, Y.~Kim, and M.-K. Kim, ``Failure prediction using sequential pattern
  mining in the wire bonding process,'' {\em IEEE Transactions on Semiconductor
  Manufacturing}, vol.~30, no.~3, pp.~285--292, 2017.

\bibitem{kowal2014computer}
M.~Kowal, ``Computer-aided diagnosis for breast tumor classification using
  microscopic images of fine needle biopsy,'' in {\em Intelligent Systems in
  Technical and Medical Diagnostics}, pp.~213--224, Springer, 2014.

\bibitem{radiya2017automated}
E.~Radiya-Dixit, D.~Zhu, and A.~H. Beck, ``Automated classification of benign
  and malignant proliferative breast lesions,'' {\em Scientific reports},
  vol.~7, no.~1, pp.~1--8, 2017.

\bibitem{hegenbart2011impact}
S.~Hegenbart, A.~Uhl, and A.~V{\'e}csei, ``Impact of endoscopic image
  degradations on lbp based features using one-class svm for classification of
  celiac disease,'' in {\em 2011 7th International Symposium on Image and
  Signal Processing and Analysis (ISPA)}, pp.~715--720, IEEE, 2011.

\bibitem{fathi2014differential}
F.~Fathi, L.~M. Kasmaee, K.~Sohrabzadeh, M.~R. Nejad, M.~Tafazzoli, and A.~A.
  Oskouie, ``The differential diagnosis of crohn’s disease and celiac disease
  using nuclear magnetic resonance spectroscopy,'' {\em Applied Magnetic
  Resonance}, vol.~45, no.~5, pp.~451--459, 2014.

\bibitem{habtamu2015sensitive}
H.~B. Habtamu, M.~Sentic, M.~Silvestrini, L.~De~Leo, T.~Not, S.~Arbault,
  D.~Manojlovic, N.~Sojic, and P.~Ugo, ``A sensitive electrochemiluminescence
  immunosensor for celiac disease diagnosis based on nanoelectrode ensembles,''
  {\em Analytical chemistry}, vol.~87, no.~24, pp.~12080--12087, 2015.

\bibitem{lionetti2014introduction}
E.~Lionetti, S.~Castellaneta, R.~Francavilla, A.~Pulvirenti, E.~Tonutti,
  S.~Amarri, M.~Barbato, C.~Barbera, G.~Barera, A.~Bellantoni, {\em et~al.},
  ``Introduction of gluten, hla status, and the risk of celiac disease in
  children,'' {\em New England Journal of Medicine}, vol.~371, no.~14,
  pp.~1295--1303, 2014.

\bibitem{pagadala2013diagnosis}
M.~R. Pagadala, M.~S. Lemyre, R.~Lopez, A.~Kumaravel, W.~D. Carey, N.~N. Zein,
  {\em et~al.}, ``Diagnosis of celiac disease in adults based on serology test
  results, without small-bowel biopsy,'' {\em Clinical Gastroenterology and
  Hepatology}, vol.~11, no.~5, pp.~511--516, 2013.

\bibitem{syed2018195}
S.~Syed, M.~Al-Boni, K.~Sadiq, N.~T. Iqbal, C.~A. Moskaluk, A.~Ali, S.~Moore,
  and D.~Brown, ``195-convolutional neural networks image analysis of duodenal
  biopsies robustly distinguishes environmental enteropathy from healthy
  controls and identifies secretory cell lineages as high activation
  locations,'' {\em Gastroenterology}, vol.~154, no.~6, pp.~S--52, 2018.

\bibitem{murphy2006naive}
K.~P. Murphy, ``Naive bayes classifiers,'' {\em University of British
  Columbia}, 2006.

\bibitem{rish2001empirical}
I.~Rish, ``An empirical study of the naive bayes classifier,'' in {\em IJCAI
  2001 workshop on empirical methods in artificial intelligence}, vol.~3,
  pp.~41--46, IBM, 2001.

\bibitem{mccann2012local}
S.~McCann and D.~G. Lowe, ``Local naive bayes nearest neighbor for image
  classification,'' in {\em 2012 IEEE Conference on Computer Vision and Pattern
  Recognition}, pp.~3650--3656, IEEE, 2012.

\bibitem{yu2009learning}
C.-N.~J. Yu and T.~Joachims, ``Learning structural svms with latent
  variables,'' in {\em Proceedings of the 26th annual international conference
  on machine learning}, pp.~1169--1176, ACM, 2009.

\bibitem{tong2001support}
S.~Tong and D.~Koller, ``Support vector machine active learning with
  applications to text classification,'' {\em Journal of machine learning
  research}, vol.~2, no.~Nov, pp.~45--66, 2001.

\bibitem{kabir2015bangla}
F.~Kabir, S.~Siddique, M.~R.~A. Kotwal, and M.~N. Huda, ``Bangla text document
  categorization using stochastic gradient descent (sgd) classifier,'' in {\em
  Cognitive Computing and Information Processing (CCIP), 2015 International
  Conference on}, pp.~1--4, IEEE, 2015.

\bibitem{cox2018analysis}
D.~R. Cox, {\em Analysis of binary data}.
\newblock Routledge, 2018.

\bibitem{fan2008liblinear}
R.-E. Fan, K.-W. Chang, C.-J. Hsieh, X.-R. Wang, and C.-J. Lin, ``Liblinear: A
  library for large linear classification,'' {\em Journal of machine learning
  research}, vol.~9, no.~Aug, pp.~1871--1874, 2008.

\bibitem{genkin2007large}
A.~Genkin, D.~D. Lewis, and D.~Madigan, ``Large-scale bayesian logistic
  regression for text categorization,'' {\em Technometrics}, vol.~49, no.~3,
  pp.~291--304, 2007.

\bibitem{rao2011classification}
A.~Rao, Y.~Lee, A.~Gass, and A.~Monsch, ``Classification of alzheimer's disease
  from structural mri using sparse logistic regression with optional spatial
  regularization,'' in {\em 2011 Annual International Conference of the IEEE
  Engineering in Medicine and Biology Society}, pp.~4499--4502, IEEE, 2011.

\bibitem{juan2002use}
A.~Juan and E.~Vidal, ``On the use of bernoulli mixture models for text
  classification,'' {\em Pattern Recognition}, vol.~35, no.~12, pp.~2705--2710,
  2002.

\bibitem{cheng2009combining}
W.~Cheng and E.~H{\"u}llermeier, ``Combining instance-based learning and
  logistic regression for multilabel classification,'' {\em Machine Learning},
  vol.~76, no.~2-3, pp.~211--225, 2009.

\bibitem{krishnapuram2005sparse}
B.~Krishnapuram, L.~Carin, M.~A. Figueiredo, and A.~J. Hartemink, ``Sparse
  multinomial logistic regression: Fast algorithms and generalization bounds,''
  {\em IEEE transactions on pattern analysis and machine intelligence},
  vol.~27, no.~6, pp.~957--968, 2005.

\bibitem{jiang2012improved}
S.~Jiang, G.~Pang, M.~Wu, and L.~Kuang, ``An improved k-nearest-neighbor
  algorithm for text categorization,'' {\em Expert Systems with Applications},
  vol.~39, no.~1, pp.~1503--1509, 2012.

\bibitem{vapnik1964class}
V.~Vapnik and A.~Y. Chervonenkis, ``A class of algorithms for pattern
  recognition learning,'' {\em Avtomat. i Telemekh}, vol.~25, no.~6,
  pp.~937--945, 1964.

\bibitem{boser1992training}
B.~E. Boser, I.~M. Guyon, and V.~N. Vapnik, ``A training algorithm for optimal
  margin classifiers,'' in {\em Proceedings of the fifth annual workshop on
  Computational learning theory}, pp.~144--152, ACM, 1992.

\bibitem{bo2006svm}
G.~Bo and H.~Xianwu, ``Svm multi-class classification [j],'' {\em Journal of
  Data Acquisition \& Processing}, vol.~3, p.~017, 2006.

\bibitem{camlica2015medical}
Z.~Camlica, H.~R. Tizhoosh, and F.~Khalvati, ``Medical image classification via
  svm using lbp features from saliency-based folded data,'' in {\em 2015 IEEE
  14th International Conference on Machine Learning and Applications (ICMLA)},
  pp.~128--132, IEEE, 2015.

\bibitem{manevitz2001one}
L.~M. Manevitz and M.~Yousef, ``One-class svms for document classification,''
  {\em Journal of machine Learning research}, vol.~2, no.~Dec, pp.~139--154,
  2001.

\bibitem{mohri2012foundations}
M.~Mohri, A.~Rostamizadeh, and A.~Talwalkar, {\em Foundations of machine
  learning}.
\newblock MIT press, 2012.

\bibitem{chen2016turning}
K.~Chen, Z.~Zhang, J.~Long, and H.~Zhang, ``Turning from tf-idf to tf-igm for
  term weighting in text classification,'' {\em Expert Systems with
  Applications}, vol.~66, pp.~245--260, 2016.

\bibitem{weston1998multi}
J.~Weston and C.~Watkins, ``Multi-class support vector machines,'' tech. rep.,
  Technical Report CSD-TR-98-04, Department of Computer Science, Royal
  Holloway, University of London, May, 1998.

\bibitem{morgan1963problems}
J.~N. Morgan and J.~A. Sonquist, ``Problems in the analysis of survey data, and
  a proposal,'' {\em Journal of the American statistical association}, vol.~58,
  no.~302, pp.~415--434, 1963.

\bibitem{safavian1991survey}
S.~R. Safavian and D.~Landgrebe, ``A survey of decision tree classifier
  methodology,'' {\em IEEE transactions on systems, man, and cybernetics},
  vol.~21, no.~3, pp.~660--674, 1991.

\bibitem{aggarwal2012survey}
C.~C. Aggarwal and C.~Zhai, ``A survey of text classification algorithms,'' in
  {\em Mining text data}, pp.~163--222, Springer, 2012.

\bibitem{magerman1995statistical}
D.~M. Magerman, ``Statistical decision-tree models for parsing,'' in {\em
  Proceedings of the 33rd annual meeting on Association for Computational
  Linguistics}, pp.~276--283, Association for Computational Linguistics, 1995.

\bibitem{quinlan1986induction}
J.~R. Quinlan, ``Induction of decision trees,'' {\em Machine learning}, vol.~1,
  no.~1, pp.~81--106, 1986.

\bibitem{de1991distance}
R.~L. De~M{\'a}ntaras, ``A distance-based attribute selection measure for
  decision tree induction,'' {\em Machine learning}, vol.~6, no.~1, pp.~81--92,
  1991.

\bibitem{farzi2016estimation}
R.~Farzi and V.~Bolandi, ``Estimation of organic facies using ensemble methods
  in comparison with conventional intelligent approaches: a case study of the
  south pars gas field, persian gulf, iran,'' {\em Modeling Earth Systems and
  Environment}, vol.~2, no.~2, p.~105, 2016.

\bibitem{bauer1999empirical}
E.~Bauer and R.~Kohavi, ``An empirical comparison of voting classification
  algorithms: Bagging, boosting, and variants,'' {\em Machine learning},
  vol.~36, no.~1-2, pp.~105--139, 1999.

\bibitem{subudhi2020automated}
A.~Subudhi, M.~Dash, and S.~Sabut, ``Automated segmentation and classification
  of brain stroke using expectation-maximization and random forest
  classifier,'' {\em Biocybernetics and Biomedical Engineering}, vol.~40,
  no.~1, pp.~277--289, 2020.

\bibitem{Ho1995RF}
T.~K. Ho, ``Random decision forests,'' in {\em Proceedings of 3rd International
  Conference on Document Analysis and Recognition}, vol.~1, pp.~278--282 vol.1,
  Aug 1995.

\bibitem{breiman1999random}
L.~Breiman, ``Random forests,'' {\em UC Berkeley TR567}, 1999.

\bibitem{wu2004probability}
T.-F. Wu, C.-J. Lin, and R.~C. Weng, ``Probability estimates for multi-class
  classification by pairwise coupling,'' {\em Journal of Machine Learning
  Research}, vol.~5, no.~Aug, pp.~975--1005, 2004.

\bibitem{schapire1990strength}
R.~E. Schapire, ``The strength of weak learnability,'' {\em Machine learning},
  vol.~5, no.~2, pp.~197--227, 1990.

\bibitem{freund1992improved}
Y.~Freund, ``An improved boosting algorithm and its implications on learning
  complexity,'' in {\em Proceedings of the fifth annual workshop on
  Computational learning theory}, pp.~391--398, ACM, 1992.

\bibitem{bloehdorn2004boosting}
S.~Bloehdorn and A.~Hotho, ``Boosting for text classification with semantic
  features,'' in {\em International workshop on knowledge discovery on the
  web}, pp.~149--166, Springer, 2004.

\bibitem{freund1995efficient}
Y.~Freund, M.~Kearns, Y.~Mansour, D.~Ron, R.~Rubinfeld, and R.~E. Schapire,
  ``Efficient algorithms for learning to play repeated games against
  computationally bounded adversaries,'' in {\em Foundations of Computer
  Science, 1995. Proceedings., 36th Annual Symposium on}, pp.~332--341, IEEE,
  1995.

\bibitem{breiman1996bagging}
L.~Breiman, ``Bagging predictors,'' {\em Machine learning}, vol.~24, no.~2,
  pp.~123--140, 1996.

\bibitem{ci2012multitraffic}
D.~Cire{\c{s}}An, U.~Meier, J.~Masci, and J.~Schmidhuber, ``Multi-column deep
  neural network for traffic sign classification,'' {\em Neural Networks},
  vol.~32, pp.~333--338, 2012.

\bibitem{krizhevsky2012imagenet}
A.~Krizhevsky, I.~Sutskever, and G.~E. Hinton, ``Imagenet classification with
  deep convolutional neural networks,'' in {\em Advances in neural information
  processing systems}, pp.~1097--1105, 2012.

\bibitem{lecun2015deep}
Y.~LeCun, Y.~Bengio, and G.~Hinton, ``Deep learning,'' {\em nature}, vol.~521,
  no.~7553, pp.~436--444, 2015.

\bibitem{chung2014empirical}
J.~Chung, C.~Gulcehre, K.~Cho, and Y.~Bengio, ``Empirical evaluation of gated
  recurrent neural networks on sequence modeling,'' {\em arXiv preprint
  arXiv:1412.3555}, 2014.

\bibitem{hochreiter1997long}
S.~Hochreiter and J.~Schmidhuber, ``Long short-term memory,'' {\em Neural
  computation}, vol.~9, no.~8, pp.~1735--1780, 1997.

\bibitem{mikolov2010recurrent}
T.~Mikolov, M.~Karafi{\'a}t, L.~Burget, J.~{\v{C}}ernock{\`y}, and
  S.~Khudanpur, ``Recurrent neural network based language model,'' in {\em
  Eleventh Annual Conference of the International Speech Communication
  Association}, 2010.

\bibitem{mao2016hierarchical}
Y.~Mao and Z.~Yin, ``A hierarchical convolutional neural network for mitosis
  detection in phase-contrast microscopy images,'' in {\em International
  Conference on Medical Image Computing and Computer-Assisted Intervention},
  pp.~685--692, Springer, 2016.

\bibitem{turan2017deep}
M.~Turan, Y.~Almalioglu, H.~Araujo, E.~Konukoglu, and M.~Sitti, ``Deep endovo:
  A recurrent convolutional neural network (rcnn) based visual odometry
  approach for endoscopic capsule robots,'' {\em arXiv preprint
  arXiv:1708.06822}, 2017.

\bibitem{liang2015recurrent}
M.~Liang and X.~Hu, ``Recurrent convolutional neural network for object
  recognition,'' in {\em Proceedings of the IEEE Conference on Computer Vision
  and Pattern Recognition}, pp.~3367--3375, 2015.

\bibitem{goodfellow2016deep}
I.~Goodfellow, Y.~Bengio, A.~Courville, and Y.~Bengio, {\em Deep learning},
  vol.~1.
\newblock MIT press Cambridge, 2016.

\bibitem{wang2014generalized}
W.~Wang, Y.~Huang, Y.~Wang, and L.~Wang, ``Generalized autoencoder: A neural
  network framework for dimensionality reduction,'' in {\em Proceedings of the
  IEEE conference on computer vision and pattern recognition workshops},
  pp.~490--497, 2014.

\bibitem{rumelhart1985learning}
D.~E. Rumelhart, G.~E. Hinton, and R.~J. Williams, ``Learning internal
  representations by error propagation,'' tech. rep., California Univ San Diego
  La Jolla Inst for Cognitive Science, 1985.

\bibitem{liang2017text}
H.~Liang, X.~Sun, Y.~Sun, and Y.~Gao, ``Text feature extraction based on deep
  learning: a review,'' {\em EURASIP journal on wireless communications and
  networking}, vol.~2017, no.~1, p.~211, 2017.

\bibitem{baldi2012autoencoders}
P.~Baldi, ``Autoencoders, unsupervised learning, and deep architectures,'' in
  {\em Proceedings of ICML Workshop on Unsupervised and Transfer Learning},
  pp.~37--49, 2012.

\bibitem{ap2014autoencoder}
S.~C. AP, S.~Lauly, H.~Larochelle, M.~Khapra, B.~Ravindran, V.~C. Raykar, and
  A.~Saha, ``An autoencoder approach to learning bilingual word
  representations,'' in {\em Advances in Neural Information Processing
  Systems}, pp.~1853--1861, 2014.

\bibitem{masci2011stacked}
J.~Masci, U.~Meier, D.~Cire{\c{s}}an, and J.~Schmidhuber, ``Stacked
  convolutional auto-encoders for hierarchical feature extraction,'' in {\em
  International Conference on Artificial Neural Networks}, pp.~52--59,
  Springer, 2011.

\bibitem{chen2015page}
K.~Chen, M.~Seuret, M.~Liwicki, J.~Hennebert, and R.~Ingold, ``Page
  segmentation of historical document images with convolutional autoencoders,''
  in {\em Document Analysis and Recognition (ICDAR), 2015 13th International
  Conference on}, pp.~1011--1015, IEEE, 2015.

\bibitem{geng2015high}
J.~Geng, J.~Fan, H.~Wang, X.~Ma, B.~Li, and F.~Chen, ``High-resolution sar
  image classification via deep convolutional autoencoders,'' {\em IEEE
  Geoscience and Remote Sensing Letters}, vol.~12, no.~11, pp.~2351--2355,
  2015.

\bibitem{sutskever2014sequence}
I.~Sutskever, O.~Vinyals, and Q.~V. Le, ``Sequence to sequence learning with
  neural networks,'' in {\em Advances in neural information processing
  systems}, pp.~3104--3112, 2014.

\bibitem{cho2014learning}
K.~Cho, B.~Van~Merri{\"e}nboer, C.~Gulcehre, D.~Bahdanau, F.~Bougares,
  H.~Schwenk, and Y.~Bengio, ``Learning phrase representations using rnn
  encoder-decoder for statistical machine translation,'' {\em arXiv preprint
  arXiv:1406.1078}, 2014.

\bibitem{fischer2008hematoxylin}
A.~H. Fischer, K.~A. Jacobson, J.~Rose, and R.~Zeller, ``Hematoxylin and eosin
  staining of tissue and cell sections,'' {\em Cold spring harbor protocols},
  vol.~2008, no.~5, pp.~pdb--prot4986, 2008.

\bibitem{anderson2011introduction}
J.~Anderson, ``An introduction to routine and special staining,'' {\em
  Retrieved on August}, vol.~18, p.~2014, 2011.

\bibitem{khan2014nonlinear}
A.~M. Khan, N.~Rajpoot, D.~Treanor, and D.~Magee, ``A nonlinear mapping
  approach to stain normalization in digital histopathology images using
  image-specific color deconvolution,'' {\em IEEE Transactions on Biomedical
  Engineering}, vol.~61, no.~6, pp.~1729--1738, 2014.

\bibitem{bentaieb2017adversarial}
A.~BenTaieb and G.~Hamarneh, ``Adversarial stain transfer for histopathology
  image analysis,'' {\em IEEE transactions on medical imaging}, vol.~37, no.~3,
  pp.~792--802, 2017.

\bibitem{piorkowski2018color}
A.~Pi{\'o}rkowski and A.~Gertych, ``Color normalization approach to adjust
  nuclei segmentation in images of hematoxylin and eosin stained tissue,'' in
  {\em International Conference on Information Technologies in Biomedicine},
  pp.~393--406, Springer, 2018.

\bibitem{sano2019fast}
K.~Sano, D.~Komura, and S.~Ishikawa, ``Fast and stable color normalization of
  whole slide histopathology images using deep texture and color moment
  matching,'' 2019.

\bibitem{zarella2017alternative}
M.~D. Zarella, C.~Yeoh, D.~E. Breen, and F.~U. Garcia, ``An alternative
  reference space for h\&e color normalization,'' {\em PloS one}, vol.~12,
  no.~3, 2017.

\bibitem{roberto2019classification}
G.~F. Roberto, M.~Z. Nascimento, A.~S. Martins, T.~A. Tosta, P.~R. Faria, and
  L.~A. Neves, ``Classification of breast and colorectal tumors based on
  percolation of color normalized images,'' {\em Computers \& Graphics},
  vol.~84, pp.~134--143, 2019.

\bibitem{yuan2018neural}
E.~Yuan and J.~Suh, ``Neural stain normalization and unsupervised
  classification of cell nuclei in histopathological breast cancer images,''
  {\em arXiv preprint arXiv:1811.03815}, 2018.

\bibitem{ziaei2020characterization}
D.~Ziaei, W.~Li, S.~Lam, W.-C. Cheng, and W.~Chen, ``Characterization of color
  normalization methods in digital pathology whole slide imaging,'' in {\em
  Medical Imaging 2020: Digital Pathology}, vol.~11320, p.~1132017,
  International Society for Optics and Photonics, 2020.

\bibitem{bianco2017improving}
S.~Bianco, C.~Cusano, P.~Napoletano, and R.~Schettini, ``Improving cnn-based
  texture classification by color balancing,'' {\em Journal of Imaging},
  vol.~3, no.~3, p.~33, 2017.

\bibitem{yang1999evaluation}
Y.~Yang, ``An evaluation of statistical approaches to text categorization,''
  {\em Information retrieval}, vol.~1, no.~1-2, pp.~69--90, 1999.

\bibitem{lever2016points}
J.~Lever, M.~Krzywinski, and N.~Altman, ``Points of significance:
  Classification evaluation,'' 2016.

\bibitem{manning2008matrix}
C.~D. Manning, P.~Raghavan, and H.~Sch{\"u}tze, ``Matrix decompositions and
  latent semantic indexing,'' {\em Introduction to Information Retrieval},
  pp.~403--417, 2008.

\bibitem{sebastiani2002machine}
F.~Sebastiani, ``Machine learning in automated text categorization,'' {\em ACM
  computing surveys (CSUR)}, vol.~34, no.~1, pp.~1--47, 2002.

\bibitem{tsoumakas2009mining}
G.~Tsoumakas, I.~Katakis, and I.~Vlahavas, ``Mining multi-label data,'' in {\em
  Data mining and knowledge discovery handbook}, pp.~667--685, Springer, 2009.

\bibitem{matthews1975comparison}
B.~W. Matthews, ``Comparison of the predicted and observed secondary structure
  of t4 phage lysozyme,'' {\em Biochimica et Biophysica Acta (BBA)-Protein
  Structure}, vol.~405, no.~2, pp.~442--451, 1975.

\bibitem{yonelinas2007receiver}
A.~P. Yonelinas and C.~M. Parks, ``Receiver operating characteristics (rocs) in
  recognition memory: a review.,'' {\em Psychological bulletin}, vol.~133,
  no.~5, p.~800, 2007.

\bibitem{japkowicz2002class}
N.~Japkowicz and S.~Stephen, ``The class imbalance problem: A systematic
  study,'' {\em Intelligent data analysis}, vol.~6, no.~5, pp.~429--449, 2002.

\bibitem{hanley1982meaning}
J.~A. Hanley and B.~J. McNeil, ``The meaning and use of the area under a
  receiver operating characteristic (roc) curve.,'' {\em Radiology}, vol.~143,
  no.~1, pp.~29--36, 1982.

\bibitem{pencina2008evaluating}
M.~J. Pencina, R.~B. D'Agostino, and R.~S. Vasan, ``Evaluating the added
  predictive ability of a new marker: from area under the roc curve to
  reclassification and beyond,'' {\em Statistics in medicine}, vol.~27, no.~2,
  pp.~157--172, 2008.

\bibitem{bradley1997use}
A.~P. Bradley, ``The use of the area under the roc curve in the evaluation of
  machine learning algorithms,'' {\em Pattern recognition}, vol.~30, no.~7,
  pp.~1145--1159, 1997.

\bibitem{hand2001simple}
D.~J. Hand and R.~J. Till, ``A simple generalisation of the area under the roc
  curve for multiple class classification problems,'' {\em Machine learning},
  vol.~45, no.~2, pp.~171--186, 2001.

\bibitem{syed2019mo1992}
S.~Syed, A.~Shrivastava, K.~Kant, S.~Sengupta, L.~Kang, M.~N. Khan, N.~T.
  Iqbal, K.~Sadiq, C.~A. Moskaluk, B.~Amadi, {\em et~al.}, ``Mo1992--solving
  the stain dilemma: Computational image analysis to address differential
  tissue staining color bias in duodenal biopsies,'' {\em Gastroenterology},
  vol.~156, no.~6, pp.~S--914, 2019.

\bibitem{shrivastava2019self}
A.~Shrivastava, W.~Adorno, L.~Ehsan, S.~A. Ali, S.~R. Moore, B.~C. Amadi,
  P.~Kelly, D.~E. Brown, and S.~Syed, ``Self-attentive adversarial stain
  normalization,'' {\em arXiv preprint arXiv:1909.01963}, 2019.

\bibitem{hou2016patch}
L.~Hou, D.~Samaras, T.~M. Kurc, Y.~Gao, J.~E. Davis, and J.~H. Saltz,
  ``Patch-based convolutional neural network for whole slide tissue image
  classification,'' in {\em Proceedings of the IEEE Conference on Computer
  Vision and Pattern Recognition}, pp.~2424--2433, 2016.

\bibitem{jain2010data}
A.~K. Jain, ``Data clustering: 50 years beyond k-means,'' {\em Pattern
  recognition letters}, vol.~31, no.~8, pp.~651--666, 2010.

\bibitem{kowsari2014investigation}
K.~Kowsari, {\em Investigation of fuzzyfind searching with golay code
  transformations}.
\newblock PhD thesis, M. Sc. Thesis, The George Washington University,
  Department of Computer Science, 2014.

\bibitem{kowsari2016weighted}
K.~Kowsari and M.~H. Alassaf, ``Weighted unsupervised learning for 3d object
  detection,'' {\em arXiv preprint arXiv:1602.05920}, 2016.

\bibitem{alassaf2015automatic}
M.~H. Alassaf, K.~Kowsari, and J.~K. Hahn, ``Automatic, real time, unsupervised
  spatio-temporal 3d object detection using rgb-d cameras,'' in {\em 2015 19th
  International Conference on Information Visualisation}, pp.~444--449, IEEE,
  2015.

\bibitem{yammahi2014efficient}
M.~Yammahi, K.~Kowsari, C.~Shen, and S.~Berkovich, ``An efficient technique for
  searching very large files with fuzzy criteria using the pigeonhole
  principle,'' in {\em 2014 Fifth International Conference on Computing for
  Geospatial Research and Application}, pp.~82--86, IEEE, 2014.

\bibitem{manning20introduction}
C.~D. Manning, P.~Raghavan, and H.~Schutze, ``Introduction to information
  retrieval, cambridge university press. 2008,'' {\em Ch}, vol.~20,
  pp.~405--416.

\bibitem{bianco2014error}
S.~Bianco and R.~Schettini, ``Error-tolerant color rendering for digital
  cameras,'' {\em Journal of mathematical imaging and vision}, vol.~50, no.~3,
  pp.~235--245, 2014.

\bibitem{li2014medical}
Q.~Li, W.~Cai, X.~Wang, Y.~Zhou, D.~D. Feng, and M.~Chen, ``Medical image
  classification with convolutional neural network,'' in {\em 2014 13th
  International Conference on Control Automation Robotics \& Vision (ICARCV)},
  pp.~844--848, IEEE, 2014.

\bibitem{scherer2010evaluation}
D.~Scherer, A.~M{\"u}ller, and S.~Behnke, ``Evaluation of pooling operations in
  convolutional architectures for object recognition,'' {\em Artificial Neural
  Networks--ICANN 2010}, pp.~92--101, 2010.

\bibitem{nair2010rectified}
V.~Nair and G.~E. Hinton, ``Rectified linear units improve restricted boltzmann
  machines,'' in {\em Proceedings of the 27th international conference on
  machine learning (ICML-10)}, pp.~807--814, 2010.

\bibitem{kingma2014adam}
D.~Kingma and J.~Ba, ``Adam: A method for stochastic optimization,'' {\em arXiv
  preprint arXiv:1412.6980}, 2014.

\bibitem{chollet2015keras}
F.~Chollet {\em et~al.}, ``Keras: Deep learning library for theano and
  tensorflow,'' {\em URL: https://keras. io/k}, 2015.

\bibitem{abadi2016tensorflow}
M.~Abadi, A.~Agarwal, P.~Barham, E.~Brevdo, Z.~Chen, C.~Citro, G.~S. Corrado,
  A.~Davis, J.~Dean, M.~Devin, {\em et~al.}, ``Tensorflow: Large-scale machine
  learning on heterogeneous distributed systems,'' {\em arXiv preprint
  arXiv:1603.04467}, 2016.

\bibitem{lee2009convolutional}
H.~Lee, R.~Grosse, R.~Ranganath, and A.~Y. Ng, ``Convolutional deep belief
  networks for scalable unsupervised learning of hierarchical
  representations,'' in {\em Proceedings of the 26th annual international
  conference on machine learning}, pp.~609--616, ACM, 2009.

\bibitem{kowsari2018web}
K.~Kowsari, D.~E. Brown, M.~Heidarysafa, K.~J. Meimandi, M.~S. Gerber, and
  L.~E. Barnes, ``Web of science dataset,'' {\em DOI: https://doi.
  org/10.17632/9rw3vkcfy4}, vol.~6, 2018.

\bibitem{mou2017deep}
L.~Mou, P.~Ghamisi, and X.~X. Zhu, ``Deep recurrent neural networks for
  hyperspectral image classification,'' {\em IEEE Transactions on Geoscience
  and Remote Sensing}, vol.~55, no.~7, pp.~3639--3655, 2017.

\bibitem{bengio1994learning}
Y.~Bengio, P.~Simard, and P.~Frasconi, ``Learning long-term dependencies with
  gradient descent is difficult,'' {\em IEEE transactions on neural networks},
  vol.~5, no.~2, pp.~157--166, 1994.

\bibitem{pascanu2013difficulty}
R.~Pascanu, T.~Mikolov, and Y.~Bengio, ``On the difficulty of training
  recurrent neural networks.,'' {\em ICML (3)}, vol.~28, pp.~1310--1318, 2013.

\bibitem{lai2015recurrent}
S.~Lai, L.~Xu, K.~Liu, and J.~Zhao, ``Recurrent convolutional neural networks
  for text classification.,'' in {\em AAAI}, vol.~333, pp.~2267--2273, 2015.

\bibitem{lecun1998gradient}
Y.~LeCun, L.~Bottou, Y.~Bengio, and P.~Haffner, ``Gradient-based learning
  applied to document recognition,'' {\em Proceedings of the IEEE}, vol.~86,
  no.~11, pp.~2278--2324, 1998.

\bibitem{johnson2014effective}
R.~Johnson and T.~Zhang, ``Effective use of word order for text categorization
  with convolutional neural networks,'' {\em arXiv preprint arXiv:1412.1058},
  2014.

\bibitem{shwartz2017opening}
R.~Shwartz-Ziv and N.~Tishby, ``Opening the black box of deep neural networks
  via information,'' {\em arXiv preprint arXiv:1703.00810}, 2017.

\bibitem{gray1996alternatives}
A.~Gray and S.~MacDonell, ``Alternatives to regression models for estimating
  software projects,'' 1996.

\bibitem{shrikumar2017learning}
A.~Shrikumar, P.~Greenside, and A.~Kundaje, ``Learning important features
  through propagating activation differences,'' {\em arXiv preprint
  arXiv:1704.02685}, 2017.

\bibitem{anthes2013deep}
G.~Anthes, ``Deep learning comes of age,'' {\em Communications of the ACM},
  vol.~56, no.~6, pp.~13--15, 2013.

\bibitem{lampinen2017one}
A.~K. Lampinen and J.~L. McClelland, ``One-shot and few-shot learning of word
  embeddings,'' {\em arXiv preprint arXiv:1710.10280}, 2017.

\bibitem{severyn2015learning}
A.~Severyn and A.~Moschitti, ``Learning to rank short text pairs with
  convolutional deep neural networks,'' in {\em Proceedings of the 38th
  international ACM SIGIR conference on research and development in information
  retrieval}, pp.~373--382, ACM, 2015.

\bibitem{fasano2003prevalence}
A.~Fasano, I.~Berti, T.~Gerarduzzi, T.~Not, R.~B. Colletti, S.~Drago,
  Y.~Elitsur, P.~H. Green, S.~Guandalini, I.~D. Hill, {\em et~al.},
  ``Prevalence of celiac disease in at-risk and not-at-risk groups in the
  united states: a large multicenter study,'' {\em Archives of internal
  medicine}, vol.~163, no.~3, pp.~286--292, 2003.

\bibitem{parzanese2017celiac}
I.~Parzanese, D.~Qehajaj, F.~Patrinicola, M.~Aralica, M.~Chiriva-Internati,
  S.~Stifter, L.~Elli, and F.~Grizzi, ``Celiac disease: From pathophysiology to
  treatment,'' {\em World journal of gastrointestinal pathophysiology}, vol.~8,
  no.~2, p.~27, 2017.

\bibitem{corazza2007comparison}
G.~R. Corazza, V.~Villanacci, C.~Zambelli, M.~Milione, O.~Luinetti,
  C.~Vindigni, C.~Chioda, L.~Albarello, D.~Bartolini, and F.~Donato,
  ``Comparison of the interobserver reproducibility with different histologic
  criteria used in celiac disease,'' {\em Clinical Gastroenterology and
  Hepatology}, vol.~5, no.~7, pp.~838--843, 2007.

\bibitem{gandomkar2018mudern}
Z.~Gandomkar, P.~C. Brennan, and C.~Mello-Thoms, ``Mudern: Multi-category
  classification of breast histopathological image using deep residual
  networks,'' {\em Artificial intelligence in medicine}, vol.~88, pp.~14--24,
  2018.

\bibitem{rakhlin2018deep}
A.~Rakhlin, A.~Shvets, V.~Iglovikov, and A.~A. Kalinin, ``Deep convolutional
  neural networks for breast cancer histology image analysis,'' in {\em
  International Conference Image Analysis and Recognition}, pp.~737--744,
  Springer, 2018.

\bibitem{motlagh2018breast}
N.~H. Motlagh, M.~Jannesary, H.~Aboulkheyr, P.~Khosravi, O.~Elemento,
  M.~Totonchi, and I.~Hajirasouliha, ``Breast cancer histopathological image
  classification: A deep learning approach,'' {\em bioRxiv}, p.~242818, 2018.

\bibitem{chougrad2018deep}
H.~Chougrad, H.~Zouaki, and O.~Alheyane, ``Deep convolutional neural networks
  for breast cancer screening,'' {\em Computer methods and programs in
  biomedicine}, vol.~157, pp.~19--30, 2018.

\bibitem{schaumberg2018h}
A.~J. Schaumberg, M.~A. Rubin, and T.~J. Fuchs, ``H\&e-stained whole slide
  image deep learning predicts spop mutation state in prostate cancer,'' {\em
  BioRxiv}, p.~064279, 2018.

\bibitem{korbar2017deep}
B.~Korbar, A.~M. Olofson, A.~P. Miraflor, C.~M. Nicka, M.~A. Suriawinata,
  L.~Torresani, A.~A. Suriawinata, and S.~Hassanpour, ``Deep learning for
  classification of colorectal polyps on whole-slide images,'' {\em Journal of
  pathology informatics}, vol.~8, 2017.

\bibitem{hu2018deep}
Z.~Hu, J.~Tang, Z.~Wang, K.~Zhang, L.~Zhang, and Q.~Sun, ``Deep learning for
  image-based cancer detection and diagnosis- a survey,'' {\em Pattern
  Recognition}, vol.~83, pp.~134--149, 2018.

\bibitem{aloise2009np}
D.~Aloise, A.~Deshpande, P.~Hansen, and P.~Popat, ``Np-hardness of euclidean
  sum-of-squares clustering,'' {\em Machine learning}, vol.~75, no.~2,
  pp.~245--248, 2009.

\bibitem{he2016deep}
K.~He, X.~Zhang, S.~Ren, and J.~Sun, ``Deep residual learning for image
  recognition,'' in {\em Proceedings of the IEEE conference on computer vision
  and pattern recognition}, pp.~770--778, 2016.

\bibitem{wen2018deep}
H.~Wen, J.~Shi, W.~Chen, and Z.~Liu, ``Deep residual network predicts cortical
  representation and organization of visual features for rapid
  categorization,'' {\em Scientific reports}, vol.~8, no.~1, p.~3752, 2018.

\bibitem{guillaumin2012large}
M.~Guillaumin and V.~Ferrari, ``Large-scale knowledge transfer for object
  localization in imagenet,'' in {\em 2012 IEEE Conference on Computer Vision
  and Pattern Recognition}, pp.~3202--3209, IEEE, 2012.

\bibitem{gad1982statistics}
S.~C. Gad and C.~S. Weil, ``Statistics for toxicologists,'' {\em Principles and
  methods of toxicology}, vol.~2, pp.~273--320, 1982.

\bibitem{oberhuber1999histopathology}
G.~Oberhuber, G.~Granditsch, and H.~Vogelsang, ``The histopathology of coeliac
  disease: time for a standardized report scheme for pathologists.,'' {\em
  European journal of gastroenterology \& hepatology}, vol.~11, no.~10,
  pp.~1185--1194, 1999.

\bibitem{pavik2013secreted}
I.~Pavik, P.~Jaeger, L.~Ebner, C.~A. Wagner, K.~Petzold, D.~Spichtig,
  D.~Poster, R.~P. W{\"u}thrich, S.~Russmann, and A.~L. Serra, ``Secreted
  klotho and fgf23 in chronic kidney disease stage 1 to 5: a sequence suggested
  from a cross-sectional study,'' {\em Nephrology Dialysis Transplantation},
  vol.~28, no.~2, pp.~352--359, 2013.

\bibitem{dumais2000hierarchical}
S.~Dumais and H.~Chen, ``Hierarchical classification of web content,'' in {\em
  Proceedings of the 23rd annual international ACM SIGIR conference on Research
  and development in information retrieval}, pp.~256--263, 2000.

\end{thebibliography}
\cleardoublepage
\addcontentsline{toc}{chapter}{References}

\end{document}